\documentclass[preprint]{elsarticle}

\usepackage{hyperref}

\journal{Artificial Intelligence Journal}

\usepackage{graphicx}
\usepackage{amsmath}
\usepackage{amsthm}
\usepackage{amssymb}
\usepackage{color}
\usepackage{cleveref}
\usepackage{tikz}
\usetikzlibrary{patterns,shapes,positioning,calc,decorations.pathreplacing}
\usepackage{multirow}
\usepackage{forest}
\usepackage[linesnumbered,ruled,vlined,noend]{algorithm2e}
\usepackage{subcaption}

\newtheorem{theorem}{Theorem}
\theoremstyle{remark}

\newtheorem{corollary}{Corollary}
\newtheorem{lemma}{Lemma}
\newtheorem{definition}{Definition}
\newtheorem{example}{Example}
\newtheorem{property}{Property}

\newcommand{\red}[1]{\textcolor{red}{#1}}
\newcommand{\noo}{\overline}
\newcommand{\scope}[1]{{\mathit{scope}(#1)}}
\newcommand{\tterminal}{$\top$-terminal}
\newcommand{\fterminal}{$\bot$-terminal}
\newcommand{\true}{$\mathit{true}$\xspace}
\newcommand{\false}{$\mathit{false}$\xspace}

\usepackage[normalem]{ulem}
\usepackage[xcolor,pdftex]{changebar}
\cbcolor{blue}

\usepackage[colorinlistoftodos]{todonotes}
\usepackage{soul}

\newcommand{\ull}[1]{\underline{\smash{#1}}}
\newcommand{\comment}[1]{}

\begin{document}
 	
\begin{frontmatter}

\title{SAT Encodings for Pseudo-Boolean Constraints Together With At-Most-One Constraints}

\author[udg]{Miquel Bofill}
\ead{miquel.bofill@imae.udg.edu}
\author[amu]{Jordi Coll\corref{mycorrespondingauthor}}
\ead{jordi.coll@lis-lab.fr}
\author[uoy]{Peter Nightingale}
\ead{peter.nightingale@york.ac.uk}
\author[udg]{Josep Suy}
\ead{josep.suy@imae.udg.edu}
\author[uoy]{Felix Ulrich-Oltean}
\ead{fvuo500@york.ac.uk}
\author[udg]{Mateu Villaret}
\ead{mateu.villaret@imae.udg.edu}

\cortext[mycorrespondingauthor]{Corresponding author}
\address[udg]{Universitat de Girona, Girona, Spain}
\address[amu]{Aix Marseille Univ, Universit\'e de Toulon, CNRS, LIS, Marseille, France}
\address[uoy]{University of York, York, United Kingdom}

\begin{abstract}

When solving a combinatorial problem using propositional satisfiability (SAT), the encoding of the problem is of vital importance. 
We study encodings of Pseudo-Boolean (PB) constraints, a common type of arithmetic constraint that appears in a wide variety of
combinatorial problems such as timetabling, scheduling, and resource allocation. 
In some cases PB constraints occur together with at-most-one (AMO) constraints over subsets of their variables (forming PB(AMO) constraints). 
Recent work has shown that taking account of AMOs when encoding PB constraints using decision diagrams can produce a dramatic improvement in solver efficiency. 
In this paper we extend the approach to other state-of-the-art encodings of PB constraints, developing several new encodings for PB(AMO) constraints. 
Also, we present a more compact and efficient version of the popular Generalized Totalizer encoding, named Reduced Generalized Totalizer. This new encoding is also adapted for PB(AMO) constraints for a further gain. 
Our experiments show that the encodings of PB(AMO) constraints can be substantially smaller than those of PB constraints. PB(AMO) encodings allow many more instances to be solved within a time limit, and solving time is improved by more than one order of magnitude in some cases. 
We also observed that there is no single overall winner among the considered encodings, but efficiency of each encoding may depend on PB(AMO) characteristics such as the magnitude of coefficient values.    

Published in Artificial Intelligence Journal: \url{https://doi.org/10.1016/j.artint.2021.103604}
  
\end{abstract}


\end{frontmatter}


\section{Introduction}\label{sec:intro}

Discrete decision-making problems crop up in many contexts in the modern world. 
 Such problems  can be expressed as constraint satisfaction (or optimisation) problems (CSPs or COPs), then solved using a variety of solver types. An increasingly popular and successful approach to solving CSPs and COPs is to encode them into Boolean formulas and then to apply an off-the-shelf SAT solver. This approach is attractive because of the power of modern conflict-directed clause learning (CDCL) SAT solvers, such as CaDiCaL~\cite{Biere-SAT-Race-2019-solvers} and Glucose~\cite{audemard2018glucose}, which incorporate conflict learning, powerful search heuristics, and fast propagation of the Boolean constraints.

Linear equations and inequalities are ubiquitous in constraint
problems such as  scheduling, routing, resource allocation, and many
other hard combinatorial problems.   Pseudo-Boolean (PB) constraints
are a particular type of linear constraint -- PB constraints
are of the form  $ \sum_{i=1}^{n} q_ix_i\: \#\: K $, where $\# \in
\{<,\leq,=,\geq,>\}$,  $q_1,\dots,q_n$ and $K$ are integer constants, and $x_1,\dots,x_n$ are 0/1 variables.  There has been a great deal of work on encoding PB constraints to SAT, some of which is reviewed by Philipp and Steinke~\cite{philipp2015pblib}. State-of-the-art encodings are based on Binary Decision Diagrams~\cite{een2006translating,abio2012new}, Sequential Weight Counters~\cite{holldobler2012compact}, Generalized Totalizers~\cite{joshi2015generalized,zha2019n}, and Polynomial Watchdog schemes~\cite{BailleuxBR09,manthey2014more}. At-most-one (AMO) constraints (i.e.\ constraints of the form $ \sum_{i=1}^{m} x_i\: \leq \: 1 $) are also very common, with the most basic being a mutual exclusion between two 0/1 variables. In this paper we extend several of the PB encodings mentioned above, and demonstrate substantially improved performance when PB constraints intersect with AMO constraints. 

Bofill, Coll, Suy, and Villaret~\cite{BCSV2017amopb,bofill2020mdd} proposed a SAT encoding based on Multi-valued Decision
Diagrams (MDDs) for a conjunction of a PB constraint
with a set of AMO constraints over the variables of the
PB constraint. Such conjunctions are referred to as PB(AMO)
constraints. The AMO constraints, which can be encoded to SAT in any
way, allow certain interpretations to be erased from decision diagrams,
and to represent the PB constraint as an MDD instead of as a Binary
Decision Diagram (BDD). The encoding of the MDD is notably smaller than
the encoding of an equivalent BDD, and the solving time is substantially
reduced. This technique has been used to provide efficient
formulations of particular kinds of scheduling problems \cite{BCSV2017amopb,BofillCSV17mrcpspmax}. Also,
Ans\'otegui et al~\cite{AnsoteguiBCDEMN19} integrated the MDD-based SAT encoding of PB(AMO)
constraints into the automatic reformulation pipeline of
Savile Row~\cite{aij-savilerow}, showing important size and solving
time improvements compared to a BDD-based encoding oblivious to the
existence of AMO constraints. 

Efficient encodings of the conjunction of PB and AMO constraints can have a significant impact on solving a wide range of CSPs. This combination of constraints appears in settings where one option has to be chosen among a set of incompatible options, and the decision has an associated cost. 
This pattern occurs in numerous applications, for example logistics~\cite{basnet2005heuristics}, resource allocation~\cite{ma1982task}, capital budgeting~\cite{pisinger2001budgeting}, telecommunications~\cite{watson2001packet}, combinatorial auctions~\cite{de2003combinatorial}, and routing~\cite{miller1960integer}, among many others. In short, any problem which is essentially a multi-choice knapsack problem is likely to contain both PB and AMO constraints.  Moreover, as stated in Bofill et al~\cite{bofill2020mdd}, any Linear Integer Arithmetic expression, which are also ubiquitous in CSP models, can be easily transformed to a PB(AMO) constraint. Therefore finding new and better SAT encodings of PB(AMO) constraints is of wide interest.

As a motivating example, consider the PB constraint \(2x_1 + 3x_2 + 4x_3 + 2x_4 + 3x_5 +4x_6 \leq 7\). 
Also, suppose there are two AMO constraints: \(x_1 + x_2 +x_3 \leq 1\) and \(x_4 + x_5 + x_6\leq 1\).  Encoding the PB constraint alone would require several clauses and (depending on the chosen encoding) multiple additional variables. For example, the Generalized Totalizer encoding~\cite{joshi2015generalized} has 23 additional variables and 56 clauses. However, the two AMO constraints rule out most of the values that the sum (\(2x_1 + \cdots +4x_6\)) could take, and almost all such values that break the PB constraint. Encoding the PB constraint together with the two AMO constraints requires just one clause to prevent \(x_3\) and \(x_6\) being assigned true together. This simple observation underpins all the PB(AMO) encodings presented in this paper.

\subsection{Contributions}

The main contribution of this paper is to generalize five state-of-the-art SAT encodings of PB
constraints to encode PB(AMO) constraints. We generalize each of the following encodings: 
Sequential Weight Counter (SWC), Generalized Totalizer (GT), n-Level Modulo
Totalizer (MTO), Global Polynomial Watchdog (GPW), and Local Polynomial
Watchdog (LPW). In each case we demonstrate substantial reductions in size, and also
improvements in solving time using two recent CDCL SAT solvers. Compared to their PB counterparts, 
PB(AMO) encodings allow many more instances to be solved within a time limit, and solving time is improved by more than one order of magnitude in some cases.

Since PB(AMO) constraints generalize PB constraints,
we follow the convention of naming the new encodings after the
original encoding, prefixing them with the word \emph{Generalized},
e.g., from the Sequential Weight Counter (SWC) encoding we provide the
Generalized Sequential Weight Counter (GSWC) encoding.  We also show
that the new encodings preserve the propagation properties of the
original ones.

Another contribution of this paper is a new encoding of PB(AMO) constraints 
called Reduced Generalized Generalized Totalizer (RGGT). This encoding does
not directly generalize any existing PB encoding. RGGT first constructs a Generalized Generalized Totalizer (GGT) tree, then applies a reduction algorithm in order to obtain a more compact representation that replaces individual numeric values with intervals.  In some cases RGGT will detect that terms in the PB(AMO) constraint are redundant. When this occurs the redundant terms are removed and the entire encoding process is repeated (until a fixpoint is reached). RGGT is frequently substantially smaller than GGT and this translates to improved solver efficiency. RGGT can also be used as an encoding of PB constraints (without collateral AMO constraints). In this case, we refer to it as the Reduced Generalized Totalizer (RGT) encoding. 
We also present a new heuristic called \textit{minRatio} to build the binary trees required by GT, GGT, RGT, and RGGT encodings. The experimental results show that the minRatio heuristic has a substantial positive effect on the size and performance of GT and GGT encodings when compared to a simple balanced tree. 

Our experimental results show that the size of the SAT encodings of PB
constraints can be dramatically reduced thanks to taking AMO
constraints into account, and that there can be a huge 
improvement in solving time when using the new generalized encodings. We
provide new benchmarks (and use others from the literature) which contain
AMO constraints and PB constraints in different configurations, and
we show empirically that some encodings are better than others for
particular kinds of PB(AMO).

The rest of the paper is structured as follows:
\begin{itemize}
	\item Section~\ref{sec:pre} presents preliminary concepts and notations used in this paper.
	\item Section~\ref{sec:encodings} describes the normalisation processes that we perform before encoding a PB(AMO) constraint.
	\item For the sake of completeness, Section~\ref{sec:enc_bdd} summarises the decision diagram encoding of PB constraints \cite{abio2012new} and its generalization to PB(AMO) constraints~\cite{bofill2020mdd}.
	\item Sections \ref{sec:enc_gswc}, \ref{sec:enc_ggt}, \ref{sec:enc_gmto}, \ref{sec:enc_ggpw}, and \ref{sec:enc_glpw} present the new PB(AMO) encodings. 	Each section describes the existing PB encoding followed by its generalization for PB(AMO) constraints. 
	\item Section~\ref{sec:pp} describes the propagation properties of the presented encodings.
	\item Section~\ref{sec:results} presents our experimental results comparing the PB(AMO) encodings to each other and also to their corresponding PB encoding. 
    \item Section~\ref{sec:related} surveys other work related to PB(AMO) encoding.
	\item Section~\ref{sec:conclusions} is devoted to conclusions and future work.
\end{itemize}

This work is an extension of the previous work~\cite{BofillCSV19};
here we provide more detailed explanations and examples, and the
following additional content: the set of preprocesses used to simplify
PB(AMO) constraints before they are encoded, in Section~\ref{sec:pre};
a summary of the PB(AMO) based on MDDs presented
in~\cite{bofill2020mdd} (in order to collect all existing PB(AMO)
encodings in this paper), in Section~\ref{sec:enc_bdd}; the Reduced
Generalized Totalizer (RGT) encoding for PB constraints and its
generalized version for PB(AMO) constraints in
Section~\ref{sec:enc_rggt}; the minRatio heuristic to build generalized
totalizers in Section~\ref{sec:heur_ggt}; the Generalized n-Level
Modulo Totalizer (GMTO) encoding in Section~\ref{sec:enc_gmto}; and the
Generalized Local Polynomial Watchdog (GLPW) encoding in
Section~\ref{sec:enc_glpw}.
We give further proofs of propagation properties of GGPW and the new encodings in Section~\ref{sec:pp}.
The experimental section has also been extended: we study in detail the size reduction achieved by the new encodings of PB and PB(AMO) constraints, namely RGT and RGGT; we study the results of the other new PB(AMO) encodings, namely GMTO and GLPW; 
we consider additional benchmark sets from the Combinatorial Auctions problem, the resource-constrained scheduling problems MRCPSP and RCPSP/t, and the Nurse Scheduling Problem;  and in addition to the original experiments run with Glucose, we run all experiments with
the SAT solver CaDiCaL.


\section{Preliminaries}\label{sec:pre}
A \emph{Boolean variable} is a variable than can take truth values 0
(false) and 1 (true).  A \emph{literal} is a Boolean variable $x$ or
its negation $\noo{x}$. A \emph{clause} is a disjunction of
literals. A \emph{propositional formula in conjunctive normal form}
(CNF) is a conjunction of clauses.  We will assume that all formulas
are in CNF. Clauses are usually seen as sets of literals, and formulas
as sets of clauses. A \emph{Boolean function} is a function of the
form $f: \{0,1\}^n \rightarrow \{0,1\}$.

We will consider constraints that are defined over a finite set of
Boolean variables, i.e., Boolean functions. An \emph{assignment} is a
mapping of Boolean variables to truth values; it can also be seen as a
set of literals, e.g. $\{x=1,y=0,z=0\}$ is usually denoted
$\{x,\noo{y},\noo{z}\}$.   By $\scope{C}$ we denote the set of
variables occurring in a constraint $C$. 

A~\emph{satisfying assignment} of a Boolean
function $f$ is an assignment that makes $f$ evaluate to 1. In
particular, an assignment $A$ satisfies a formula $F$ in CNF if at
least one literal $l$ of each clause in $F$ belongs to $A$. Such an
assignment is called a \emph{model} of the formula.  Given two Boolean
functions $F$ and $G$, we say that $G$ is logically implied by $F$,
iff every model of $F$ is also a model of $G$.

\begin{definition}
  An \emph{at-most-one} (AMO) constraint is a Boolean function of the
  form $\sum_{i=1}^{n} x_i \leq 1$, where all $x_i$ are Boolean 
  variables.
\end{definition}

\begin{definition}
  A \emph{pseudo-Boolean} (PB) constraint is a Boolean function of the
  form $ \sum_{i=1}^{n} q_ix_i \,\#\, K $ where $K$ and all $q_i$ are
  integer constants, all $x_i$ are Boolean variables, and
  $\# \in \{<, \leq, =, \geq, >\}$.
\end{definition}

\begin{definition}
  By \emph{PB(AMO)} constraint we refer to a constraint of the form
  $P \land M_1 \land \dots \land M_N$, where $P$ is a PB constraint,
  and $M_1,\dots,M_N$ are AMO constraints such that
  $\{\mathit{scope}(M_1),\dots,\mathit{scope}(M_N)\}$ is a partition
  of $\mathit{scope}(P)$.
\end{definition}

Since a single variable constitutes a trivial at-most-one
constraint, PB constraints are a particular case of PB(AMO)
constraints, i.e., a PB constraint of the form
$\sum_{i=1}^{n} q_ix_i \leq K$ is a PB(AMO) constraint of the form
$\sum_{i=1}^{n} q_ix_i \leq K \land x_1 \leq 1 \land \dots \land x_n
\leq 1$.

\begin{example}
  $2x_1+3x_2+ 3x_3+7x_4\leq 8 \wedge x_1+x_2\leq 1 \wedge x_3+x_4\leq
  1$ is a PB(AMO) constraint. Notice that, for instance, the assignment
  ${x_1,x_2,x_3,\noo{x_4}}$ satisfies the PB constraint
  $2x_1+3x_2+ 3x_3+7x_4\leq 8$ but does not satisfy the PB(AMO)
  constraint because of $x_1+x_2\leq 1$.
\end{example}

\begin{definition}[Encoding]
We say that a formula $G$ is an \emph{encoding} of a Boolean function
$F$ if the following holds: given an assignment $A$ over the variables
of $F$, $A$ satisfies $F$ iff $A$ can be extended to a satisfying
assignment of $G$.
\end{definition}

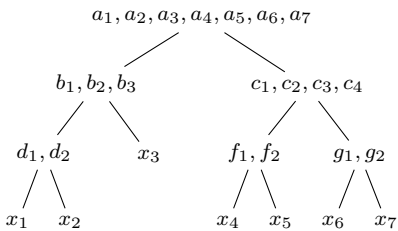
\begin{figure}[!t]
	\centering
	\footnotesize
		\begin{tikzpicture}[auto,node distance= 0.7cm and 0.1cm,>=stealth]
		\tikzstyle{tex}=[]

		\node[tex] (H) []             {$x_1$};
		\node[tex] (I) [right of = H] {$x_2$};
		\node[tex] (J) [right of = I] {\phantom{$x_0$}};
		\node[tex] (K) [right of =J] {\phantom{$x_4$}};
		\node[tex] (L) [right of =K] {$x_4$};
		\node[tex] (M) [right of = L] {$x_5$};
		\node[tex] (N) [right of =M] {$x_6$};
		\node[tex] (O) [right of = N] {$x_7$};

		\coordinate (Middle) at ($(H)!0.5!(I)$); \node[tex] (D) [above = of Middle] {$d_1,d_2$};
		\coordinate (Middle) at ($(J)!0.5!(K)$); \node[tex] (E) [above = of Middle] {$x_3$};
		\coordinate (Middle) at ($(L)!0.5!(M)$); \node[tex] (F) [above = of Middle] {$f_1,f_2$};
		\coordinate (Middle) at ($(N)!0.5!(O)$); \node[tex] (G) [above = of Middle] {$g_1,g_2$};

		\coordinate (Middle) at ($(D)!0.5!(E)$); \node[tex] (B) [above = of Middle] {$b_1,b_2,b_3$};
		\coordinate (Middle) at ($(F)!0.5!(G)$); \node[tex] (C) [above = of Middle] {$c_1,c_2,c_3,c_4$};

		\coordinate (Middle) at ($(B)!0.5!(C)$); \node[tex] (A) [above = of Middle] {$a_1,a_2,a_3,a_4,a_5,a_6,a_7$};

		\path[-] (A) edge (B); \path[-] (A) edge (C);

		\path[-] (B) edge (D); \path[-] (B) edge (E);
		\path[-] (C) edge (F); \path[-] (C) edge (G);

		\path[-] (D) edge (H); \path[-] (D) edge (I);
		\path[-] (F) edge (L); \path[-] (F) edge (M);
		\path[-] (G) edge (N); \path[-] (G) edge (O);

		\end{tikzpicture}
\caption[]{Tree representation of a totalizer for variables $x_1,x_2,x_3,x_4,x_5,x_6,x_7$.}
\label{fig:totalizer}
\end{figure}

A large number of encodings have been proposed for AMO
constraints and PB constraints in the literature, as well as encodings
for \emph{cardinality constraints}, i.e., PB constraints with all
coefficients $q_i=1$. 
Here we revisit some basics of the Totalizer encoding for cardinality constraints~\cite{BailleuxB03}, since it is closely related to many of the encodings presented in the paper.
Given a set of variables $x_1,\dots,x_n$, a \emph{totalizer} is a binary tree that contains a different variable $x_i$ at each leaf. An example is given in Figure~\ref{fig:totalizer}. For any subtree, the root of the subtree contains a list with as many Boolean variables as leaves in the subtree. These variables are constrained to represent in unary notation the sum of the variables of the leaves of the subtree or, equivalently, the list will be a decreasing ordering of the values of the leaves. In particular, the list of variables $a_1,\dots,a_n$ contained in the root of the totalizer represents the decreasing ordering of the values of $x_1,\dots,x_n$. 
Then, an encoding of the cardinality constraint $\sum_{i=1}^n x_i \leq K$ is completed by forbidding that variable $a_{K+1}$ is true.

\section{Encoding Technique and Normalisation of PB(AMO) Constraints}\label{sec:encodings}

Given a PB(AMO) constraint of the form $P \land M_1 \land \dots
\land M_N$, a straightforward approach to encode it is to generate a
formula $F$ of the form $G \land H_1 \land \dots \land H_N$, where $G$
is an encoding of $P$, and $H_i$ is an encoding of $M_i$ for all $i\in
1..N$.
Instead, similarly to the MDD-based approach of~\cite{bofill2020mdd}, we
propose to encode PB(AMO) constraints in a combined way.  
We encode the conjunction of AMO constraints in the usual way,
i.e., each AMO is encoded separately and we use the conjunction of all the
resulting clauses.
However, we encode the PB constraint assuming that the accompanying
AMO constraints are already enforced. This is what will let us
significantly reduce the size of the PB constraint encoding.
 Lemma~\ref{lem:pbamo-correct} states the correctness of our encoding technique.
\begin{lemma}\label{lem:pbamo-correct}
	Let $\cal P$ be a PB(AMO) constraint of the form
	$P\land M_1 \land \dots \land M_N$, where ${\cal X}=\{X_1,\dots,X_N\}$
	is a partition of the variables in $P$ such that
	$X_i = \mathit{scope(M_i)}$. Let $E$ be a Boolean formula such that, given any assignment $A$ of the variables of $\cal P$ that satisfies $M_1 \land \dots \land M_N$, $A$ can be extended to a model of $E$ iff $A$ satisfies $P$. Then the conjunction of
	$E$ with an encoding of
	$M_1 \land \dots \land M_N$ is an encoding of $\cal P$.
\end{lemma}
The encoding procedures described in the following sections generate the $E$ formula of Lemma~\ref{lem:pbamo-correct}. It is assumed
that the AMO constraints \(M_1\) to \(M_N\) are encoded elsewhere.

We will not restrict ourselves to any particular encoding for the AMO
constraints when encoding a PB(AMO) constraint. Furthermore, in the context of a larger formula, if the
AMO constraints are not explicit but logically implied by the formula,
then the encoding of the PB constraint will suffice to obtain a
correct encoding of the PB(AMO) constraint.

Given a PB(AMO) constraint ${\cal P}$ of the form
$P \land M_1 \land \dots \land M_N$, the encoding procedures described
in the following sections take as input the pair $(P,{\cal X})$, where
${\cal
  X}=\{X_1,\dots,X_N\}=\{\mathit{scope}(M_1),\dots,\mathit{scope}(M_N)\}$
is a partition of the variables of the PB constraint $P$. 
Moreover, we soundly preprocess the PB constraint $P$ and the partition
${\cal X}$ so that they satisfy a set of properties. Some of the
properties are required for some of the encodings presented in this
paper, while others are used to reduce the size of the
constraints or to deal efficiently with trivial constraints or
monomials in the PB constraints. The properties are enforced in the order they are given, 
and clauses and variables may be introduced at each step. 

\begin{property}\label{prop:monotonicdecreasing}
	 $P$ is of the form $\sum_{i=1}^nq_ix_i \leq K$,
  with $q_i \geq 0$.
\end{property}

A PB(AMO) satisfying this property is a monotonic decreasing Boolean function, i.e.\ any of its models remains a model after flipping assignments from 1 to 0. 
The PB constraint $P$ of a PB(AMO) constraint can always be
transformed to have this property~\cite{bofill2020mdd}. If $P$ is of
the form $\sum_{i=1}^nq_ix_i = K$, it is replaced by the conjunction
of two constraints $\sum_{i=1}^nq_ix_i \leq K$ and
$\sum_{i=1}^nq_ix_i \geq K$, both with partition ${\cal X}$.
The operators $\geq$ and $>$ can be transformed to $\leq$ or $<$,
respectively, by multiplying both sides of the inequality by
$-1$. Furthermore, $<$ can be replaced by $\leq$ by subtracting $1$
from the right-hand side. Finally, in order to get only positive
coefficients, we proceed as follows. For each AMO group $X_i$ with
some negative coefficient in the PB constraint, consider such minimum
negative coefficient $q$ and perform the following transformation:
increase by $-q$ all coefficients of the variables of the AMO-group,
as well as $K$. Moreover, add a new variable $y$ defined by
$y\leftrightarrow \bigwedge_{x_l{\in}X_i}\noo{x_l}$ and add it to the
PB constraint with coefficient $-q$. This way the inequality is
preserved, since the left-hand side will be incremented exactly by
$-q$ whatever variables are set to true, thanks to the AMO
constraint. As an example, consider the PB constraint
$4x_1-7x_2+\dots\leq 15$, and assume $x_1$ and $x_2$ constitute an AMO
group, i.e., we have $x_1+x_2\leq 1$. According to our transformation
we would get $11x_1+0x_2+7y+\dots\leq 22$, with
$y\leftrightarrow \noo{x_1}\land\noo{x_2}$. Observe that, on the one
hand, the contribution of each variable of the AMO group is
incremented by $7$, so the right-hand side must be incremented by $7$
as well and, on the other hand, if neither of the variables of the AMO
group is set to true, then $y$ contributes $7$ to the left-hand
side, so the original relation is preserved. Note moreover that the variables
originally with the minimum coefficient ($x_2$ in the example) can be
removed from the PB constraint, as their coefficient becomes $0$ after
the transformation.

\begin{property}\label{prop:Kgt0}
$K>0$.
\end{property}
If $K<0$ (the constraint is
  unsatisfiable) the encoding of $(P,{\cal X})$ is just the empty
  clause. If $K=0$, the encoding is just the union
  of the unit clauses $\noo{x_l}$ for all terms \(q_lx_l\) where $q_l > 0$.

\begin{property}\label{prop:qgt0}
	All coefficients satisfy $q_l > 0$.
\end{property}
Otherwise,
  for any $q_l=0$ where $x_l \in X_i$, remove the term $q_lx_l$ from
  $P$, remove $x_l$ from $X_i$, and remove any set $X_i \in {\cal X}$
  that becomes empty.

\begin{property}\label{prop:qleqK}
	All coefficients satisfy $q_l \leq K$.
\end{property}
This can
  be achieved by adding the unit clause $\noo{x_l}$ for all $q_l >
  K$. For such $q_l$, where $x_l \in X_i$, remove the corresponding
  monomial $q_lx_l$ from $P$, remove $x_l$ from $X_i$, and
  remove any set $X_i \in {\cal X}$ that becomes empty.

\begin{property}\label{prop:ngt1}
	$N > 1$.
\end{property}
Otherwise, the encoding of
  $(P,{\cal X})$ is just the empty CNF (note that this is sound given
  Property~\ref{prop:qleqK} and the AMOs assumption).

\begin{property}\label{prop:notautology}
	$\sum_{i=1}^N \max_{x_l \in X_i}(q_l) > K$.
\end{property}
This property means that the constraint is not trivially true when all AMO constraints
  hold. Otherwise the encoding of $(P,{\cal X})$ is just the empty
  CNF.

\begin{property}\label{prop:difcoeffs}
	There are no two variables $x_l,x_{l'}$
  belonging to the same $X_i$ with the same coefficient
  $q_l=q_{l'}$.
\end{property}
   Otherwise, for all maximal subsets of variables
  $X'_{i}\subseteq X_i$ with the same coefficient $q$ (i.e. 
  $\forall x_l{\in}X'_{i},\; q_l=q$): add an auxiliary
  variable $y$; update $X_i$ as $X_i=(X_i \setminus X_i') \cup \{y\}$;
  remove from $P$ all monomials $q_lx_l$, for all $x_l{\in}X'_{i}$;
  add to $P$ the monomial $qy$; and add to the encoding of $(P,{\cal X})$
  the clauses $\noo{x_l} \lor y$, for all $x_l{\in}X'_i$.

\textbf{Remark:} Henceforth, we assume that PB(AMO) constraints satisfy all these properties.


\section{Decision Diagram encoding}\label{sec:enc_bdd}

In order to cover all existing encodings of PB(AMO) constraints, here we provide an overview of using Binary Decision Diagrams (BDDs) to encode PB constraints following~\cite{abio2012new}, and in Subsection~\ref{ssec:enc_mdd} we provide an overview of using  Multivalued Decision Diagrams (MDDs) to encode PB(AMO) constraints as is done in~\cite{bofill2020mdd}.

\subsection{Binary Decision Diagram}\label{ssec:enc_bdd}

Many BDD-based encodings of PB constraints have been proposed. In this
work we consider the encoding for monotonic decreasing PB constraints
presented in~\cite{abio2012new}, which in this
paper will be referred to as the \emph{BDD} encoding. 
Given a PB constraint of the form $\sum_{i=1}^{n} q_ix_i \leq K$ with
$q_i\geq 0$ for all $i$, the BDD encoding consists of, first of all,
representing the constraint as a Reduced Ordered BDD (ROBDD), and then encoding that BDD as a set of clauses. We use the following definition of a BDD:

\begin{definition}[Definition 5 in \cite{bofill2020mdd}]
	A \emph{Binary Decision Diagram} (BDD) is a rooted, directed,
        acyclic graph which represents a Boolean function. BDDs have
        two terminal nodes, namely \fterminal{} and
        \tterminal{}.  Each nonterminal node has an associated Boolean
        variable (\emph{selector}), and two outgoing edges,
        representing the \true and the \false assignment of the
        selector. Every truth assignment of the variables follows a
        path from the root to the \tterminal{} when it satisfies the
        formula, or to the \fterminal{} otherwise.
\end{definition}

A BDD is called \emph{ordered} if different variables appear in the
same order on all paths from the root.  A BDD is said to be
\emph{reduced} if it satisfies the following two conditions: it contains no isomorphic sub-BDDs, and there is no node whose true and false child are the same.

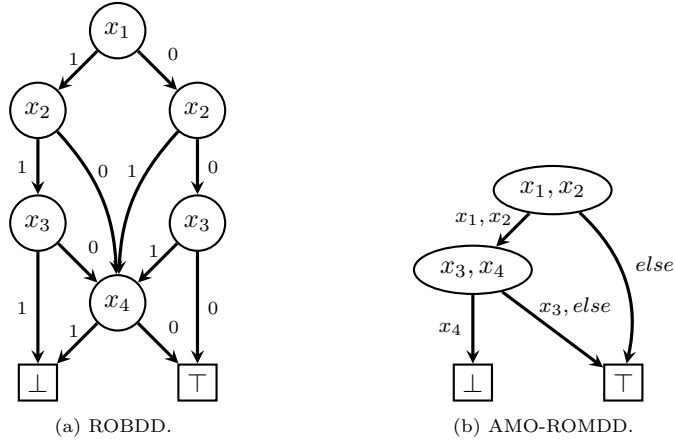
\begin{figure}[!ht]
	\centering
\subcaptionbox{ROBDD.\label{fig:bdd}}{
	\begin{tikzpicture}[auto,node distance=1.5cm,>=stealth]
	\tikzstyle{nonterminal}=[circle,thick,draw,minimum size=4mm]
	\tikzstyle{terminal}=[rectangle,thick,draw,minimum size=4mm]
	\tikzstyle{labelfont}=[font=\scriptsize,line width=1.2pt]
	\node[nonterminal] (1) [] {$x_1$};
	\node[nonterminal] (2) [below left of = 1] {$x_2$};
	\node[nonterminal] (3) [below right of = 1] {$x_2$};
	\node[nonterminal] (4) [below of = 2] {$x_3$};
	\node[nonterminal] (5) [below of = 3] {$x_3$};
	\node[nonterminal] (6) [below right of = 4] {$x_4$};
	\node[terminal] (T) [below right of = 6] {$\top$};
	\node[terminal] (F) [below left of = 6] {$\bot$};
	
	\path[->,labelfont] (1) edge  [left] node [pos=0.2] {1} (2);
	\path[->,labelfont] (1) edge node {0} (3);
	\path[->,labelfont] (2) edge [left] node {1} (4);
	\path[->,labelfont] (2) edge [right, bend left = 20] node  [pos=0.3]{0} (6);
	\path[->,labelfont] (3) edge node {0} (5);
\path[->,labelfont] (3) edge [left, bend right = 20] node [pos=0.3] {1} (6);
	\path[->,labelfont] (4) edge [left] node {1} (F);
	\path[->,labelfont] (4) edge node {0} (6);
	\path[->,labelfont] (5) edge node {0} (T);
	\path[->,labelfont] (5) edge [left] node [pos=0.2] {1} (6);
	\path[->,labelfont] (6) edge node {0} (T);
	\path[->,labelfont] (6) edge [left] node [pos=0.2] {1} (F);
	\end{tikzpicture}
}
	\hspace{2cm}
\subcaptionbox{AMO-ROMDD.\label{fig:mdd}}{
	\begin{tikzpicture}[auto,node distance=1.5cm,>=stealth]
	\tikzstyle{nonterminal}=[ellipse,thick,draw,minimum size=4mm]
	\tikzstyle{terminal}=[rectangle,thick,draw,minimum size=4mm]
	\tikzstyle{labelfont}=[font=\footnotesize,line width=1.2pt]
	\node[nonterminal] (1) [] {$x_1,x_2$};
	\node[nonterminal] (2) [below left of = 1] {$x_3,x_4$};
	\node[terminal] (F) [below of = 2] {$\bot$};
	\node[terminal] (T) [right of = F,node distance=2cm] {$\top$};
	
	\path[->,labelfont] (1) edge [left] node [pos=0.2]{$x_1,x_2$} (2);
	\path[->,labelfont] (1) edge  [bend left = 30] node {$else$} (T);
	\path[->,labelfont] (2) edge [left]  node {$x_4$} (F);
	\path[->,labelfont] (2) edge  [anchor=west, pos=0.2] node [outer sep=3pt] {$x_3,else$} (T);
	\end{tikzpicture}
}
	\caption[]{(a): ROBDD for the PB constraint $2x_1 + 3x_2 + 4x_3 + 7x_4 \leq 8$ with variable ordering $x_1 \prec x_2 \prec x_3 \prec x_4$. (b): AMO-ROMDD for $2x_1 + 3x_2 + 4x_3 + 7x_4 \leq 8$,
		with ordered partition $\{x_1, x_2\} \prec \{x_3, x_4\}$. Multiple edges between two nodes are represented as a single edge with multiple labels.}
\end{figure}

Figure~\ref{fig:bdd} contains an example BDD representing a PB constraint. There exist algorithms to construct a ROBDD representing a PB constraint that run in polynomial time w.r.t. the size of the resulting ROBDD.

 Once the ROBDD has been constructed, an auxiliary variable $v$ is
introduced for each node. The encoding enforces $v$ to be false
whenever the sub-ROBDD rooted at that node follows a path
to the \fterminal{} with a given assignment. That is, it enforces
$\noo{v_0}\land\noo{x_i}\to\noo{v}$ and
$\noo{v_1}\land x_i\to\noo{v}$, where $x_i$ is the variable in the
node of the ROBDD, and $v_0$ and $v_1$ are the auxiliary variables of
the false child and the true child of that node,
respectively. However, since the considered constraints are monotonic,
we can simplify the constraint $\noo{v_0}\land\noo{x_i}\to\noo{v}$ to
$\noo{v_0}\to\noo{v}$, because if $x_i=0$ falsifies the PB constraint,
so will $x_i=1$. Therefore, for each nonterminal node, the encoding
introduces the following clauses:
\begin{align}
v_0 \lor \noo{v} \label{eq:bdd_fchild}\\
v_1 \lor \noo{x_i} \lor \noo{v} \label{eq:bdd_tchild}
\end{align}

The encoding is completed by adding three unary clauses:
\begin{align}
v_r \quad \land \quad \noo{v_\bot} \quad \land \quad v_\top \label{eq:bdd_unary}
\end{align}
\noindent where $v_r$, $v_\bot$ and  $v_\top$ are the auxiliary variables of the root node, the \fterminal{} and the \tterminal{} respectively.

\subsection{Multivalued Decision Diagram}\label{ssec:enc_mdd}

Bofill, Coll, Suy, and Villaret~\cite{BCSV2017amopb,bofill2020mdd} presented a generalization of the BDD encoding for PB(AMO) constraints. Like the new encodings presented in this paper, the MDD encoding receives as input a PB constraint $P$ and a partition $\cal X$ of its variables. 
This encoding uses at-most-one Multivalued Decision Diagrams (AMO-MDD), which is a more generic structure than a BDD. In an AMO-MDD, as defined in~\cite{bofill2020mdd}, each node contains a set of selector variables. Given the partition ${\cal X} = \{X_1,\dots,X_N\}$, the nodes of the $i$-th layer of the AMO-MDD contain the variables of $X_i$ as selectors. An example is given in  Figure~\ref{fig:mdd}. For each node there are multiple possible choices (outgoing edges): assigning one of the selectors to true, or assigning all of them to false (the \emph{else} edge). Therefore, the AMO-MDDs correctly represent all assignments where at most one variable in each set $X_i$ is true. Similarly to BDD, first of all an AMO-ROMDD is constructed, given an order of the elements of ${\cal X}$. Then, one auxiliary variable is introduced for each node and Clauses~\eqref{eq:bdd_unary} are introduced. Also, for each nonterminal node, the following clauses are introduced:
\begin{align}
&v_{0} \lor \noo{v}\label{eq:mdd_else}\\
&v_j \lor \noo{x_j} \lor \noo{v}
& \forall x_j \in X_i \text{ s.t. } v_j \neq v_{0}\label{eq:mdd_selector}
\end{align}
\noindent where $v$ is the auxiliary variable of the encoded node, $v_j$ is the auxiliary
variable of the child node selected by the selector variable $x_j$, and
$v_{0}$ is the auxiliary variable of the \emph{else} child.

The BDD encoding requires $O(nK)$ auxiliary variables and $O(nK)$ clauses,
while the AMO-MDD encoding requires $O(NK)$ auxiliary variables and
$O(nK)$ clauses.


\section{Sequential Weight Counter Encoding}\label{sec:enc_gswc}
In this section we first recall  the Sequential Weight Counter encoding for PB constraints
from~\cite{holldobler2012compact} and in Subsection~\ref{ssec:enc_gswc} we provide its generalization to encode PB(AMO) constraints.

\subsection{Sequential Weight Counter}\label{ssec:enc_swc}

The idea of the \emph{Sequential Weight Counter} (SWC) encoding for PB constraints
 is to build a circuit that sequentially sums from left to
right the coefficients (a.k.a.\ weights) $q_i$ whose variable $x_i$ is
set to true.
Specifically, given a PB constraint
$\sum_{i=1}^{n} q_ix_i \leq K$, there is a sequence of $n$ counters of
$K$ inputs and $K$ outputs, where the $i$-th counter is associated to
the variable $x_i$. Each counter receives as input a vector of Boolean
variables, which is the unary representation of an integer value, and
adds the weight $q_i$ to the output if the associated variable $x_i$
is set to true. Therefore, the $i$-th counter receives as input
$\sum_{j=1}^{i-1}q_jx_j$ and outputs $\sum_{j=1}^{i}q_jx_j$. Note that
the output of the counter number $i-1$ is the input of the $i$-th
counter.

An example of a sequence of counters is shown in
Figure~\ref{fig:swc}.  The encoding introduces $n\cdot K$ variables,
denoted $s_{i,j}$, with $1 \leq i \leq n$, $1 \leq j \leq K$, where
$s_{i,j}$ is the $j$-th output of the $i$-th counter and
also the $j$-th input of the $(i+1)$-th counter.  The encoding
introduces the following clauses:
\begin{align}
 \noo{s_{i-1,j}} \lor s_{i,j} & \qquad 2 \leq i < n, 1 \leq j \leq K\label{eq:swcelse} \\
 \noo{x_i}  \lor s_{i,j} & \qquad 1 \leq i < n, 1 \leq j \leq q_i \label{eq:swczerosum} \\
 \noo{s_{i-1,j}} \lor \noo{x_i} \lor s_{i,j+q_i} & \qquad 2 \leq i < n, 1 \leq j \leq K - q_i\label{eq:swcsum} \\
 \noo{s_{i-1,K+1-q_i}} \lor \noo{x_i} & \qquad 2 \leq i \leq n\label{eq:swcfail}
\end{align}
where $s_{0,j}$ is the constant 0 for all $j$, to represent the input
of the first counter which is the empty sum.  Clauses
\eqref{eq:swcelse} state that
$\sum_{j=1}^{i}{q_jx_j} \geq \sum_{j=1}^{i-1}{q_jx_j}$. Clauses
\eqref{eq:swczerosum} and \eqref{eq:swcsum} enforce that if a variable
$x_i$ is true then its coefficient is added to the input of the next
counter.  Finally, Clauses~\eqref{eq:swcfail} enforce that the sum
never exceeds $K$.
\begin{figure}[!t]
\centering
\footnotesize
\subcaptionbox{SWC.\label{fig:swc}}{
\resizebox{10cm}{!}{
\begin{tikzpicture}[auto,node distance=1.5cm,>=stealth]
\tikzstyle{box}=[rectangle,thick,draw,minimum width=1cm,minimum height=3.4cm,anchor=south west]
\tikzstyle{txt}=[anchor=south]
\foreach \x in {0,...,7}{
	\pgfmathtruncatemacro{\y}{8-\x}
	\draw (-0.2, 0.3 + 0.4*\x) -- (0, 0.3 + 0.4*\x);
	\node [txt] at (-0.8, 0.2 + 0.4*\x) [] {$s_{0,\y} = 0$};

}
\node[box] at (0,0) [] {$+2x_1$};
\foreach \x in {0,...,7}{
	\pgfmathtruncatemacro{\y}{8-\x}
	\draw (1, 0.3 + 0.4*\x) -- (2,0.3 + 0.4*\x);
	\node [txt] at (1.5, 0.2 + 0.4*\x) [] {$s_{1,\y}$};
}
\node[box] at (2,0) [] {$+3x_2$};
\foreach \x in {0,...,7}{
	\pgfmathtruncatemacro{\y}{8-\x}
	\draw (3, 0.3 + 0.4*\x) -- (4,0.3 + 0.4*\x);
	\node [txt] at (3.5, 0.2 + 0.4*\x) [] {$s_{2,\y}$};
}
\node[box] at (4,0) [] {$+4x_3$};
\foreach \x in {0,...,7}{
	\pgfmathtruncatemacro{\y}{8-\x}
	\draw (5, 0.3 + 0.4*\x) -- (6,0.3 + 0.4*\x);
	\node [txt] at (5.5, 0.2 + 0.4*\x) [] {$s_{3,\y}$};
}
\node[box] at (6,0) [] {$+7x_4$};
\foreach \x in {0,...,7}{
	\pgfmathtruncatemacro{\y}{8-\x}
	\draw (7, 0.3 + 0.4*\x) -- (7.2,0.3 + 0.4*\x);
	\node [txt] at (7.5, 0.2 + 0.4*\x) [] {$s_{4,\y}$};
}
\end{tikzpicture}
}
}

\vspace{0.5cm}
\subcaptionbox{GSWC.\label{fig:gswc}}{
\resizebox{7cm}{!}{
\footnotesize
\begin{tikzpicture}[auto,node distance=1.5cm,>=stealth]
\tikzstyle{box}=[rectangle,thick,draw,minimum width=1.5cm,minimum height=3.4cm,anchor=south west]
\tikzstyle{txt}=[anchor=south]
\foreach \x in {0,...,7}{
	\pgfmathtruncatemacro{\y}{8-\x}
	\draw (-0.2, 0.3 + 0.4*\x) -- (0, 0.3 + 0.4*\x);
	\node [txt] at (-0.8, 0.2 + 0.4*\x) [] {$s_{0,\y} = 0$};
}
\draw [decorate,decoration={brace,amplitude=3pt},xshift=-4pt,yshift=0pt]
(0.8,1.3) -- (0.8,2.2) node [black,midway,xshift=-0.1cm]
{\footnotesize $+$};
\node[box] at (0,0) [align=right] { \phantom{....} $2x_1$ \\ \phantom{....} $3x_2$};

\foreach \x in {0,...,7}{
	\pgfmathtruncatemacro{\y}{8-\x}
	\draw (1.5, 0.3 + 0.4*\x) -- (2.5,0.3 + 0.4*\x);
	\node [txt] at (2, 0.2 + 0.4*\x) [] {$s_{1,\y}$};
}
\draw [decorate,decoration={brace,amplitude=3pt},xshift=-4pt,yshift=0pt]
(3.3,1.3) -- (3.3,2.2) node [black,midway,xshift=-0.1cm]
{\footnotesize $+$};
\node[box] at (2.5,0) [align=left] {\phantom{....} $4x_3$ \\ \phantom{....} $7x_4$};
\foreach \x in {0,...,7}{
	\pgfmathtruncatemacro{\y}{8-\x}
	\draw (4, 0.3 + 0.4*\x) -- (4.2,0.3 + 0.4*\x);
	\node [txt] at (4.5, 0.2 + 0.4*\x) [] {$s_{2,\y}$};
}
\end{tikzpicture}
}
}
\caption[]{(a): high level circuit representation of $\mathit{SWC}(2x_1 + 3x_2 + 4x_3 + 7x_4 \leq 8)$. (b):
high level circuit representation of $\mathit{GSWC}(2x_1 + 3x_2 + 4x_3 + 7x_4 \leq 8, \{\{x_1,x_2\},\{x_3,x_4\}\})$.}
\label{fig:_swc}
\end{figure}

\subsection{Generalized Sequential Weight Counter (GSWC)}\label{ssec:enc_gswc}

We define the GSWC encoding by, instead of associating a single
monomial $q_ix_i$ from the PB constraint to each counter, associating
a set of monomials to each of them. In our generalization, given a
partition ${\cal X}=\{X_1,\dots,X_N\}$ of the variables of the PB
constraint, the resulting formulation will have just $N$ counters,
where the $i$-th counter will handle all the monomials ${q_lx_l}$ for
the variables $x_l$ in $X_i$. If the variables in each set $X_i$ are
subject to an AMO constraint then, given an assignment satisfying
those constraints, at most one coefficient $q_l$ will be added by each
counter, and the output of the whole circuit will correspond to the
value of 
$\sum_{i=1}^{n}q_ix_i$. As in the original encoding, we will enforce
that a sum exceeding $K$ is not reached.  The GSWC encoding introduces
the following clauses:

\begin{align}
 \noo{s_{i-1,j}} \lor s_{i,j} & \qquad 2 \leq i < N,\, 1 \leq j \leq K \label{eq:gswcelse}\\
 \noo{x_l}  \lor s_{i,j} & \qquad 1 \leq i < N,\, x_l \in X_i,\, 1 \leq j \leq q_l\label{eq:gswczerosum}\\
 \noo{s_{i-1,j}} \lor \noo{x_l} \lor s_{i,j+q_l} &\qquad  2 \leq i < N,\,  x_l \in X_i,\, 1 \leq j \leq K - q_l \label{eq:gswcsum}\\
 \noo{s_{i-1,K+1-q_l}} \lor \noo{x_l} & \qquad 2 \leq i \leq N,\, x_l \in X_i \label{eq:gswcfail}
\end{align}

Clauses~\eqref{eq:gswcelse} propagate the accumulated sum in the same
way as Clauses~\eqref{eq:swcelse}.  Clauses \eqref{eq:gswczerosum} and
\eqref{eq:gswcsum} enforce $S_{i} \geq S_{i-1} + q_lx_l$, for all
$x_l \in X_i$, where $S_{i-1}$ and $S_{i}$ are respectively the input
and output value of the $i$-th counter.  Clauses~\eqref{eq:gswcfail}
enforce that the sum never exceeds $K$.  A high level circuit
representation of a GSWC encoding is shown in
Figure~\ref{fig:gswc}.

The main difference between the SWC and GSWC encodings is that the
latter has only $N$ counters, instead of $n$, and therefore introduces
fewer auxiliary variables (assuming $N < n$). Also, the number of
Clauses~\eqref{eq:gswcelse} in the GSWC encoding is smaller than the
number of Clauses~\eqref{eq:swcelse} in the SWC encoding.  The SWC
encoding requires $O(nK)$ auxiliary variables and $O(nK)$ clauses,
while the GSWC encoding requires $O(NK)$ auxiliary variables and
$O(nK)$ clauses.

\subsubsection{Comparison with decision diagrams}
There is a close relationship between sequential weight counter circuits and decision diagrams, i.e.\ between BDD and SWC encodings, and between MDD and GSWC encodings. In particular, a reduced ordered decision diagram can be seen as a sequential weight counter where the $i$-th counter only has output pins representing sum values that can be obtained from a subset of the first $i$ variable coefficients. Moreover, two sum values may share the same decision diagram node if this does not affect satisfiability, and this reduction is not performed in sequential counters. Also, reduced ordered decision diagrams shortcut intermediate nodes when all their output edges point to the same child. This is usually referred to as a long edge, and is also not used by sequential weight counters. Therefore, there are possible sum values that are not represented by any variable in decision diagram encodings. This means that decision diagram encodings are smaller, but in some models it might be useful having a variable representing each sum value. Both the results of the original paper of SWC~\cite{holldobler2012compact} and our results show that sometimes encodings based on sequential counters perform better than those based on decision diagrams.



\section{Generalized Totalizer Encoding}\label{sec:enc_ggt}

In this section we first revisit the  Generalized Totalizer encoding for PB constraints
introduced in~\cite{joshi2015generalized}, then  in Subsection~\ref{ssec:enc_ggt} we provide its generalization to encode PB(AMO) constraints. In Subsection~\ref{sec:enc_rggt} we introduce a new encoding for PB(AMO) constraints, the Reduced Generalized Generalized Totalizer, which is able to merge sets of equivalent values into intervals, potentially reducing the size of the encoding. Subsection~\ref{sec:heur_ggt} provides a heuristic to build the binary trees used in the totalizer encodings.

\subsection{Generalized Totalizer}\label{ssec:enc_gt}

The \emph{Generalized Totalizer} (GT) encoding was
presented in~\cite{joshi2015generalized} as a generalization of the
Totalizer encoding for cardinality constraints~\cite{BailleuxB03}. The
overall idea of GT is to represent a PB constraint
$\sum_{i=1}^{n} q_ix_i \leq K$ as a binary tree where each leaf represents a term of the sum, and non-leaf nodes represents the sum of the terms beneath. In each non-leaf node, every possible value of the sum is represented with one Boolean variable. 

\begin{figure}[!t]
\centering
\subcaptionbox{GT.\label{fig:gt}}{
\footnotesize
\resizebox{12cm}{!}{
\begin{tikzpicture}[auto,node distance= 0.5cm and 0.8cm,>=stealth]
\tikzstyle{tex}=[anchor=south]

\node[tex] (H) []             {$H:h_2$};
\node[tex] (I) [right = 0.3cm of H] {$I:i_3$};
\node[tex] (J) [right = 0.4cm of I] {$J:j_4$};
\node[tex] (K) [right = 0.3cm of J] {$K:k_5$};
\node[tex] (L) [right = 0.4cm of K] {$L:l_3$};
\node[tex] (M) [right = 0.3cm of L] {$M:m_4$};
\node[tex] (N) [right = 0.4cm of M] {$N:n_6$};
\node[tex] (O) [right = 0.3cm of N] {$O:o_8$};

\coordinate (Middle) at ($(H)!0.5!(I)$); \node[tex] (D) [above = of Middle] {$D:d_2,d_3,d_5$};
\coordinate (Middle) at ($(J)!0.5!(K)$); \node[tex] (E) [above = of Middle] {$E:e_4,e_5,e_9$};
\coordinate (Middle) at ($(L)!0.5!(M)$); \node[tex] (F) [above = of Middle] {$F:f_3,f_4,f_7$};
\coordinate (Middle) at ($(N)!0.5!(O)$); \node[tex] (G) [above = of Middle] {$G:g_6,g_8,g_{11}$};

\coordinate (Middle) at ($(D)!0.5!(E)$); \node[tex] (B) [above = of Middle] {$B:b_2,b_3,b_4,b_5,b_6,b_7,b_8,b_9,b_{10},b_{11}$};
\coordinate (Middle) at ($(F)!0.5!(G)$); \node[tex] (C) [above = of Middle] {$C:c_3,c_4,c_6,c_7,c_8,c_9,c_{10},c_{11}$};

\coordinate (Middle) at ($(B)!0.5!(C)$); \node[tex] (A) [above = of Middle] {$A:a_2,a_3,a_4,a_5,a_6,a_7,a_8,a_9,a_{10},a_{11}$};

\path[-] (A) edge (B); \path[-] (A) edge (C);

\path[-] (B) edge (D); \path[-] (B) edge (E);
\path[-] (C) edge (F); \path[-] (C) edge (G);

\path[-] (D) edge (H); \path[-] (D) edge (I);
\path[-] (E) edge (K); \path[-] (E) edge (J);
\path[-] (F) edge (L); \path[-] (F) edge (M);
\path[-] (G) edge (N); \path[-] (G) edge (O);

\node [below right = 0.3cm and -0.45 cm of H,rotate=90] {$=$}; \node [below right = 0.1cm and -0.55 cm of H] {$x_1$};
\node [below right = 0.3cm and -0.45 cm of I,rotate=90] {$=$}; \node [below right = 0.1cm and -0.55 cm of I] {$x_2$};
\node [below right = 0.3cm and -0.45 cm of J,rotate=90] {$=$}; \node [below right = 0.1cm and -0.55 cm of J] {$x_3$};
\node [below right = 0.3cm and -0.45 cm of K,rotate=90] {$=$}; \node [below right = 0.1cm and -0.55 cm of K] {$x_4$};
\node [below right = 0.3cm and -0.45 cm of L,rotate=90] {$=$}; \node [below right = 0.1cm and -0.55 cm of L] {$x_5$};
\node [below right = 0.3cm and -0.45 cm of M,rotate=90] {$=$}; \node [below right = 0.1cm and -0.55 cm of M] {$x_6$};
\node [below right = 0.3cm and -0.45 cm of N,rotate=90] {$=$}; \node [below right = 0.1cm and -0.55 cm of N] {$x_7$};
\node [below right = 0.3cm and -0.45 cm of O,rotate=90] {$=$}; \node [below right = 0.1cm and -0.55 cm of O] {$x_8$};
\end{tikzpicture}
}
}

\vspace{0.3cm}

\subcaptionbox{GGT.\label{fig:ggt}}{
\footnotesize
\resizebox{10cm}{!}{
\begin{tikzpicture}[auto,node distance= 0.5cm and 0.8cm,>=stealth]
\tikzstyle{tex}=[anchor=south]

\coordinate (Middle) at ($(D)!0.5!(E)$); \node[tex] (B) [above = of Middle] {$B:b_2,\quad b_3,\quad b_4, \quad b_5$};
\coordinate (Middle) at ($(F)!0.5!(G)$); \node[tex] (C) [above = of Middle] {$C:c_3,\quad c_4,\quad c_6, \quad c_8$};

\coordinate (Middle) at ($(B)!0.5!(C)$); \node[tex] (A) [above = of Middle] {$A:a_2,a_3,a_4,a_5,a_6,a_7,a_8,a_9,a_{10},a_{11}$};

\path[-] (A) edge (B); \path[-] (A) edge (C);

\node [below right = 0.25cm and -2.70 cm of B,rotate=90] {$=$}; \node [below right = 0.1cm and -2.80 cm of B] {$x_1$};
\node [below right = 0.25cm and -1.90 cm of B,rotate=90] {$=$}; \node [below right = 0.1cm and -2.00 cm of B] {$x_2$};
\node [below right = 0.25cm and -1.20 cm of B,rotate=90] {$=$}; \node [below right = 0.1cm and -1.30 cm of B] {$x_3$};
\node [below right = 0.25cm and -0.45 cm of B,rotate=90] {$=$}; \node [below right = 0.1cm and -0.55 cm of B] {$x_4$};
\node [below right = 0.25cm and -2.70 cm of C,rotate=90] {$=$}; \node [below right = 0.1cm and -2.80 cm of C] {$x_5$};
\node [below right = 0.25cm and -1.90 cm of C,rotate=90] {$=$}; \node [below right = 0.1cm and -2.00 cm of C] {$x_6$};
\node [below right = 0.25cm and -1.20 cm of C,rotate=90] {$=$}; \node [below right = 0.1cm and -1.30 cm of C] {$x_7$};
\node [below right = 0.25cm and -0.45 cm of C,rotate=90] {$=$}; \node [below right = 0.1cm and -0.55 cm of C] {$x_8$};
\end{tikzpicture}
}
}
\caption[]{(a): binary tree of
  $\mathit{GT}(2x_1 + 3x_2 + 4x_3 + 5x_4 + 3x_5 + 4x_6 + 6x_7 + 8x_8
  \leq 10)$.  (b): binary tree of
  $\mathit{GGT}(2x_1 + 3x_2 + 4x_3 + 5x_4 + 3x_5 + 4x_6 + 6x_7 + 8x_8
  \leq 10,\{\{x_1,x_2,x_3,x_4\},\{x_5,x_6,x_7,x_8\}\})$.} 
\label{fig:_gt}
\end{figure}
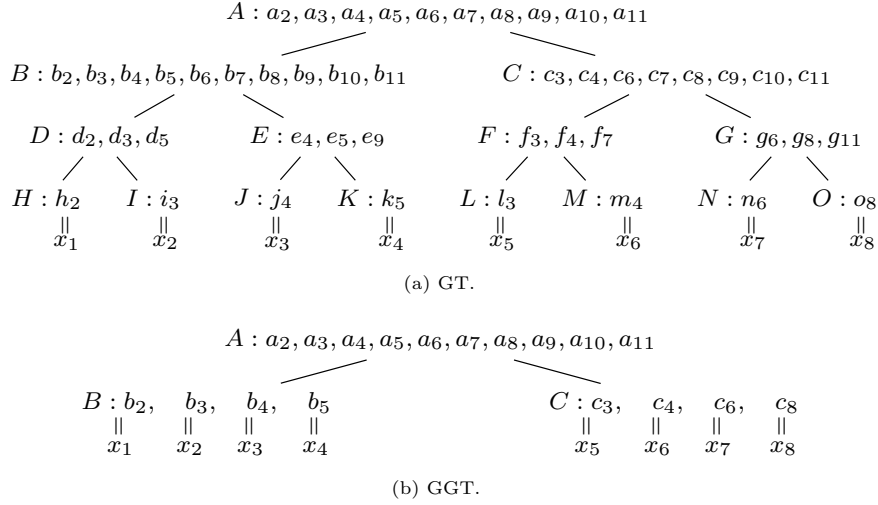

In the original presentation, every node of the
tree has a distinct label and an attribute $\mathit{vars}$ which consists of a
set of Boolean variables. We add another attribute \(\mathit{vals}\),  a set of values. In GT, \(\mathit{vals}\) corresponds to \(\mathit{vars}\) (each non-zero value in \(\mathit{vals}\) has a corresponding variable in \(\mathit{vars}\)), but this is not the case in one of our generalizations of GT.  
Each variable $x_i$ of the PB constraint is
placed into the attribute $\mathit{vars}$ of a different leaf node,
and is renamed after the label of the node and its associated
coefficient $q_i$ (e.g., given the monomial $3x_1$, if the variable
$x_1$ is inserted into a leaf node labelled by letter $O$, then the
variable is named $o_3$). The attribute \(\mathit{vals}\) of the leaf node is simply \(\{0,q_i\}\). The attribute $\mathit{vars}$ of any
non-leaf node labelled $O$ contains a variable $o_{w}$ for every variable $l_w$ of its left child and for every variable $r_w$ of its right child. Moreover, $O.\mathit{vars}$ also contains a variable $o_w$ for  every value $w$ in the range $[1,K]$ resulting from summing any pair of values $w_1$ and  $w_2$, where $l_{w_1}$ and $r_{w_2}$ are variables of the left and right children respectively.
Also, $\mathit{vars}$ contains a variable $o_{K+1}$
iff any of the sums is larger than $K$.
For each variable \(o_c \in O.\mathit{vars}\), value \(c\) is inserted into \(O.\mathit{vals}\), along with value 0.
Figure~\ref{fig:gt} illustrates an example binary tree.

Once the tree is properly constructed, the GT encoding introduces the
following clauses for each non-leaf node $O$ with children $L$ and
$R$:
\begin{align}
\noo{t_w} \lor o_{w}  										& \quad t_w \in
L.\mathit{vars} \cup R.\mathit{vars}\label{eq:gt_sumone}\\
\noo{l_{w_1}} \lor \noo{r_{w_2}} \lor o_{w} 	&  \quad	l_{w_1} \in L.\mathit{vars},\; r_{w_2} \in R.\mathit{vars},\;w=\min(w_1+w_2,K+1)\label{eq:gt_sumtwo}
\end{align}
It also introduces the unary clause:
\begin{align}
 \noo{a_{K+1}}\label{eq:gt_root}
\end{align}
where $A$ is the root node of the tree and
$a_{K+1} \in A.\mathit{vars}$ (otherwise the constraint would be trivially
satisfied). 

Clauses~\eqref{eq:gt_sumone} enforce that the
variable $o_w$ will be set to true by UP if some child has a variable
$t_w$ set to true. 
Clauses~\eqref{eq:gt_sumtwo} enforce that the variable
$o_{w}$ will be set to true by UP if there exists a pair of
variables $l_{w_1},r_{w_2}$ from the children nodes that are set to
true and such that
$w=\min(w_1+w_2,K+1)$. Finally, Clause~\eqref{eq:gt_root} states that the
sum of the tree (i.e., the value of the left hand side expression of
the PB constraint) cannot be greater than~$K$.

We apply a minor optimisation to the GT encoding as well as the GGT encoding presented below.  Variables $a_0,\dots,a_{K}$ of the root node only appear in Clauses~\eqref{eq:gt_sumone} and~\eqref{eq:gt_sumtwo} and they are never negated. Therefore, we do not introduce variables $a_0,\dots,a_{K}$ of the root node nor their associated clauses.

\subsubsection{Comparison with Decomposable Negation Normal Form}\label{sec:rgt_dnnf}
The encodings described in \Cref{sec:enc_bdd} use decision diagrams (ROBDD in particular). Ordered decision diagrams are one target language for knowledge compilation \cite{darwiche2002knowledge}. It is natural to consider whether GT and its generalizations in this section (GGT, RGT, and RGGT) also correspond to a knowledge compilation language. 
The GT encoding and its generalizations (GGT, RGT, and RGGT) can be translated straightforwardly to the knowledge compilation language Decomposable Negation Normal Form (DNNF) \cite{darwiche2002knowledge}. 
For each non-leaf, non-root node \(O\), for each value \(w\) in \(O.\mathit{vals}-\{0\}\), \(w\) can be represented in DNNF as a disjunction with one disjunct for each pair of values \(w_1\) and \(w_2\) (from the left and right children respectively) where \(\min(w_1+w_2, K+1)=w\). Each disjunct would be a conjunction of the two DNNF terms representing \(w_1\) and \(w_2\) respectively (reusing terms where possible). The root node is treated similarly but all values in the range \([1,K]\) are represented by a single disjunction (since the exact value is irrelevant). For a leaf \(q_ix_i\), value \(q_i\) is represented by \(x_i\) and value 0 by \textit{true} (subsequently removed by partial evaluation). This construction is in DNNF but not d-DNNF (it does not have the \textit{determinism} property \cite{darwiche2002knowledge}). DNNF can be encoded into SAT using the Tseitin encoding, and in this case it would be completed by asserting \(\noo{a_{K+1}}\) where \(a_{K+1}\) is the Tseitin variable for the root node value \(K+1\). 
GT has a specialised SAT encoding for the tree which is more compact than the Tseitin encoding of this DNNF formula. 
To adapt this DNNF formula for the RGT and RGGT encodings (described below), values are replaced with intervals.

\subsection{Generalized Generalized Totalizer (GGT)}\label{ssec:enc_ggt}

In our generalization of the GT encoding, we will use the same
definition of the binary tree, but the leaves will be instantiated
differently. Instead of introducing a leaf node for each variable of
the PB constraint, we introduce a leaf node for each of
the sets in the partition $\cal X$.  The leaf node $O$ associated with
set $X_i$ will contain a variable $o_{q_l}$ in its $\mathit{vars}$
attribute for each distinct coefficient $q_l$ such that
$x_l \in X_i$. Note that, due to Property~\ref{prop:difcoeffs}, $q_l \neq q_{l'}$ for
every two distinct variables $x_l,x_{l'} \in X_i$. As in the GT encoding,
every variable $x_l$ is renamed as $o_{q_l}$ and placed in
$O.\mathit{vars}$. Each distinct coefficient \(q_l\) is placed in \(O.\mathit{vals}\), along with value 0. Then, the GGT encoding introduces Clauses
\eqref{eq:gt_sumone}, \eqref{eq:gt_sumtwo} and~\eqref{eq:gt_root} as
in the GT encoding.  Figure~\ref{fig:ggt} depicts the binary tree of a
GGT encoding.

Note that assuming that an AMO constraint over each set
$X_i$ is satisfied, at most one of the variables in each leaf node
will be true, and therefore the encoding correctly evaluates
$\sum_{i=1}^{n} q_ix_i \leq K$.

The GT encoding requires $O(nK)$ auxiliary variables and $O(nK^2)$
clauses, while the GGT encoding requires $O(NK)$ auxiliary variables
and $O(NK^2)$ clauses.  However, as stated
in~\cite{joshi2015generalized}, this size depends on the number of
unique sums of coefficients, and hence the previously given size
bounds are only accurate when the number of unique sums is close to $K$.

\subsection{Reduced Generalized Generalized Totalizer (RGGT)}
\label{sec:enc_rggt}

In the GT and GGT encodings it is possible to have two values \(i,j\) of a tree node \(O\) such that the choice of \(i\) or \(j\) does not affect the semantics of the constraint: informally, it makes no difference whether the node takes value \(i\) or \(j\). We present a new encoding that makes use of this idea, named Reduced Generalized Generalized Totalizer (RGGT). Since PB constraints are a particular case of PB(AMO) constraints, this encoding is also useful to encode PB constraints. We refer to the PB encoding as Reduced Generalized Totalizer (RGT), and given an input PB constraint $\sum_{i=1}^{n} q_i x_i \leq K$, RGT consists of the RGGT encoding with partition ${\cal X} = \{\{x_1\},\dots,\{x_n\}\}$.

The key idea of the RGGT is to merge sets of equivalent values into intervals and use only one variable per interval in the encoding. The RGGT was inspired by the ROBDD encoding of pseudo-Boolean constraints~\cite{abio2012new}, which never contains two equivalent states by construction. In RGGT, tree nodes have an attribute \(\mathit{intervals}\) which is a sequence of integer intervals. Within each interval, the choice of value does not change the semantics of the constraint. 

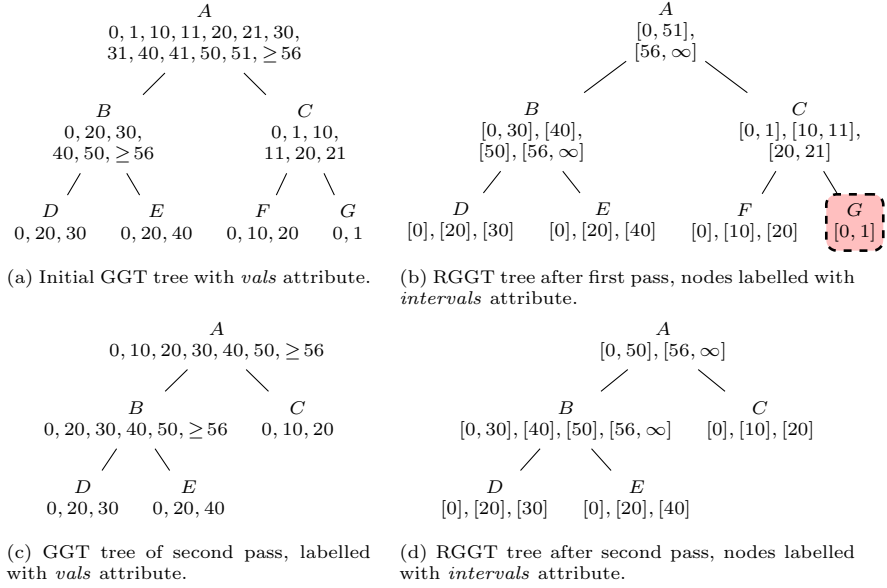
\begin{figure}
  \centering
  \scriptsize
  \makebox[\width][c]{
    \subcaptionbox{Initial GGT tree with \emph{vals} attribute.\label{fig:rggt_a}}[0.4\textwidth]{
      \begin{forest}
        for tree = {s sep=7pt, align=center}
        [{$A$\\ $0,1,10,11,20,21,30,$\\$31,40,41,50,51,\geq\!56$}
          [{$B$\\ $0,20,30,$\\$40,50,\geq\!56$}
            [{$D$\\ $0,20,30$}] [{$E$\\$0,20,40$}]
          ]
          [{$C$\\ $0,1,10,$\\$11,20,21$}
            [{$F$\\ $0,10,20$}] [{$G$\\ $0,1$}]
          ]
        ]
      \end{forest}
    }
    \quad
    \subcaptionbox{RGGT tree after first pass, nodes labelled with  \emph{intervals} attribute.\label{fig:rggt_b}}[0.5\textwidth]{
      \begin{forest}
        for tree = {s sep=7pt,align=center}
        [{$A$\\$[0,51],$\\$[56,\infty]$}
          [{$B$\\$[0,30],[40],$\\$[50],[56,\infty]$}
            [{$D$\\$[0],[20],[30]$}]
            [{$E$\\$[0],[20],[40]$}]
          ]
          [{$C$\\$[0,1], [10,11],$\\$ [20,21]$}
            [{$F$\\$[0],[10],[20]$}]
            [{$G$\\$[0,1]$},fill=pink,dashed,line width=1pt,rectangle,rounded corners=4pt,draw]
          ]
        ]
      \end{forest}
    }
  }
  \\
  \subcaptionbox{GGT tree of second pass, labelled with \emph{vals} attribute.\label{fig:rggt_c}}[0.4\textwidth]{
    \begin{forest}
      for tree = {s sep=7pt,align=center}
      [{$A$\\ $0,10,20,30,40,50,\geq\!56$}
        [{$B$\\ $0,20,30,40,50,\geq\!56$}
          [{$D$\\ $0,20,30$}] [{$E$\\ $0,20,40$}]
        ]
        [{$C$\\ $0,10,20$}]
      ]
    \end{forest}
  }
  \quad
  \subcaptionbox{RGGT tree after second pass, nodes labelled with  \emph{intervals} attribute.\label{fig:rggt_d}}[0.5\textwidth]{
    \begin{forest}
      for tree = {s sep=7pt,align=center}
      [{$A$\\$[0,50],[56,\infty]$}
        [{$B$\\$[0,30], [40],[50], [56,\infty]$}
          [{$D$\\$[0], [20], [30]$}]
          [{$E$\\$[0], [20], [40]$}]
        ]
        [{$C$\\$[0], [10], [20]$}]
      ]
    \end{forest}
  }
  \caption[]{Two passes of the RGGT algorithm reach a stable tree for the PB(AMO) $20x_1+30x_2+20x_3+40x_4+10x_5+20x_6+x_7 \leq 55,\mathcal{X}=\{\{x_1,x_2\},\{x_3,x_4\},\{x_5,x_6\},\{x_7\}\}.$}
  \label{fig:rggt}
\end{figure}

Take for example the constraint $20x_1 + 30x_2 + 20x_3 +
40x_4 + 10x_5 + 20x_6 + x_7 \leq 55$ with AMO partition  \(\mathcal{X}=\{\{x_1,x_2\},\{x_3,x_4\},\{x_5,x_6\},\{x_7\}\}\). \Cref{fig:rggt_a} shows a GGT for this constraint (in this case the \(\mathit{vals}\) attribute is shown in each node). First, all values of the root node (A) that are \(\leq 55\) are equivalent: they all unconditionally satisfy the constraint. Therefore the root node has two intervals: \([0,51],[56,\infty]\).
Next, in node B, values 0, 20, and 30 are equivalent: for all values \(c\in C.\mathit{vals}\), the sums \(0+c\), \(20+c\), and \(30+c\) are in the same interval of \(A\) therefore the choice of 0, 20, or 30 is not significant regardless of the value of \(C\). Collecting values into intervals is called \textit{reduction} and parents are always reduced before children. An interval is non-trivial if it contains more than one value. 

\Cref{fig:rggt_b} shows the RGGT tree created by one pass of reduction. Nodes A, B, C, and G have non-trivial intervals. Node G is a leaf node, representing term \(x_7\). Non-trivial intervals in leaf nodes trigger changes to the PB(AMO) constraint, the details of these changes will be described later. In this case, term \(x_7\) and its AMO group are deleted because the term does not affect the satisfaction of the constraint (values 0 and 1 are in the same interval). Whenever the PB constraint is modified the entire RGGT process is repeated from scratch. \Cref{fig:rggt_c} shows the GGT tree constructed in the second pass, with 2 fewer nodes than \Cref{fig:rggt_a}. Finally, \Cref{fig:rggt_d} shows the RGGT tree produced by the second pass. The final RGGT tree would require 6 SAT variables to encode, whereas the original GGT in \Cref{fig:rggt_a} would require 11.

\begin{algorithm}[ht!]
\SetAlgoLined
\DontPrintSemicolon

\SetKwProg{Proc}{Procedure}{is}{end}
\SetKwFunction{ProcMCI}{makeChildIntervals}
\SetKw{KwBreak}{break}

\KwData{A PB(AMO) constraint $ \sum_{i=1}^{n} q_i x_i \leq K$, with variable partition \(\mathcal{X}\)}
\KwResult{A Reduced Generalized Generalized Totalizer tree}

\BlankLine
\Repeat{No changes to PB(AMO) constraint}{
Build a GGT tree with root node \(A\) without \(\mathit{vars}\) attributes\;
\(A.\mathit{intervals}\leftarrow [[0,\mathrm{max}(A.\mathit{vals}\setminus\{K+1\})], [K+1,\infty]]\)\;
\ProcMCI{A}\;
\ForEach{\(L\): leaf node with non-trivial interval}{
    \ForEach{\([l,u]\): non-trivial interval (\(u>l\)) in \(L.\mathit{intervals}\)}{
        \ForEach{\(q_ix_i\) linked to \(L\) where \(l< q_i \leq u\)}{\(q_i\leftarrow l\)} 
    }
}
Delete terms \(q_ix_i\) where \(q_i=0\)\; 
Delete empty cells of partition \(\mathcal{X}\)\;
}

\BlankLine

\Proc{\ProcMCI{$U$}}{
    \ForEach{child $V$ (with sibling \(W\)) of $U$ }{
        \(V.\mathit{intervals}\leftarrow [ [a,a] \mid a\in V.\mathit{vals}, a<K+1]+[ [a,\infty] \mid a\in V.\mathit{vals}, a=K+1 ]\)\;
        \ForEach{adjacent pair of intervals \([a,b], [c,d]\) in \(V.\mathit{intervals}\) in ascending order}{
            \If{\(\forall w\in W.\mathit{vals}.\enspace\exists I\in U.\mathit{intervals}.\enspace(w+b)\in I \wedge (w+c)\in I\)}{
                Replace \([a,b], [c,d]\) with \([a,d]\) in \(V.\mathit{intervals}\)
            }
        }
        \If{$V$ has children}{
            \ProcMCI{$V$}
        }
    }
}

\caption{RGGT reduction algorithm.} \label{alg:rggt-reduce}
\end{algorithm}

The reduction algorithm is presented in Algorithm~\ref{alg:rggt-reduce}. The first section shows the outer loop, the body of which builds a GGT tree with only \(\mathit{vals}\) attributes. No additional Boolean variables are created at this point. Then \texttt{makeChildIntervals} is called to create the \(\mathit{intervals}\) attribute of each node. After all intervals have been constructed, for each non-trivial interval of a leaf node, corresponding terms of the PB constraint are adjusted. First, coefficients are reduced to the smallest value in the interval (lines 5--8). Any value in the interval could be used; the smallest is a heuristic choice. Terms with coefficient 0 are removed, then empty AMO sets are removed from \(\mathcal{X}\) (on lines 9--10), thus restoring Property~\ref{prop:qgt0}.

The function \texttt{makeChildIntervals} traverses the GGT tree top-down, creating the \(\mathit{intervals}\) attribute of each node \(V\) using \(\mathit{intervals}\) of the parent \(U\) and \(\mathit{vals}\) of the sibling \(W\). Two adjacent values \(a,b\) of \(V\) are placed in the same interval iff (for each value \(w\) of the sibling node) \(a+w\) and \(b+w\) are in the same interval of the parent -- i.e.\ the choice of \(a\) or \(b\) is not significant.

\begin{property}\label{lem:rggt-incl}
For each non-leaf node \(O\) with children \(L\) and \(R\), for each pair of intervals \([a,b]\in L.\mathit{intervals}\) and \([c,d]\in R.\mathit{intervals}\) there must exist an interval \([e,f]\in O.\mathit{intervals}\) that contains \([a+c,b+d]\). 
\end{property}

Note that Property~\ref{lem:rggt-incl} will hold by construction, otherwise either \([a,b]\) or \([c,d]\) could not have been constructed by lines 16--17 of Algorithm~\ref{alg:rggt-reduce}.

Finally the RGGT tree is encoded into SAT. For each non-leaf node \(U\), for each interval \([a,b]\in U.\mathit{intervals}\) where \(a>0\), one Boolean variable \(u_{a,b}\) is created and placed in \(U.\mathit{vars}\). For each leaf node \(L\), note that all intervals must be trivial (since Algorithm~\ref{alg:rggt-reduce} has completed) but some intervals may be linked to more than one term. For each interval \([a,a]\in L.\mathit{intervals}\) where \(a>0\), if exactly one term \(q_ix_i\) is linked to \(L\) and \(q_i=a\), then \(x_i\) is named \(l_{a,a}\) and added to \(L.\mathit{vars}\). Otherwise a new Boolean variable \(l_{a,a}\) is created, and for all terms \(q_ix_i\) linked to \(L\) where \(q_i=a\) the clause \(\neg x_i \vee l_{a,a}\) is added. 

The RGGT encoding introduces the following clauses for each non-leaf node $O$ with children $L$ and $R$. The following two sets of clauses are comparable to Clauses \ref{eq:gt_sumone} and \ref{eq:gt_sumtwo} of (G)GT.
\begin{align}
\noo{t_{a,b}} \lor o_{e,f}  & 
\quad t_{a,b} \in L.\mathit{vars} \cup R.\mathit{vars},\; o_{e,f}\in O.\mathit{vars},\; e\leq a,\; b\leq f \label{eq:rggt_sumone}\\
\noo{l_{a,b}} \lor \noo{r_{c,d}} \lor o_{e,f} 	&  \quad	l_{a,b} \in L.\mathit{vars},\; r_{c,d} \in R.\mathit{vars},\; o_{e,f}\in O.\mathit{vars},\; e\leq a+c,\; b+d\leq f\label{eq:rggt_sumtwo}
\end{align}
For the root node A the unary clause \(\noo{a_{K+1,\infty}}\) is added.

\subsection{A Heuristic To Build GT, GGT, RGT, and RGGT Trees} \label{sec:heur_ggt}

Each of the totalizer encodings described above (GT, GGT, RGT, and RGGT) requires a method to construct a binary tree given a set of leaf nodes. 
In this section we propose a new heuristic named minRatio to build the binary tree required by these encodings. 
The minRatio heuristic compares the number of values of a (proposed) internal node to the product of the numbers of values of its two children, greedily minimising the ratio of these two quantities. Suppose we have three tree nodes \(C,D,E\)  with value sets \(C.\mathit{vals}=\{0,3,6\}\), \(D.\mathit{vals}=\{0,3,6\}\), and \(E.\mathit{vals}=\{0,4,8\}\), and \(K=15\). If \(C\) and \(E\) shared a parent \(A\), then \(A.\mathit{vals}=\{0,3,4,6,7,8,10,11,14\}\). The size of \(A.\mathit{vals}\) is the product of \(|C.\mathit{vals}|\) and \(|E.\mathit{vals}|\).  However, if \(C\) and \(D\) shared a parent \(B\), then \(B.\mathit{vals}=\{0,3,6,9,12\}\) -- smaller than the product of its children. The minRatio heuristic would generate \(B\) in this case.

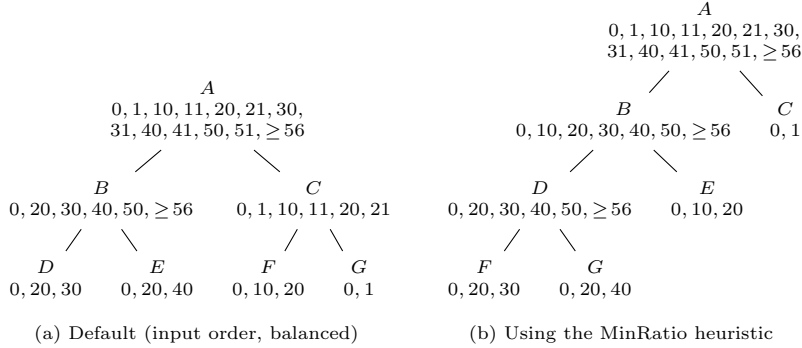
\begin{figure}
  \centering
  \scriptsize
  \makebox[\width][c]{
  \subcaptionbox{Default (input order, balanced)}{
    \begin{forest}
      for tree = {s sep=9pt,align=center}
      [{$A$ \\ $0,1,10,11,20,21,30,$\\$31,40,41,50,51,\geq\!56$ }
        [{$B$ \\ $0,20,30,40,50,\geq\!56$}
          [{$D$\\$0,20,30$}]
          [{$E$\\$0,20,40$}]
        ]
        [{$C$ \\ $0,1,10,11,20,21$}
          [{$F$\\$0,10,20$}]
          [{$G$\\$0,1$}]
        ]
      ]
    \end{forest}
  }
  \quad
  \subcaptionbox{Using the MinRatio heuristic}{
    \begin{forest}
      for tree = {s sep=9pt,align=center}
      [{$A$ \\ $0,1,10,11,20,21,30,$ \\ $31,40,41,50,51,\geq\!56$}
        [{$B$ \\ $0,10,20,30,40,50,\geq\!56$}
          [{$D$ \\ $0,20,30,40,50,\geq\!56$}
            [{$F$ \\ $0,20,30$}]
            [{$G$ \\ $0,20,40$}]
          ]
          [{$E$ \\ $0,10,20$}]
        ]
        [{$C$ \\ $0,1$}]
      ]
    \end{forest}
  }    
}
  \caption[]{Comparing tree-building heuristics for
    $\mathit{GGT}(20x_1+30x_2+20x_3+40x_4+10x_5+20x_6+x_7 \leq 55,\{\{x_1,x_2\},\{x_3,x_4\},\{x_5,x_6\},\{x_7\}\})$. 
    }
  \label{fig:minratio-trees}
\end{figure}

MinRatio works on a set \(S\) of tree nodes. Initially \(S\) contains all leaf nodes. MinRatio has completed when \(S\) contains a single node (which will be the root). At each step, two nodes \(B\) and \(C\) are removed from \(S\) and their parent \(A\) is created and added to \(S\). \(B\) and \(C\) are selected to minimise the quotient \(|A.\mathit{vals}|/(|B.\mathit{vals}|\times|C.\mathit{vals}|)\). Note that \(K+1\) is counted as a single value. 
In the experiments described below, we refer to the encodings using the minRatio heuristic simply as GT, GGT, RGT, and RGGT. 

In order to evaluate the impact of the minRatio heuristic, in Section~\ref{sec:results} we will compare its performance to that of a natural default heuristic with the GT and GGT encodings. The default heuristic constructs a balanced binary tree where the elements of ${\cal X}$ are placed in the leaf nodes from left to right in input order. When the number of leaf nodes is not a power of two, the leftmost leaves are the ones of higher depth. In the experimental section we refer to the encodings using these heuristics as GTd and GGTd (where d indicates \emph{default}). Our results show that the minRatio heuristic has a substantial positive effect on the size and performance of GT and GGT encodings when compared to GTd and GGTd.  

The default heuristic produces balanced trees of height \(\lceil \mathrm{log}_2(|\mathcal{X}|) \rceil\), whereas minRatio can produce unbalanced trees of much greater height. In fact we observed that minRatio tends to produce highly unbalanced trees. In our experiments with the Multi-Choice Multidimensional Knapsack Problem (MMKP) (described in \Cref{sec:res_setting}), when using GGT the number of leaf nodes is \(|\mathcal{X}|=15\) and the tree height is 14 in all cases except for one constraint of one instance in MMKP3 which has height 13. With just one exception, the trees are as high as possible (i.e.\ maximally unbalanced). When using GT, considering MMKP1, the trees have height at least 112 (with 150 leaf nodes). In MMKP2, trees have height at least 97 (with 150 leaves), and in MMKP3, trees have height 57 or more (with 75 leaf nodes). Minimal heights are 8 for MMKP1 and MMKP2, and 7 for MMKP3, so it is clear that minRatio is building extremely unbalanced trees. For the Combinatorial Auctions problem (see \Cref{sec:res_setting}) PB(AMO) constraints are of different sizes so we report percentages. All GGT trees are a minimum of 74\% of the maximum possible height, and GT trees are at least 38\% of the maximum possible height --- in both cases substantially higher than a balanced tree. 
\Cref{fig:minratio-trees} compares the default heuristic to minRatio on a small example with the GGT encoding. In this case, minRatio produces an unbalanced tree of maximal height by first combining the three leaf nodes with larger values (i.e. values that are multiples of 10) in one subtree. The combination of node C with any other node has a poor ratio, so node C is the last leaf node to be incorporated into the tree. 

\subsubsection{Comparison with decision diagrams}
Given that minRatio produces extremely unbalanced trees, it is natural to
compare the RGT and RGGT encodings with minRatio to the BDD and MDD encodings in  \Cref{sec:enc_bdd}. Suppose that the tree for RGT/RGGT is of maximal height, and that the order of terms or AMO groups in the tree (from the deepest leaves to the root) is consistent with the order of the decision diagram. The nodes \(\top\) and \(\bot\) of the decision diagram correspond to the two intervals of the root node of the RGT/RGGT. The two encodings collapse equivalent states in very similar ways. The most significant difference is that the decision diagram may have long edges (i.e.\ edges that bypass one or more terms or AMO groups) whereas this is not possible in the RGT/RGGT encoding.


\section{n-Level Modulo Totalizer Encoding}\label{sec:enc_gmto}

In this section we first revisit the n-Level Modulo Totalizer encoding for PB constraints from~\cite{zha2019n} and in Subsection~\ref{ssec:enc_gmto} we provide its generalization to encode PB(AMO) constraints. Subsection~\ref{ssec:const_gmto} gives some construction details.

\subsection{n-Level Modulo Totalizer}
The \emph{n-Level Modulo Totalizer} (MTO)
encoding generalises the Weighted 
Totalizer encoding which, as stated in~\cite{zha2019n}, is essentially
an equivalent definition to the GT encoding that was presented in a
parallel work. For the sake of readability we unify the nomenclature
of the MTO encoding with the one of the GT encoding, but the overall idea of the encoding summarised here and the resulting clauses are the same as in the original work.

The MTO encoding, equally to the GT encoding, consists in building a
(generalization of a) totalizer in which the root of every subtree
represents the sum of the monomials associated to its leaf nodes and,
in particular, the root of the whole totalizer represents the sum of
the monomials in the PB constraint. The main difference with respect
to GT is that the values of the nodes are represented in a mixed radix
base.

A \emph{mixed radix base}  is a vector
$\Lambda = \langle \lambda_0,\dots,\lambda_{\beta-1}\rangle$ where
$\beta\in\mathbb N_0$ and
$\forall i\in 0..\beta-1:\lambda_i\in\mathbb N, \lambda_i > 1$. A
number $I\in\mathbb N_0$ is represented in base $\Lambda$ as
$d_\beta\;d_{\beta-1}\;\dots\;d_0$ where
$\forall i\in 0..\beta-1: 0\leq d_i<\lambda_i$,
$d_\beta\in{\mathbb N}_0$, and
$I =
d_\beta\times(\lambda_0\times\lambda_1\times\dots\times\lambda_{\beta-1})
+ \dots + d_2 \times (\lambda_0\times\lambda_1) + d_1 \times
(\lambda_0) + d_0$.

For instance, numbers $0,1,\dots,10$ in base $\langle 3,2\rangle$
would be represented as 0\;0\;0, 0\;0\;1, 0\;0\;2, 0\;1\;0, 0\;1\;1,
0\;1\;2, 1\;0\;0, 1\;0\;1, 1\;0\;2, 1\;1\;0, 1\;1\;1. Number $100$
would be represented in base $\langle 3,2\rangle$ as 16\;1\;1, that is
$16\times(2\times3) + 1\times3 + 1$, and it would be represented in
base $\langle 3,2,2\rangle$ as 8\;0\;1\;1, that is
$8\times(2\times2\times3) + 0\times(2\times3) + 1\times3 + 1$.  As a
particular case, a binary base can be defined as a vector containing
an infinite number of $2$s, that we denote by
$\Lambda = \langle 2* \rangle$, and similarly a decimal base can be
defined as $\Lambda = \langle 10* \rangle$.

\begin{figure}[ht!]
	\centering
	\subcaptionbox{MTO.\label{fig:mto}}{
	\footnotesize
	\resizebox{12cm}{!}{
		\begin{tikzpicture}[auto,node distance= 1cm and1cm,>=stealth]
		\tikzstyle{tex}=[rectangle,draw,anchor=south, align=left]

		\node[tex] (H) []             {$H_2:$\\$H_1:$\\$H_0:h^0_2$};
		\node[tex] (I) [right = 0.3cm of H] {$I_2:$\\$I_1:$\\$I_0:i^0_3$};
		\node[tex] (J) [right = 0.4cm of I] {$J_2:$\\$J_1:j^1_1$\\$J_0:$};
		\node[tex] (K) [right = 0.3cm of J] {$K_2:$\\$K_1:k^1_1$\\$K_0:k^0_1$};
		\node[tex] (L) [right = 0.4cm of K] {$L_2:$\\$L_1:$\\$L_0:l^0_3$};
		\node[tex] (M) [right = 0.3cm of L] {$M_2:$\\$M_1:m^1_1$\\$M_0:$};
		\node[tex] (N) [right = 0.4cm of M] {$N_2:$\\$N_1:n^1_1$\\$N_0:n^0_2$};
		\node[tex] (O) [right = 0.3cm of N] {$O_2:$\\$O_1:o^1_2$\\$O_0:$};

		\coordinate (Middle) at ($(H)!0.5!(I)$); \node[tex] (D) [above = of Middle] {$D_2:$\\$D_1:d^1_1$\\$D_0:d^0_1,d^0_2,d^0_3, \gamma^0_D$};

		\coordinate (Middle) at ($(J)!0.5!(K)$); \node[tex] (E) [above = of Middle] {$E_2:$\\$E_1:e^1_1,e^1_2$\\$E_0:e^0_1$};

		\coordinate (Middle) at ($(L)!0.5!(M)$); \node[tex] (F) [above = of Middle] {$F_2:$\\$F_1:f^1_1$\\$F_0:f^0_3$};

		\coordinate (Middle) at ($(N)!0.5!(O)$); \node[tex] (G) [above = of Middle] {$G_2:g^2_1$\\$G_1:g^1_1,g^1_2,\gamma^1_G$\\$G_0:g^0_2$};

		\coordinate (Middle) at ($(D)!0.5!(E)$); \node[tex] (B) [above = of Middle] {$B_2:b^2_1$\\$B_1:b^1_1,b^1_2,\gamma^1_B$\\$B_0:b^0_1,b^0_2,b^0_3,\gamma^0_B$};

		\coordinate (Middle) at ($(F)!0.5!(G)$); \node[tex] (C) [above = of Middle] {$C_2:c^2_1,c^2_2$\\$C_1:c^1_1,c^1_2,\gamma^1_C$\\$C_0:c^0_1,c^0_2,c^0_3,\gamma^0_C$};

		\coordinate (Middle) at ($(B)!0.5!(C)$); \node[tex] (A) [above = of Middle] {$A_2:a^2_1,a^2_2,a^2_3,a^2_4$\\$A_1:a^1_1,a^1_2,\gamma^1_A$\\$A_0:a^0_1,a^0_2,a^0_3,\gamma^0_A$};

		\path[-] (A) edge (B); \path[-] (A) edge (C);

		\path[-] (B) edge (D); \path[-] (B) edge (E);
		\path[-] (C) edge (F); \path[-] (C) edge (G);

		\path[-] (D) edge (H); \path[-] (D) edge (I);
		\path[-] (E) edge (K); \path[-] (E) edge (J);
		\path[-] (F) edge (L); \path[-] (F) edge (M);
		\path[-] (G) edge (N); \path[-] (G) edge (O);

		\node [below right = 0.1cm and -1.5 cm of H] {$h^0_2{=}x_1$};
		\node [below right = 0.1cm and -1.3 cm of I] {$i^0_3{=}x_2$};
		\node [below right = 0.1cm and -1.4 cm of J] {$j^1_1{=}x_3$};
		\node [align=left] [below right = 0.1cm and -1.5 cm of K] {$k^0_1=k^1_1$\\ ${=}x_4$};
		\node [below right = 0.1cm and -1.3 cm of L] {$l^0_3{=}x_5$};
		\node [below right = 0.1cm and -1.6 cm of M] {$m^1_1{=}x_6$};
		\node [align=left] [below right = 0.1cm and -1.5 cm of N] {$n^0_2{=}n^1_1$\\ ${=}x_7$};
		\node [below right = 0.1cm and -1.5 cm of O] {$o^1_2{=}x_8$};
		\end{tikzpicture}
	}
}

	\vspace{0.5cm}
\makebox[\width][c]{
\subcaptionbox{Coefficients in base $\langle 4,3\rangle$.\label{fig:mtobase}}{
	\footnotesize
		
			\begin{tabular}{|r||rrr|} \hline
			2 &0 &0 &2 \\ \hline
			3 &0 &0 &3 \\ \hline
			4 &0 &1 &0 \\ \hline
			5 &0 &1 &1 \\ \hline
			6 &0 &1 &2 \\ \hline
			8 &0 &2 &0 \\ \hline
			\end{tabular}
}
\hspace{1cm}	
\subcaptionbox{GMTO.\label{fig:gmto}}{
	\footnotesize
	\resizebox{8cm}{!}{
	\begin{tikzpicture}[auto,node distance= 1cm and1cm,>=stealth]
		\tikzstyle{tex}=[rectangle,draw,anchor=south, align=left]

	    \node[tex] (B) [] {$B_2:$\\$B_1:b^1_1$\\$B_0:b^0_1,b^0_2,b^0_3$};
		\node[tex] (C) [right = 1cm of B] {$C_2:$\\$C_1:c^1_1,c^1_2$\\$C_0:c^0_2,c^0_3$};

		\node[align=left,anchor=north east] (Breif) [left = 0.1cm  of B]
		{$b^0_1=x_4$\\
		$b^0_2=x_1$\\
		$b^0_3=x_2$\\
		$b^1_1 \leftarrow x_3\lor x_4$};

		\node[align=left,anchor=north west] (Creif) [right = 0.1cm  of C]
		{$c^0_2=x_7$\\
		$c^0_3=x_5$\\
		$c^1_1 \leftarrow x_6 \lor x_7$\\
		$c^1_2=x_8$};

		\coordinate (Middle) at ($(B)!0.5!(C)$); \node[tex] (A) [above = of Middle] {$A_2:a^2_1$\\$A_1:a^1_1,a^1_2,\gamma^1_A$\\$A_0:a^0_1,a^0_2,a^0_3,\gamma^0_A$};

		\path[-] (A) edge (B); \path[-] (A) edge (C);

		\end{tikzpicture}
	}
}
}
	\caption[]{Example with base  $\Lambda=\langle 4,3 \rangle$, $P: 2x_1 + 3x_2 + 4x_3 + 5x_4 + 3x_5 + 4x_6 + 6x_7 + 8x_8
		\leq 10$.
		(a): binary tree of
		$\mathit{MTO}(P)$.  (b): representation in base $\Lambda$ of the coefficients of $P$. (c): binary tree of
		$\mathit{GMTO}(P,\{\{x_1,x_2,x_3,x_4\},\{x_5,x_6,x_7,x_8\}\})$.}
	\label{fig:_mto}
\end{figure}
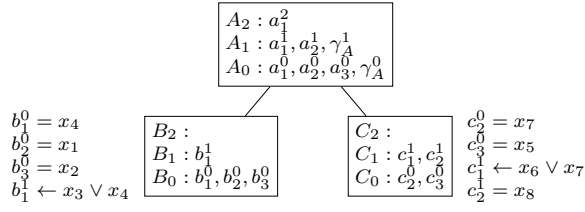

The MTO encoding builds a totalizer where the value of a node is
represented in a mixed radix base, and each digit is represented with a distinct symbol as in the GT encoding.  An example is
given in Figure~\ref{fig:mto}.  For each node $O$ and digit $d_h$,
with $h \in 0..\beta-1$, we define a list of variables $O_h$ of
maximum length $\lambda_{h}$, which will contain a subset of the
variables $o^h_0,\dots,o^h_{\lambda_{h}-1}$. We also define a list
$O_\beta$ (i.e., the list for the digit of most weight), that can
contain as many variables as required to represent the value of the
node. If a variable $o^h_{i}$ is true, it means that the $d_h$ digit
of the value of $O$ is at least $i$. The element $o^h_0$ is always
present in a list $O_h$, and is defined to be a 1 constant (it is
omitted in Figure~\ref{fig:_mto}). The other variables only appear if
required to represent the node values, similarly to GT, as follows:

\begin{itemize}
\item At leaf nodes we place the variables of the PB constraint $P$
  transformed into variables of $O_h$ lists. Note that each variable
  in $P$ can correspond to more than one variable in the $O_h$
  lists. For instance, in Figure~\ref{fig:mto}, variable $x_4$ has
  coefficient 5, which is represented in base $\langle 4,3\rangle$ as
  1\;1, and therefore corresponds to the variables $k^0_1$ and
  $k^1_1$ in the lists $K_0$ and $K_1$ of node $K$.
  
\item At a non-leaf node $O$ with left and right children $L$ and $R$, we
  add to list $O_h$ the required new variables to represent any sum
  of the variables in $L_h$ and $R_h$.  First of all, we introduce a new variable to represent the carry digit whenever the sum of two
  weighted digits from $L_h$ and $R_h$ is greater or equal than
  $\lambda_h$. We name this variable $\gamma^h_O$, and it is only
  introduced if there exist two variables $l^h_i$, $r^h_j$ such that
  $i+j\geq\lambda_h$, or such that $i+j+1\geq\lambda_h$ if there
  exists the carry digit $\gamma^{h-1}_O$. Otherwise, the carry
  variable $\gamma^h_O$ is assumed to be false.  Then, $O_{h}$
  contains a variable $o^h_\sigma$, with
  $\sigma=i+j(\mathit{mod}\ \lambda_h)$, for any two variables
  $l^h_i$, $r^h_j$. Also, if $\gamma^{h-1}_O$ exists, $O_h$ contains
  variables $o^h_\sigma$, with
  $\sigma=i+j+1(\mathit{mod}\ \lambda_h)$.
\end{itemize}
Note that the GT encoding can be seen as a particular case of the MTO encoding where an empty base $\Lambda=\langle\rangle$ is used.

The encoding contains, firstly, the clauses needed to
propagate the sums of values from the leaf nodes to the root
node. For every non-leaf node $O$, with children $L$ and $R$,  we add
the following clauses.\footnote{The nomenclature of this formulation
  has been slightly changed w.r.t. the one in~\cite{zha2019n} without
  affecting the final result, for the sake of a unified notation in
  the current work.}

\noindent  Sums when the carry-in does not exist.  For all $h \in 0..\beta-1$, $l^h_i \in L_h$, $r^h_j \in R_h$, where $\sigma=i+j$:
\begin{align}
&\noo{l^h_i} \lor \noo{r^h_j} \lor o^h_\sigma \lor \gamma^h_O & \mathit{if}\ \sigma<\lambda_h\label{eq:mto_sumnocarry_1}\\
&\noo{l^h_i} \lor \noo{r^h_j} \lor \gamma^h_O & \mathit{if}\ \sigma\geq\lambda_h\label{eq:mto_sumnocarry_2}\\
&\noo{l^h_i} \lor \noo{r^h_j} \lor o^h_{\sigma\ \mathit{mod}\ \lambda_h}
 & \mathit{if}\ \sigma>\lambda_h\label{eq:mto_sumnocarry_3}
\end{align}

\noindent  Sums when the carry-in does exist.  For all $h \in 1..\beta-1$, $l^h_i \in L_h$, $r^h_j \in R_h$, where $\sigma=i+j+1$:
\begin{align}
&\noo{\gamma^{h-1}_O} \lor \noo{l^h_i} \lor \noo{r^h_j} \lor o^h_\sigma  \lor \gamma^h_O & \mathit{if}\ \sigma< \lambda_h\label{eq:mto_sumcarry_1}\\
&\noo{\gamma^{h-1}_O} \lor \noo{l^h_i} \lor \noo{r^h_j} \lor \gamma^h_O & \mathit{if}\ \sigma\geq\lambda_h\label{eq:mto_sumcarry_2}\\
&\noo{\gamma^{h-1}_O} \lor \noo{l^h_i} \lor \noo{r^h_j} \lor o^h_{\sigma\ \mathit{mod}\ \lambda_h}
& \mathit{if}\ \sigma>\lambda_h\label{eq:mto_sumcarry_3}
\end{align}

\noindent Sums for the uppermost digits. For all $l^\beta_i \in L_\beta$, $r^\beta_j \in R_\beta$, where $\sigma=i+j$:
\begin{align}
&\noo{l^\beta_i} \lor \noo{r^\beta_j} \lor o^\beta_\sigma  & \label{eq:mto_sumupper_1}\\
&\noo{\gamma^{h-1}_O} \lor \noo{l^\beta_i} \lor \noo{r^\beta_j} \lor o^\beta_{\sigma+1}
& \label{eq:mto_sumupper_2}
\end{align}

Secondly, the encoding enforces that the value of the root node is not greater than the constant $K$ of the PB constraint. To impose this constraint, $K$ must also be represented in base $\Lambda$, and we refer to the $d_h$ digit of this representation by $K^h$. The clauses to be added are the following, specified from the uppermost to the lowest digit, where $O$ is the root node of the tree:
\begin{align}
&\noo{o^\beta_i}  && \forall o^\beta_i \in O_\beta, i > K^\beta \\
&\noo{o^\beta_{K^\beta}} \lor \noo{o^{\beta-1}_i}  && \forall o^{\beta-1}_i \in O_{\beta-1}, i > K^{\beta-1} \\
&\vdots&& \nonumber\\
&\noo{o^\beta_{K^\beta}} \lor \noo{o^{\beta-1}_{K^{\beta-1}}} \lor \dots \lor \noo{o^2_{K^2}} \lor \noo{o^1_i} && \forall o^1_i \in O_1, i > K^1 \\
&\noo{o^\beta_{K^\beta}} \lor \noo{o^{\beta-1}_{K^{\beta-1}}} \lor \dots \lor \noo{o^1_{K^1}} \lor \noo{o^0_i} && \forall o^0_i \in O_0, i > K^0 \label{eq:mto_last_k}
\end{align}

Note that this series of formulas can be stopped if we find an index $h$ such that $o^h_{K^h}\not\in O_h$, because $\noo{o^h_{K^h}}$ is always true and is part of all the clauses from that point on.

\subsubsection{Comparison with Decomposable Negation Normal Form}
In \Cref{sec:rgt_dnnf} we showed that GT, GGT, RGT, and RGGT trees can be straightforwardly encoded into the knowledge compilation language Decomposable Negation Normal Form (DNNF) \cite{darwiche2002knowledge}. This is not the case for MTO: a straightforward translation into NNF does not have the \textit{decomposable} property (that conjuncts do not share variables). Consider a non-leaf node \(O\) (with children \(L\) and \(R\)), and digit \(h\). The carry $\gamma^{h}_O$ depends on values of \(L_h\) and \(R_h\). The values of the digit \(O_h\) (represented in MTO with variables \(o^h_\sigma\)) depend on \(L_h\) and \(R_h\) as well as the carry $\gamma^{h}_O$ (see clause~\eqref{eq:mto_sumcarry_1}) and as a result the NNF terms corresponding to variables \(o^h_\sigma\) may not be decomposable. As a concrete example, suppose \(\lambda_h=3\) and there is no carry-in. \(o^h_2\) is true when \(l^h_1\) and \(r^h_1\) are both true and $\gamma^{h}_O$ is false (conjunction \textit{A}). $\gamma^{h}_O$ is true when \(l^h_1\) and \(r^h_2\) (for example), so \(l^h_1\) is mentioned in two conjuncts of \textit{A} and the NNF is not decomposable.

\subsection{Generalized n-Level Modulo Totalizer (GMTO)}\label{ssec:enc_gmto}
The generalization from MTO to GMTO is analogous to the generalization from GT to GGT. We will instantiate the leaves of the totalizer in a way that each set of variables $X_i \in {\cal X}$ is represented by a single leaf node.
The leaf node $O$ associated to
set $X_i$ will contain all variables of the form $o^h_\sigma$ involved in the representation in the selected mixed radix base $\Lambda$ of the coefficients $q_j$ of variables $x_j \in X_i$. We denote as $q_j^h$ the $d_h$ digit of the representation of $q_j$ in base $\Lambda$.
For each set $X_i \in {\cal X}$, with leaf node name $O$, for each $d_h$ digit with $h \in 0..\beta$, and for each value $\sigma \in 1..\lambda_h$:
\begin{itemize}
	\item If there is no coefficient $q_j$ such that $x_j \in X_i$ and $q_j^h=\sigma$, then list $O_h$ does not contain variable $o^h_\sigma$.
	\item If there is only one coefficient $q_j$ such that $x_j \in X_i$ and $q_j^h=\sigma$, then variable $o^h_\sigma\in O_h$ is the same as variable $x_j$.
	\item If there is more than one coefficient $q_j$ such that $x_j \in X_i$ and $q_j^h=\sigma$, then variable $o^h_\sigma\in O_h$ is a new variable, and we add the clauses $\noo{x_j} \lor o^h_\sigma$ for any such $q_j$.

\end{itemize}
As in MTO,  $o_0^h$ is defined as the 1 constant and is always present for each  $h\in0..\beta$ and for all nodes. An example is given in Figure~\ref{fig:gmto} (the constants $o_0^h$ are not represented).

Therefore, the GMTO encoding consists of the previous definition of the totalizer together with clauses~\eqref{eq:mto_sumnocarry_1}--\eqref{eq:mto_last_k}.
Note that assuming that an AMO constraint over each set
$X_i$ is satisfied, the value of the leaf node associated to $X_i$ will be at least $q_j$, where $x_j\in X_i$ is the variable that is set to true, and therefore the encoding correctly evaluates
$\sum_{i=1}^{n} q_ix_i \leq K$.

The MTO encoding requires $O(n\beta\lambda_0)$ auxiliary variables and
$O(n\beta\lambda_0^2)$ clauses (assuming
$\lambda_0=\dots=\lambda_{\beta-1}=\lceil
K^{\frac{1}{\beta}}\rceil$)~\cite{zha2019n}, while the GMTO encoding
requires $O(N\beta\lambda_0)$ auxiliary variables and
$O(N\beta\lambda_0^2)$ clauses.

\subsection{Construction of (G)MTO}\label{ssec:const_gmto}
In our experiments, in order to build the n-Level Modulo Totalizer, we
follow Algorithm~5 from~\cite{zha2019n} which produces a balanced
binary tree, with very small sizes and good performance in our
experiments. Regarding the
selection of a mixed radix base, we have implemented a greedy
heuristic based on the description given in Section~4.3
of~\cite{zha2019n}. We keep adding values to $\Lambda$ until
$\prod_{\lambda_h \in \Lambda} \lambda_h > K$. To select the new value
$\lambda_h$ to add to $\Lambda$, we choose the number greater than 1
which is a divisor of the largest number of coefficients of the PB
constraint and, in case of a tie, we choose the highest value. Each
time we add a new value $\lambda_h$ to $\Lambda$, all coefficients
$q_i$ are updated as \(q_i= \lfloor q_i/\lambda_h \rfloor\) in order to
find the best divisor (base) for the next digit.



\section{Global Polynomial Watchdog Encoding}\label{sec:enc_ggpw}

In this section we first revisit the Global Polynomial Watchdog encoding for PB constraints from~\cite{BailleuxBR09} and in Subsection~\ref{ssec:enc_ggpw} we provide its generalization to encode PB(AMO) constraints.

\subsection{Global Polynomial Watchdog}

The \emph{Global Polynomial Watchdog} (GPW) encoding was introduced
by Bailleux et al~\cite{BailleuxBR09}.  It uses as its basis a \emph{polynomial watchdog}
formula, denoted $PW(P)$, which is associated with a PB constraint
$P$, and contains a variable named the \emph{output} variable,
denoted~$w$. The $PW(P)$ formula satisfies the following property:

\begin{lemma}[Lemma~1 in~\cite{BailleuxBR09}]\label{lemma:pw_check}
  For any partial assignment to the variables of $P$, unit propagation on $PW(P)$
  assigns 1 to~$w$ if and only if this partial assignment cannot be
  extended to a model of $P$.
\end{lemma}

We first summarise how to construct the formula $PW(P)$ and then
complete the definition of the GPW encoding. The first step is to rewrite
the constraint into the form $T + \sum_{i=1}^n{q_ix_i} < m\cdot2^p$, with a
strict inequality, where $p$, $T$ and $m$ are defined as follows:
$p = \lfloor \log_2(\max_{i=1..n}(q_i)) \rfloor$ is the index of the
most significant bit in the binary representation of the largest
coefficient $q_i$, where 0 is the index of the least significant bit. In
other words, $p+1$ is the number of bits needed to represent $q_i$ in
binary notation; $T$ is the smallest non-negative integer such that
$K + 1 + T$ is a multiple of $2^p$; $m = (K+1+T)/2^p$.

Once the constraint is rewritten to this form, a set
$B_r$ of variables of $P$ (called \emph{bucket}) is computed for each bit
$0 \leq r \leq p$. We denote by $b_r(q_i)$ the $r$-th bit of the
binary representation of the integer $q_i$. Bucket $B_r$ contains all
the variables $x_i$ such that $b_r(q_i)=1$. Bucket $B_r$ also contains
a 1 constant if $b_r(T)=1$.
\begin{example}\label{ex:pw}
  The following is the transformation to apply to the PB constraint
  $2x_1 + 3x_2 + 4x_3 + 7x_4 \leq 8$. We have $p = 2$, and $T=3$ is
  the smallest integer such that $K+1+T=12$ is a multiple of $2^p$, with
  $m=3$. Therefore, the constraint is expressed as
  $3 + 2x_1 + 3x_2 + 4x_3 + 7x_4 < 12$. The content of buckets
  $B_0$, $B_1$ and $B_2$ is illustrated in Figure~\ref{fig:pw}.
\end{example}
\begin{figure}[!t]
\centering
\subcaptionbox{$PW(2x_1 + 3x_2 + 4x_3 + 7x_4 \leq 8)$.\label{fig:gpw}}{
\footnotesize
\resizebox{12cm}{!}{
\begin{tikzpicture}[auto,node distance=1.5cm,>=stealth]

\tikzstyle{box}=[rectangle,thick,draw,minimum width=2.7cm,minimum height=1cm,anchor=south west]
\tikzstyle{txt}=[anchor=south]
\node[box] at (0,0) [] {$\phi(\langle B_0\rangle)$};
\node[box] at (4.1,0) [] {$\phi(\langle B_1\rangle)$};
\node[box] at (8.2,0) [] {$\phi(\langle B_2\rangle)$};

\node[box] at (4.1,-2) [] {$\psi(U(\langle B_1\rangle),S_0^{1/2})$};
\node[box] at (8.2,-2) [] {$\psi(U(\langle B_2\rangle),S_1^{1/2})$};

\draw (0.3, 1.2) -- (0.3, 1); \node [txt] at (0.3, 1.2) [] {$1$};   \draw (0.3, 0) -- (0.3, -0.4);
\draw (1.0, 1.2) -- (1.0, 1); \node [txt] at (1.0, 1.2) [] {$x_2$}; \draw (1.0, 0) -- (1.0, -1.7);
\draw (1.7, 1.2) -- (1.7, 1); \node [txt] at (1.7, 1.2) [] {$x_4$}; \draw (1.7, 0) -- (1.7, -0.4);
\node [txt] at (0.3, 1.15) [anchor=south east] {$B_0:\langle$};
\node [txt] at (1.8, 1.15) [anchor=south west] {$\rangle$};

\draw (1.0, -1.7) -- (4.1, -1.7);
\node [txt] at (2.75, -0.8) [] {$U(\langle B_0 \rangle)=S_0$};
\node [txt] at (3.7, -2.55) [] {$S_0^{1/2}$};
\draw (3.6,-1.5) ellipse (0.05 and 0.3);

\draw (4.4, 1.2) -- (4.4, 1); \node [txt] at (4.4, 1.2) [] {$1$};   \draw (4.4, 0) -- (4.4, -1);
\draw (5.1, 1.2) -- (5.1, 1); \node [txt] at (5.1, 1.2) [] {$x_1$}; \draw (5.1, 0) -- (5.1, -1);
\draw (5.8, 1.2) -- (5.8, 1); \node [txt] at (5.8, 1.2) [] {$x_2$}; \draw (5.8, 0) -- (5.8, -1);
\draw (6.5, 1.2) -- (6.5, 1); \node [txt] at (6.5, 1.2) [] {$x_4$}; \draw (6.5, 0) -- (6.5, -1);
\node [txt] at (4.4, 1.15) [anchor=south east] {$B_1:\langle$};
\node [txt] at (6.6, 1.15) [anchor=south west] {$\rangle$};

\draw (6.8, -1.1) -- (7, -1.1);
\draw (6.8, -1.3) -- (8.2, -1.3);
\draw (6.8, -1.5) -- (7, -1.5);
\draw (6.8, -1.7) -- (8.2, -1.7);
\draw (6.8, -1.9) -- (7, -1.9);
\node [txt] at (7.2, -0.8) [] {$U(\langle B_1 \rangle)$};

\draw (1.35,-0.2) ellipse (1.35 and 0.05);
\draw (5.45,-0.2) ellipse (1.35 and 0.05);
\draw (9.55,-0.2) ellipse (1.35 and 0.05);

\draw (7.8,-1.5) ellipse (0.05 and 0.3);
\node [txt] at (7.9, -2.55) [] {$S_1^{1/2}$};
\draw (7,-1.5) ellipse (0.05 and 0.5);
\node[txt] at (7.08,-2.5) {$S_1$};

\draw (8.5, 1.2) -- (8.5, 1); \node [txt] at (8.5, 1.2) [] {$x_3$}; \draw (8.5, 0) -- (8.5, -1);
\draw (9.2, 1.2) -- (9.2, 1); \node [txt] at (9.2, 1.2) [] {$x_4$}; \draw (9.2, 0) -- (9.2, -1);
\node [txt] at (8.45, 1.15) [anchor=south east] {$B_2:\langle$};
\node [txt] at (9.3, 1.15) [anchor=south west] {$\rangle$};

\draw (10.9, -1.3) -- (11.1, -1.3);
\draw (10.9, -1.5) -- (11.4, -1.5);  \node [txt] at (11.4, -1.5) [anchor=west] {$w$};
\draw (10.9, -1.7) -- (11.1, -1.7);
\draw (10.9, -1.9) -- (11.1, -1.9);
\node [txt] at (9.9, -0.8) [] {$U(\langle B_2 \rangle)$};

\draw (11.10,-1.5) ellipse (0.05 and 0.5);

\node[txt] at (11.18,-2.5) {$S_2$};

\end{tikzpicture}
}
}

\vspace{0.5cm}

\subcaptionbox{$PW(2x_1 + 3x_2 + 4x_3 + 7x_4  \leq 8,\{\{x_1,x_2\},\{x_3,x_4\}\})$.\label{fig:ggpw}}{
\footnotesize
\resizebox{12cm}{!}{
\begin{tikzpicture}[auto,node distance=1.5cm,>=stealth]
\tikzstyle{box}=[rectangle,thick,draw,minimum width=2.7cm,minimum height=1cm,anchor=south west]
\tikzstyle{txt}=[anchor=south]
\node[box] at (0,0) [] {$\phi(\langle B_0\rangle)$};
\node[box] at (4.1,0) [] {$\phi(\langle B_1\rangle)$};
\node[box] at (8.2,0) [] {$\phi(\langle B_2\rangle)$};

\node[box] at (4.1,-2) [] {$\psi(U(\langle B_1\rangle),S_0^{1/2})$};
\node[box] at (8.2,-2) [] {$\psi(U(\langle B_2\rangle),S_1^{1/2})$};

\draw (0.3, 1.2) -- (0.3, 1); \node [txt] at (0.3, 1.2) [] {$1$};   \draw (0.3, 0) -- (0.3, -0.4);
\draw (1.0, 1.2) -- (1.0, 1); \node [txt] at (1.0, 1.2) [] {$y_{1,0}$}; \draw (1.0, 0) -- (1.0, -1.7);
\draw (1.7, 1.2) -- (1.7, 1); \node [txt] at (1.7, 1.2) [] {$y_{2,0}$}; \draw (1.7, 0) -- (1.7, -0.4);
\node [txt] at (0.3, 1.15) [anchor=south east] {$B_0:\langle$};
\node [txt] at (1.8, 1.15) [anchor=south west] {$\rangle$};

\node [txt] at (1.0, 1.8) [] {$x_2$};
\node [txt] at (1.2, 1.7) [rotate=90] {$=$};

\node [txt] at (1.7, 1.8) [] {$x_4$};
\node [txt] at (1.9, 1.7) [rotate=90] {$=$};

\draw (1.0, -1.7) -- (4.1, -1.7);
\node [txt] at (2.75, -0.8) [] {$U(\langle B_0 \rangle)=S_0$};
\node [txt] at (3.7, -2.55) [] {$S_0^{1/2}$};
\draw (3.6,-1.5) ellipse (0.05 and 0.3);

\draw (4.4, 1.2) -- (4.4, 1); \node [txt] at (4.4, 1.2) [] {$1$};   \draw (4.4, 0) -- (4.4, -1);
\draw (5.1, 1.2) -- (5.1, 1); \node [txt] at (5.1, 1.2) [] {$y_{1,1}$}; \draw (5.1, 0) -- (5.1, -1);
\draw (5.8, 1.2) -- (5.8, 1); \node [txt] at (5.8, 1.2) [] {$y_{2,1}$}; \draw (5.8, 0) -- (5.8, -1);
\node [txt] at (4.4, 1.15) [anchor=south east] {$B_1:\langle$};
\node [txt] at (5.9, 1.15) [anchor=south west] {$\rangle$};

\node [txt] at (4.9, 1.8) [] {$x_1$};
\node [txt] at (4.7, 1.7) [rotate=-80] {$\rightarrow$};
\node [txt] at (5.3, 1.8) [] {$x_2$};
\node [txt] at (5.5, 1.7) [rotate=77] {$\leftarrow$};

\node [txt] at (5.8, 1.8) [] {$x_4$};
\node [txt] at (6.0, 1.7) [rotate=90] {$=$};

\draw (6.8, -1.3) -- (8.2, -1.3);
\draw (6.8, -1.5) -- (7, -1.5);
\draw (6.8, -1.7) -- (8.2, -1.7);
\draw (6.8, -1.9) -- (7, -1.9);
\node [txt] at (7.2, -0.8) [] {$U(\langle B_1 \rangle)$};

\draw (1.35,-0.2) ellipse (1.35 and 0.05);
\draw (5.45,-0.2) ellipse (1.35 and 0.05);
\draw (9.55,-0.2) ellipse (1.35 and 0.05);

\draw (7.8,-1.5) ellipse (0.05 and 0.3);
\node [txt] at (7.9, -2.55) [] {$S_1^{1/2}$};
\draw (7,-1.5) ellipse (0.05 and 0.5);
\node[txt] at (7.08,-2.5) {$S_1$};

\draw (8.5, 1.2) -- (8.5, 1); \node [txt] at (8.5, 1.2) [] {$y_{2,2}$}; \draw (8.5, 0) -- (8.5, -1);
\node [txt] at (8.3, 1.15) [anchor=south east] {$B_2:\langle$};
\node [txt] at (8.65, 1.15) [anchor=south west] {$\rangle$};

\node [txt] at (8.3, 1.8) [] {$x_3$};
\node [txt] at (8.1, 1.7) [rotate=-80] {$\rightarrow$};
\node [txt] at (8.7, 1.8) [] {$x_4$};
\node [txt] at (8.9, 1.7) [rotate=77] {$\leftarrow$};

\draw (10.9, -1.5) -- (11.4, -1.5);  \node [txt] at (11.4, -1.5) [anchor=west] {$w$};
\draw (10.9, -1.7) -- (11.1, -1.7);
\draw (10.9, -1.9) -- (11.1, -1.9);
\node [txt] at (9.9, -0.8) [] {$U(\langle B_2 \rangle)$};

\draw (11.10,-1.5) ellipse (0.05 and 0.5);

\node[txt] at (11.18,-2.5) {$S_2$};
\end{tikzpicture}
}
}
\caption[]{(a): circuit representation of $PW(2x_1 + 3x_2 + 4x_3 + 7x_4 \leq 8)$.
  (b): circuit representation of $PW(2x_1 + 3x_2 + 4x_3 + 7x_4  \leq 8,\{\{x_1,x_2\},\{x_3,x_4\}\})$.
}
\label{fig:pw}
\end{figure}

The idea is to decompose each coefficient in its binary representation
and sum each bit having the same weight.

The formula $PW(P)$ can be represented as a circuit, as can be seen in
Figure~\ref{fig:pw} corresponding to Example~\ref{ex:pw}.  We denote
by $\langle B_r \rangle$ a vector with an arbitrary order containing
the elements of bucket $B_r$.  The formula $PW(P)$ uses two main
components: the formulas $\phi(V)$ and $\psi(V_1,V_2)$.  The formula
$\phi(V)$ has as input a vector of Boolean variables $V$, and has as
output a vector of $|V|$ variables named $U(V)$. The formula
$\phi(V)$ enforces that $U(V)$ is the unary representation of the sum
of the input variables.  The formula $\psi(V_1,V_2)$, has as input two
vectors of variables $V_1$ and $V_2$, which are the unary
representation of two integers, and has as output a vector of
$|V_1|+|V_2|$ variables named $S$.  The formula $\psi(V_1,V_2)$
enforces that $S$ is the unary representation of $V_1$ + $V_2$.  In
the definition of $PW(P)$, we denote by $S_r$ the output of the $\psi$
formula related with bucket $B_r$, for $1 \leq r \leq p$, and we
define $S_0 = U(\langle B_0\rangle)$.
Half of the value of $S_k$ for a weight $2^k$, denoted as $S_k^{1/2}$ is integrated in the sum for weight $2^{k+1}$.
Then, the formula $PW(P)$ is defined as the conjunction of these two formulas:
\begin{align}
&\phi(\langle B_r\rangle) 						& 0 \leq r \leq p \label{eq:phi}\\
&\psi(U(\langle B_r \rangle),S_{r-1}^{1/2})		& 1 \leq r \leq p \label{eq:psi}
\end{align}
The GPW encoding is defined as:
\begin{align}
PW(P) \\
\noo{w} \label{eq:gpw_w}
\end{align}
The basic idea is that the $m$-th bit of $S_p$, represented with
variable $w$, is set to 1 by UP if the sum of the constraint is
greater or equal than $m\cdot2^p = K+1+T$. If $w$ is set to 1 the formula
is not satisfied.  We build formulas $\phi$ and $\psi$ as in Bailleux et al~\cite{BailleuxBR09}, where $\phi$ is encoded with a totalizer, and
$\psi$ with an adder of unary numbers.

\subsection{Generalized Global  Polynomial Watchdog (GGPW)}\label{ssec:enc_ggpw}

We define  GGPW by using a \emph{generalized polynomial
  watchdog formula} $PW(P,{\cal X})$ instead of the original polynomial
watchdog formula.
Again, $P$ is normalised to the form
$T + \sum_{i=1}^n{q_ix_i} < m\cdot2^p$ in the same way as in $PW(P)$.
For each set $X_i$, $PW(P,{\cal X})$ will contain a vector of
variables $Y_i=\langle y_{i,p},y_{i,p-1},\dots,y_{i,0} \rangle$.

$Y_i$ is interpreted as a binary number, where for all $x_l\in
X_i$ such that $x_l$ is true, at least the bits
corresponding to the binary representation of $q_l$  are set to one. 
Therefore, when exactly
one $x_l$ is true, $Y_i$ will be greater than or equal to $q_l$.
The following clauses define the variables $Y_i$:
\begin{align}
\noo{x_l}\lor y_{i,r} \qquad  & 0 \leq r \leq p,\; 1 \leq i \leq N,\;x_l \in X_i,\; b_r(q_l) = 1 \label{eq:pw_y_def}
\end{align}

In this case bucket $B_r$, for each bit $0 \leq r \leq p$, will
contain variables $y_{1,r},y_{2,r},\dots,y_{N,r}$. Bucket $B_r$
will also contain a 1 constant if $b_r(T)=1$.

The formula $PW(P,{\cal X})$ is defined as the conjunction
of~\eqref{eq:phi}, \eqref{eq:psi} and~\eqref{eq:pw_y_def}.  Some
considerations can be taken into account on Clauses~\eqref{eq:pw_y_def}
in order to optimise the encoding:
\begin{itemize}
\item If there is no $x_l \in X_i$ such that $b_r(q_l)=1$, and
  therefore variable $y_{i,r}$ does not appear in any
  clause of~\eqref{eq:pw_y_def}, then this variable is not created nor
  included in any bucket.
\item If there is only one variable $x_l \in X_i$ such that
  $b_r(q_l)=1$, then variable $y_{i,r}$ is the variable $x_l$ itself, and
  Clause~\eqref{eq:pw_y_def} is not added for $y_{i,r}$.
\item Otherwise, $y_{i,r}$ is indeed a new variable and
  Clause~\eqref{eq:pw_y_def} is added.
\end{itemize}
Figure~\ref{fig:pw} contains a circuit representation of $PW(P,{\cal X})$.

The GGPW encoding is defined by:
\begin{align}
PW(P,{\cal X})\\
\noo{w}\label{eq:ggpw_w}
\end{align}
Just as with the other newly
introduced encodings, given an assignment that satisfies an AMO
constraint over each $X_i \in {\cal X}$, this encoding represents the
PB constraint $\sum_{i=1}^{n} q_ix_i \leq K$ in a more compact way.

The GPW encoding introduces $O(n \log(n) \log(q_{max}))$ auxiliary variables
and $O(n^2\log(n)\log(q_{max}))$ clauses, while the GGPW
introduces $O(N \log(N) \log(q_{max}))$ auxiliary variables and
$O(N^2 \log(N) \log(q_{max}))$ clauses, where
$q_{max}=\max_{i=1}^n{q_i}$.  This follows from the fact that a
totalizer $\phi$ with $n$ input variables requires $O(n \log(n))$
auxiliary variables and $O(n^2 \log(n))$ clauses, and an adder $\psi$
of unary numbers with $n$ input variables requires $O(n)$ auxiliary
variables and $O(n^2)$ clauses; see~\cite{BailleuxBR09}.

\section{Local Polynomial Watchdog Encoding}\label{sec:enc_glpw}

In this section we briefly revisit the Local Polynomial Watchdog encoding for PB constraints from~\cite{BailleuxBR09} and in Subsection~\ref{ssec:enc_glpw} we provide its generalization to encode PB(AMO) constraints. For this generalization, we provide an efficient implementation that reuses many variables and clauses to obtain smaller formulas.

\subsection{Local Polynomial Watchdog}

In Section~\ref{sec:pp} we give details of the propagation strength of each encoding, and note that GPW does not have the same propagation strength as BDD, GT, RGT, and SWC.  The Local Polynomial Watchdog (LPW) does have the same propagation strength as the others mentioned, at the cost of additional variables and clauses (specifically by including a different PW formula for each variable of the PB constraint). 
The definition of $PW(P)$ is the same as in GPW, that is Constraints~\eqref{eq:phi} and~\eqref{eq:psi}. Then, LPW is defined as:

\begin{align}
PW(P[x_i]) && 1 \leq i \leq n\label{eq:lpw_pw}\\
\noo{w(P[x_i])} \lor \noo{x_i} && 1 \leq i \leq n \label{eq:lpw_w}
\end{align}
where $P[x_i]$ is the resulting PB constraint of setting $x_i$ to 1 in
$P$, and $w(P[x_i])$ is the output variable of
$PW(P[x_i])$.

\subsection{Generalized Local Polynomial Watchdog (GLPW)}\label{ssec:enc_glpw}
Similarly to LPW, we can define the GLPW to be a GAC encoding for PB(AMO) constraints. We can do that by simply modifying Formulas~\eqref{eq:lpw_pw} and~\eqref{eq:lpw_w} as follows:
\begin{align}
PW(P[x_l,X_i],\ {\cal X}\setminus \{X_i\}) && X_i \in {\cal X},\ x_l \in X_i \label{eq:glpw_pw}\\
\noo{w(P[x_l,X_i])} \lor \noo{x_l} && X_i \in {\cal X},\ x_l \in X_i \label{eq:glpw_w}
\end{align}
where $P[x_l,X_i]$ is the PB constraint resulting from setting $x_l$ to 1 in $P$ and setting any other variable $x_{l'} \in X_i$ to 0, and $w(P[x_l,X_i])$ is the output variable of  $PW(P[x_l,X_i],\ {\cal X}\setminus \{X_i\})$.

The LPW introduces $O(n^2 \log(n)\log(q_{max}))$ auxiliary variables
and $O(n^3\log(n)\log(q_{max}))$ clauses, while the GLPW encoding
introduces $O(nN \log(N) \log(q_{max}))$ auxiliary variables and
$O(nN^2 \log(N) \log(q_{max}))$ clauses, where
$q_{max}=\max_{i=1}^n{q_i}$.  Basically, the sizes are multiplied by $n$ w.r.t.\ those of GPW and GGPW, since we encode $n$ polynomial watchdog formulas, one for each variable in $P$.

Bailleux et al~\cite{BailleuxBR09} stated that the LPW encoding can be compacted by reusing some of the components of the different polynomial watchdogs introduced by Constraints~\eqref{eq:lpw_pw}, and some hints were given, but no method was detailed nor evaluated for this purpose. Here we provide implementation techniques to obtain small GLPW encodings.

Firstly, we build the totalizers for formulas $\phi$ in a way
that (i) any two formulas
$PW(P[x_l,X_i],\ {\cal X}\setminus \{X_i\})$,
$PW(P[x_{l'},X_i],\ {\cal X}\setminus \{X_i\})$ contain exactly the
same set of totalizers, and (ii) the totalizers of any two formulas
$PW(P[x_l,X_i])$, $PW(P[x_{l'},X_j])$, with $i\neq j$, share most of
the nodes. This is explained in Section~\ref{sec:glpw_totalizer}.

Secondly, we reuse auxiliary variables and clauses while constructing
formulas~\eqref{eq:glpw_pw} . This can
help not only in GLPW but also to encode PB constraints without AMOs,
i.e., in LPW encodings.  Variables and clauses are reused globally
across the different generalized local polynomial watchdog formulas
for all $X_i \in {\cal X},\ x_l \in X_i$. This is explained in
Subsection~\ref{sec:glpw_dynamic}.

\subsubsection{Structure of Totalizers}\label{sec:glpw_totalizer}
The difference between the two formulas $PW(P[x_l,X_i],\ {\cal X}\setminus \{X_i\})$ and $PW(P[x_{l'},X_i],\ {\cal X}\setminus \{X_i\})$ is just the value of the right hand side $K$ in the PB constraint.
\begin{example}
	Consider the PB constraint and partition:
	$$(P,{\cal X}) = (2x_1 + 3x_2 + 4x_3 + 7x_4  \leq 8,\{\{x_1,x_2\},\{x_3,x_4\}\})$$
	\begin{center}
		\begin{tabular}{ll}
			With $x_1=1$: &$PW(P[x_1,X_1],{\cal X}{\setminus}\{X_1\}) \equiv PW(4x_3 + 7x_4  \leq 6,\{\{x_3,x_4\}\})$\\
			With $x_2=1$: & $PW(P[x_2,X_1],{\cal X}{\setminus}\{X_1\}) \equiv PW(4x_3 + 7x_4  \leq 5,\{\{x_3,x_4\}\})$\\
			With $x_3=1$: &$PW(P[x_3,X_2],{\cal X}{\setminus}\{X_2\}) \equiv PW(2x_1 + 3x_2  \leq 4,\{\{x_1,x_2\}\})$\\
			With $x_4=1$: & $PW(P[x_4,X_2],{\cal X}{\setminus}\{X_2\}) \equiv PW(2x_1 + 3x_2  \leq 1,\{\{x_1,x_2\}\})$\\
		\end{tabular}
	\end{center}
   The only difference between $P[x_1,X_1]$ and $P[x_2,X_1]$ is the right hand side constant in $P[x_1,X_1]$ and $P[x_2,X_1]$. The same happens with
	$P[x_3,X_2]$ and $P[x_4,X_2]$.
\end{example}

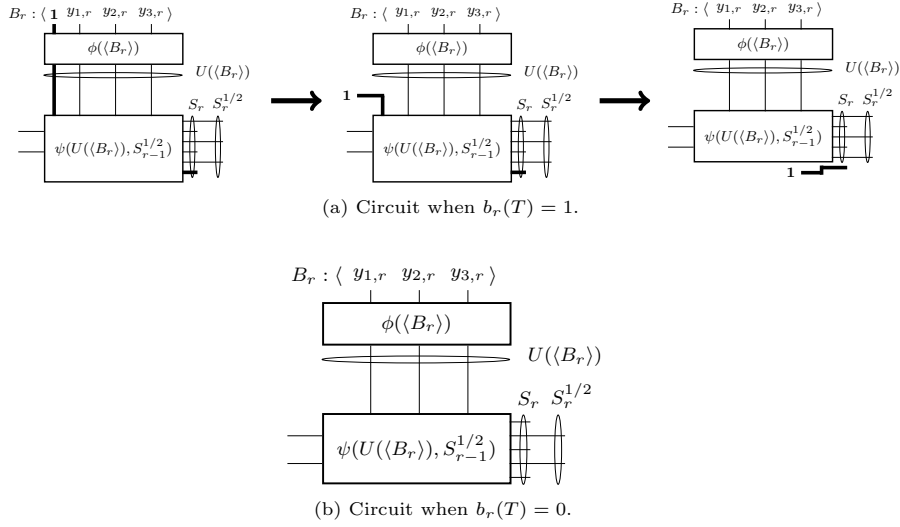
\begin{figure}[!t]
	\centering
	\footnotesize

\subcaptionbox{Circuit when $b_r(T)=1$.\label{fig:glpw_move1_a}}{
	\resizebox{\textwidth}{!}{
		\begin{tikzpicture}[auto,node distance=1.5cm,>=stealth]
		\tikzstyle{box}=[rectangle,thick,draw,minimum width=2.7cm,minimum height=1cm,anchor=south west]
		\tikzstyle{txt}=[anchor=south]

		\node[box,minimum height=0.6cm] at (4.1,0.4) [] {$\phi(\langle B_r\rangle)$};
		\node[box,minimum height=1.3cm] at (4.1,-1.9) [] {$\psi(U(\langle B_r\rangle),S_{r-1}^{1/2})$};

		\draw [line width=2pt] (4.3, 1.2) -- (4.3, 1); \node [txt] at (4.3, 1.2) [] {$\mathbf{1}$};   \draw [line width=2pt] (4.3, 0.4) -- (4.3, -0.6);
		\draw (4.8, 1.2) -- (4.8, 1); \node [txt] at (4.8, 1.2) [] {$y_{1,r}$}; \draw (4.8, 0.4) -- (4.8, -0.6);
		\draw (5.5, 1.2) -- (5.5, 1); \node [txt] at (5.5, 1.2) [] {$y_{2,r}$}; \draw (5.5, 0.4) -- (5.5, -0.6);
		\draw (6.2, 1.2) -- (6.2, 1); \node [txt] at (6.2, 1.2) [] {$y_{3,r}$}; \draw (6.2, 0.4) -- (6.2, -0.6);
		\node [txt] at (4.25, 1.15) [anchor=south east] {$B_r:\langle$};
		\node [txt] at (6.4, 1.15) [anchor=south west] {$\rangle$};
		\node [txt] at (7.6, 0) [] {$U(\langle B_r \rangle)$};

		\draw (3.6, -1.3) -- (4.1, -1.3);
		\draw (3.6, -0.9) -- (4.1, -0.9);

		\draw (6.8, -0.7) -- (7.6, -0.7);
		\draw (6.8, -0.9) -- (7.1, -0.9);
		\draw (6.8, -1.1) -- (7.6, -1.1);
		\draw (6.8, -1.3) -- (7.1, -1.3);
		\draw (6.8, -1.5) -- (7.6, -1.5);
		\draw [line width=2pt](6.8, -1.7) -- (7.1, -1.7);

		\draw (5.45,0.2) ellipse (1.35 and 0.05);

		\draw (7.5,-1.2) ellipse (0.05 and 0.6);
		\node [txt] at (7.7, -0.6) [] {$S_r^{1/2}$};
		\draw (7,-1.2) ellipse (0.05 and 0.6);
		\node[txt] at (7.08,-0.6) {$S_r$};

		\end{tikzpicture}
		\begin{tikzpicture}
		\node [] (dummy) at (0,-1.5) [] {};
		\node [] (s) at (0,0) [] {};
		\node [] (d) at (1.2,0) [] {};
		\path[->,line width=1mm] (s) edge (d);
		\end{tikzpicture}
		\begin{tikzpicture}[auto,node distance=1.5cm,>=stealth]
		\tikzstyle{box}=[rectangle,thick,draw,minimum width=2.7cm,minimum height=1cm,anchor=south west]
		\tikzstyle{txt}=[anchor=south]

		\node[box,minimum height=0.6cm] at (4.1,0.4) [] {$\phi(\langle B_r\rangle)$};
		\node[box,minimum height=1.3cm] at (4.1,-1.9) [] {$\psi(U(\langle B_r\rangle),S_{r-1}^{1/2})$};


		\draw (4.8, 1.2) -- (4.8, 1); \node [txt] at (4.8, 1.2) [] {$y_{1,r}$}; \draw (4.8, 0.4) -- (4.8, -0.6);
		\draw (5.5, 1.2) -- (5.5, 1); \node [txt] at (5.5, 1.2) [] {$y_{2,r}$}; \draw (5.5, 0.4) -- (5.5, -0.6);
		\draw (6.2, 1.2) -- (6.2, 1); \node [txt] at (6.2, 1.2) [] {$y_{3,r}$}; \draw (6.2, 0.4) -- (6.2, -0.6);
		\node [txt] at (4.5, 1.15) [anchor=south east] {$B_r:\langle$};
		\node [txt] at (6.4, 1.15) [anchor=south west] {$\rangle$};
		\node [txt] at (7.6, 0) [] {$U(\langle B_r \rangle)$};
		\node [txt] at (3.6,-0.4) [] {$\mathbf{1}$}; \draw [line width=2pt] (3.8, -0.2) -- (4.33, -0.2);   \draw [line width=2pt] (4.3, -0.2) -- (4.3, -0.6);

		\draw (3.6, -1.3) -- (4.1, -1.3);
		\draw (3.6, -0.9) -- (4.1, -0.9);

		\draw (6.8, -0.7) -- (7.6, -0.7);
		\draw (6.8, -0.9) -- (7.1, -0.9);
		\draw (6.8, -1.1) -- (7.6, -1.1);
		\draw (6.8, -1.3) -- (7.1, -1.3);
		\draw (6.8, -1.5) -- (7.6, -1.5);
		\draw [line width=2pt] (6.8, -1.7) -- (7.1, -1.7);

		\draw (5.45,0.2) ellipse (1.35 and 0.05);

		\draw (7.5,-1.2) ellipse (0.05 and 0.6);
		\node [txt] at (7.7, -0.6) [] {$S_r^{1/2}$};
		\draw (7,-1.2) ellipse (0.05 and 0.6);
		\node[txt] at (7.08,-0.6) {$S_r$};
		\end{tikzpicture}
		\begin{tikzpicture}
		\node [] (dummy) at (0,-1.5) [] {};
		\node [] (s) at (0,0) [] {};
		\node [] (d) at (1.2,0) [] {};
		\path[->,line width=1mm] (s) edge (d);
		\end{tikzpicture}
		\begin{tikzpicture}[auto,node distance=1.5cm,>=stealth]
		\tikzstyle{box}=[rectangle,thick,draw,minimum width=2.7cm,minimum height=1cm,anchor=south west]
		\tikzstyle{txt}=[anchor=south]

		\node[box,minimum height=0.6cm] at (4.1,0.4) [] {$\phi(\langle B_r\rangle)$};
		\node[box,minimum height=1.0cm] at (4.1,-1.6) [] {$\psi(U(\langle B_r\rangle),S_{r-1}^{1/2})$};


		\draw (4.8, 1.2) -- (4.8, 1); \node [txt] at (4.8, 1.2) [] {$y_{1,r}$}; \draw (4.8, 0.4) -- (4.8, -0.6);
		\draw (5.5, 1.2) -- (5.5, 1); \node [txt] at (5.5, 1.2) [] {$y_{2,r}$}; \draw (5.5, 0.4) -- (5.5, -0.6);
		\draw (6.2, 1.2) -- (6.2, 1); \node [txt] at (6.2, 1.2) [] {$y_{3,r}$}; \draw (6.2, 0.4) -- (6.2, -0.6);
		\node [txt] at (4.5, 1.15) [anchor=south east] {$B_r:\langle$};
		\node [txt] at (6.4, 1.15) [anchor=south west] {$\rangle$};
		\node [txt] at (7.6, 0) [] {$U(\langle B_r \rangle)$};

		\draw (3.6, -1.3) -- (4.1, -1.3);
		\draw (3.6, -0.9) -- (4.1, -0.9);

		\draw (6.8, -0.7) -- (7.6, -0.7);
		\draw (6.8, -0.9) -- (7.1, -0.9);
		\draw (6.8, -1.1) -- (7.6, -1.1);
		\draw (6.8, -1.3) -- (7.1, -1.3);
		\draw (6.8, -1.5) -- (7.6, -1.5);

		\node [txt] at (6,-2) [] {$\mathbf{1}$};
		 \draw [line width=2pt](6.2, -1.8) -- (6.6, -1.8);
		  \draw [line width=2pt](6.6, -1.84) -- (6.6, -1.66);
		  \draw [line width=2pt](6.6, -1.7) -- (7.1, -1.7);

		\draw (5.45,0.2) ellipse (1.35 and 0.05);

		\draw (7.5,-1.1) ellipse (0.05 and 0.5);
		\node [txt] at (7.7, -0.6) [] {$S_r^{1/2}$};
		\draw (7,-1.1) ellipse (0.05 and 0.5);
		\node[txt] at (7.08,-0.6) {$S_r$};
		\end{tikzpicture}
}
}

\vspace{0.5cm}

\subcaptionbox{Circuit when $b_r(T)=0$.\label{fig:glpw_move1_b}}{
	\resizebox{4.5cm}{!}{
		\begin{tikzpicture}[auto,node distance=1.5cm,>=stealth]
		\tikzstyle{box}=[rectangle,thick,draw,minimum width=2.7cm,minimum height=1cm,anchor=south west]
		\tikzstyle{txt}=[anchor=south]
		
		\node[box,minimum height=0.6cm] at (4.1,0.4) [] {$\phi(\langle B_r\rangle)$};
		\node[box,minimum height=1.0cm] at (4.1,-1.6) [] {$\psi(U(\langle B_r\rangle),S_{r-1}^{1/2})$};
		
		
		\draw (4.8, 1.2) -- (4.8, 1); \node [txt] at (4.8, 1.2) [] {$y_{1,r}$}; \draw (4.8, 0.4) -- (4.8, -0.6);
		\draw (5.5, 1.2) -- (5.5, 1); \node [txt] at (5.5, 1.2) [] {$y_{2,r}$}; \draw (5.5, 0.4) -- (5.5, -0.6);
		\draw (6.2, 1.2) -- (6.2, 1); \node [txt] at (6.2, 1.2) [] {$y_{3,r}$}; \draw (6.2, 0.4) -- (6.2, -0.6);
		\node [txt] at (4.5, 1.15) [anchor=south east] {$B_r:\langle$};
		\node [txt] at (6.4, 1.15) [anchor=south west] {$\rangle$};
		\node [txt] at (7.6, 0) [] {$U(\langle B_r \rangle)$};

		\draw (3.6, -1.3) -- (4.1, -1.3);
		\draw (3.6, -0.9) -- (4.1, -0.9);
		
		\draw (6.8, -0.7) -- (7.1, -0.7);
		\draw (6.8, -0.9) -- (7.6, -0.9);
		\draw (6.8, -1.1) -- (7.1, -1.1);
		\draw (6.8, -1.3) -- (7.6, -1.3);
		\draw (6.8, -1.5) -- (7.1, -1.5);

		\draw (5.45,0.2) ellipse (1.35 and 0.05);
		
		\draw (7.5,-1.1) ellipse (0.05 and 0.5);
		\node [txt] at (7.7, -0.6) [] {$S_r^{1/2}$};
		\draw (7,-1.1) ellipse (0.05 and 0.5);
		\node[txt] at (7.08,-0.6) {$S_r$};
		\end{tikzpicture}
	}
}

	\caption[]{(a): example of moving the 1 constants introduced by $T$ from the bucket to the output of formulas $\psi$. (b): representation of the corresponding formulas $\phi$ and $\psi$ when no 1 constant is introduced by $T$.
	}
	\label{fig:glpw_move1}
\end{figure}

The only thing that prevents the contents of the buckets of  $PW(P[x_l,X_i],\ {\cal X}\setminus \{X_i\})$, $PW(P[x_{l'},X_i],\ {\cal X}\setminus \{X_i\})$, and hence of the $\phi$ formulas,  to be the same, are the possible 1 constants introduced by $T$ in the normalisation step. However, since formulas $\phi$ sort the buckets, instead of putting a 1 constant in the input of $\phi$ we can append a 1 constant directly to the first position of the output of $\phi$ when required, i.e., the 1 constant now goes directly into formula $\psi$ (see first step of Figure~\ref{fig:glpw_move1_a}). This way, the content of the buckets for $PW(P[x_l,X_i],\ {\cal X}\setminus \{X_i\})$, $PW(P[x_{l'},X_i],\ {\cal X}\setminus \{X_i\})$ is exactly the same regardless of whether there is an input 1-constant. Therefore, we will only introduce one formula $\phi$, encoded as a totalizer, for each bit $r$ and for each $X_i \in {\cal X}$, instead of one totalizer for each bit $r$ and variable $x_l$. In fact we can go one step further, and since formulas $\psi$ behave as mergers of two sorted lists, we can move the 1 constant directly to the output of formula $\psi$ (see second step of Figure~\ref{fig:glpw_move1_a}). This second move of the 1 constant reduces to a small extent  the sizes of formulas $\psi$, and most importantly, makes the content of $S_r$ independent of any 1 constant in bucket $B_r$. This is illustrated in Figure~\ref{fig:glpw_move1}, where $S_r$ is exactly the same when  $b_r(T)=1$ and when $b_r(T)=0$.  This lets us reuse the output of formulas $\psi$ as will be explained in Section~\ref{sec:glpw_dynamic}, thus saving variables and clauses.

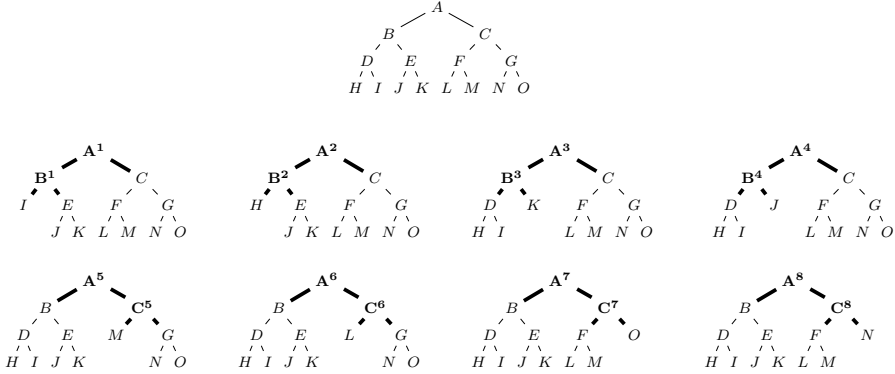
\begin{figure}[!t]
	\centering
	\footnotesize
	\resizebox{2.7cm}{!}{
		\begin{tikzpicture}[auto,node distance= 0.3cm and 0.1cm,>=stealth]
		\tikzstyle{tex}=[anchor=south]

		\node[tex] (H) []             {$H$};
		\node[tex] (I) [right = 0cm of H] {$I$};
		\node[tex] (J) [right = 0cm of I] {$J$};
		\node[tex] (K) [right = 0cm of J] {$K$};
		\node[tex] (L) [right = 0cm of K] {$L$};
		\node[tex] (M) [right = 0cm of L] {$M$};
		\node[tex] (N) [right = 0cm of M] {$N$};
		\node[tex] (O) [right = 0cm of N] {$O$};

		\coordinate (Middle) at ($(H)!0.5!(I)$); \node[tex] (D) [above = of Middle] {$D$};
		\coordinate (Middle) at ($(J)!0.5!(K)$); \node[tex] (E) [above = of Middle] {$E$};
		\coordinate (Middle) at ($(L)!0.5!(M)$); \node[tex] (F) [above = of Middle] {$F$};
		\coordinate (Middle) at ($(N)!0.5!(O)$); \node[tex] (G) [above = of Middle] {$G$};

		\coordinate (Middle) at ($(D)!0.5!(E)$); \node[tex] (B) [above = of Middle] {$B$};
		\coordinate (Middle) at ($(F)!0.5!(G)$); \node[tex] (C) [above = of Middle] {$C$};

		\coordinate (Middle) at ($(B)!0.5!(C)$); \node[tex] (A) [above = of Middle] {$A$};

		\path[-] (A) edge (B); \path[-] (A) edge (C);

		\path[-] (B) edge (D); \path[-] (B) edge (E);
		\path[-] (C) edge (F); \path[-] (C) edge (G);

		\path[-] (D) edge (H); \path[-] (D) edge (I);
		\path[-] (E) edge (K); \path[-] (E) edge (J);
		\path[-] (F) edge (L); \path[-] (F) edge (M);
		\path[-] (G) edge (N); \path[-] (G) edge (O);

		\end{tikzpicture}
	}
\vspace{0.5cm}

	\makebox[\width][c]{
	\resizebox{2.7cm}{!}{
	\begin{tikzpicture}[auto,node distance= 0.3cm and 0.1cm,>=stealth]
	\tikzstyle{tex}=[anchor=south]

	\node[tex] (H) []             {\phantom{$H$}};
	\node[tex] (I) [right = 0cm of H] {\phantom{$I$}};
	\node[tex] (J) [right = 0cm of I] {$J$};
	\node[tex] (K) [right = 0cm of J] {$K$};
	\node[tex] (L) [right = 0cm of K] {$L$};
	\node[tex] (M) [right = 0cm of L] {$M$};
	\node[tex] (N) [right = 0cm of M] {$N$};
	\node[tex] (O) [right = 0cm of N] {$O$};

	\coordinate (Middle) at ($(H)!0.5!(I)$); \node[tex] (D) [above = of Middle] {$I$};
	\coordinate (Middle) at ($(J)!0.5!(K)$); \node[tex] (E) [above = of Middle] {$E$};
	\coordinate (Middle) at ($(L)!0.5!(M)$); \node[tex] (F) [above = of Middle] {$F$};
	\coordinate (Middle) at ($(N)!0.5!(O)$); \node[tex] (G) [above = of Middle] {$G$};

	\coordinate (Middle) at ($(D)!0.5!(E)$); \node[tex] (B) [above = of Middle] {$\mathbf{B^1}$};
	\coordinate (Middle) at ($(F)!0.5!(G)$); \node[tex] (C) [above = of Middle] {$C$};

	\coordinate (Middle) at ($(B)!0.5!(C)$); \node[tex] (A) [above = of Middle] {$\mathbf{A^1}$};

	\path[-, line width=2pt] (A) edge (B); \path[-, line width=2pt] (A) edge (C);

	\path[-, line width=2pt] (B) edge (D); \path[-, line width=2pt] (B) edge (E);
	\path[-] (C) edge (F); \path[-] (C) edge (G);

	\path[-] (E) edge (K); \path[-] (E) edge (J);
	\path[-] (F) edge (L); \path[-] (F) edge (M);
	\path[-] (G) edge (N); \path[-] (G) edge (O);

	\end{tikzpicture}
}\hspace{0.3cm}
	\resizebox{2.7cm}{!}{
	\begin{tikzpicture}[auto,node distance= 0.3cm and 0.1cm,>=stealth]
	\tikzstyle{tex}=[anchor=south]

	\node[tex] (H) []             {\phantom{$H$}};
\node[tex] (I) [right = 0cm of H] {\phantom{$I$}};
	\node[tex] (J) [right = 0cm of I] {$J$};
	\node[tex] (K) [right = 0cm of J] {$K$};
	\node[tex] (L) [right = 0cm of K] {$L$};
	\node[tex] (M) [right = 0cm of L] {$M$};
	\node[tex] (N) [right = 0cm of M] {$N$};
	\node[tex] (O) [right = 0cm of N] {$O$};

	\coordinate (Middle) at ($(H)!0.5!(I)$); \node[tex] (D) [above = of Middle] {$H$};
	\coordinate (Middle) at ($(J)!0.5!(K)$); \node[tex] (E) [above = of Middle] {$E$};
	\coordinate (Middle) at ($(L)!0.5!(M)$); \node[tex] (F) [above = of Middle] {$F$};
	\coordinate (Middle) at ($(N)!0.5!(O)$); \node[tex] (G) [above = of Middle] {$G$};

	\coordinate (Middle) at ($(D)!0.5!(E)$); \node[tex] (B) [above = of Middle] {$\mathbf{B^2}$};
	\coordinate (Middle) at ($(F)!0.5!(G)$); \node[tex] (C) [above = of Middle] {$C$};

	\coordinate (Middle) at ($(B)!0.5!(C)$); \node[tex] (A) [above = of Middle] {$\mathbf{A^2}$};

	\path[-, line width=2pt] (A) edge (B); \path[-, line width=2pt] (A) edge (C);

	\path[-, line width=2pt] (B) edge (D); \path[-, line width=2pt] (B) edge (E);
	\path[-] (C) edge (F); \path[-] (C) edge (G);

	\path[-] (E) edge (K); \path[-] (E) edge (J);
	\path[-] (F) edge (L); \path[-] (F) edge (M);
	\path[-] (G) edge (N); \path[-] (G) edge (O);

	\end{tikzpicture}
}\hspace{0.3cm}
	\resizebox{2.7cm}{!}{
	\begin{tikzpicture}[auto,node distance= 0.3cm and 0.1cm,>=stealth]
	\tikzstyle{tex}=[anchor=south]

	\node[tex] (H) []             {$H$};
	\node[tex] (I) [right = 0cm of H] {$I$};
	\node[tex] (J) [right = 0cm of I] {\phantom{$J$}};
	\node[tex] (K) [right = 0cm of J] {\phantom{$K$}};
	\node[tex] (L) [right = 0cm of K] {$L$};
	\node[tex] (M) [right = 0cm of L] {$M$};
	\node[tex] (N) [right = 0cm of M] {$N$};
	\node[tex] (O) [right = 0cm of N] {$O$};

	\coordinate (Middle) at ($(H)!0.5!(I)$); \node[tex] (D) [above = of Middle] {$D$};
	\coordinate (Middle) at ($(J)!0.5!(K)$); \node[tex] (E) [above = of Middle] {$K$};
	\coordinate (Middle) at ($(L)!0.5!(M)$); \node[tex] (F) [above = of Middle] {$F$};
	\coordinate (Middle) at ($(N)!0.5!(O)$); \node[tex] (G) [above = of Middle] {$G$};

	\coordinate (Middle) at ($(D)!0.5!(E)$); \node[tex] (B) [above = of Middle] {$\mathbf{B^3}$};
	\coordinate (Middle) at ($(F)!0.5!(G)$); \node[tex] (C) [above = of Middle] {$C$};

	\coordinate (Middle) at ($(B)!0.5!(C)$); \node[tex] (A) [above = of Middle] {$\mathbf{A^3}$};

	\path[-, line width=2pt] (A) edge (B); \path[-, line width=2pt] (A) edge (C);

	\path[-, line width=2pt] (B) edge (D); \path[-, line width=2pt] (B) edge (E);
	\path[-] (C) edge (F); \path[-] (C) edge (G);

	\path[-] (D) edge (H); \path[-] (D) edge (I);
	\path[-] (F) edge (L); \path[-] (F) edge (M);
	\path[-] (G) edge (N); \path[-] (G) edge (O);
	\end{tikzpicture}
}	\hspace{0.3cm}
	\resizebox{2.7cm}{!}{
		\begin{tikzpicture}[auto,node distance= 0.3cm and 0.1cm,>=stealth]
		\tikzstyle{tex}=[anchor=south]

		\node[tex] (H) []             {$H$};
		\node[tex] (I) [right = 0cm of H] {$I$};
		\node[tex] (J) [right = 0cm of I] {\phantom{$J$}};
		\node[tex] (K) [right = 0cm of J] {\phantom{$K$}};
		\node[tex] (L) [right = 0cm of K] {$L$};
		\node[tex] (M) [right = 0cm of L] {$M$};
		\node[tex] (N) [right = 0cm of M] {$N$};
		\node[tex] (O) [right = 0cm of N] {$O$};

		\coordinate (Middle) at ($(H)!0.5!(I)$); \node[tex] (D) [above = of Middle] {$D$};
		\coordinate (Middle) at ($(J)!0.5!(K)$); \node[tex] (E) [above = of Middle] {$J$};
		\coordinate (Middle) at ($(L)!0.5!(M)$); \node[tex] (F) [above = of Middle] {$F$};
		\coordinate (Middle) at ($(N)!0.5!(O)$); \node[tex] (G) [above = of Middle] {$G$};

		\coordinate (Middle) at ($(D)!0.5!(E)$); \node[tex] (B) [above = of Middle] {$\mathbf{B^4}$};
		\coordinate (Middle) at ($(F)!0.5!(G)$); \node[tex] (C) [above = of Middle] {$C$};

		\coordinate (Middle) at ($(B)!0.5!(C)$); \node[tex] (A) [above = of Middle] {$\mathbf{A^4}$};

		\path[-, line width=2pt] (A) edge (B); \path[-, line width=2pt] (A) edge (C);

		\path[-, line width=2pt] (B) edge (D); \path[-, line width=2pt] (B) edge (E);
		\path[-] (C) edge (F); \path[-] (C) edge (G);

		\path[-] (D) edge (H); \path[-] (D) edge (I);
		\path[-] (F) edge (L); \path[-] (F) edge (M);
		\path[-] (G) edge (N); \path[-] (G) edge (O);
		\end{tikzpicture}
	}
}

\vspace{0.3cm}
\makebox[\width][c]{
		\resizebox{2.7cm}{!}{
			\begin{tikzpicture}[auto,node distance= 0.3cm and 0.1cm,>=stealth]
			\tikzstyle{tex}=[anchor=south]

			\node[tex] (H) []             {$H$};
			\node[tex] (I) [right = 0cm of H] {$I$};
			\node[tex] (J) [right = 0cm of I] {$J$};
			\node[tex] (K) [right = 0cm of J] {$K$};
			\node[tex] (L) [right = 0cm of K] {\phantom{$L$}};
			\node[tex] (M) [right = 0cm of L] {\phantom{$M$}};
			\node[tex] (N) [right = 0cm of M] {$N$};
			\node[tex] (O) [right = 0cm of N] {$O$};

			\coordinate (Middle) at ($(H)!0.5!(I)$); \node[tex] (D) [above = of Middle] {$D$};
			\coordinate (Middle) at ($(J)!0.5!(K)$); \node[tex] (E) [above = of Middle] {$E$};
			\coordinate (Middle) at ($(L)!0.5!(M)$); \node[tex] (F) [above = of Middle] {$M$};
			\coordinate (Middle) at ($(N)!0.5!(O)$); \node[tex] (G) [above = of Middle] {$G$};

			\coordinate (Middle) at ($(D)!0.5!(E)$); \node[tex] (B) [above = of Middle] {$B$};
			\coordinate (Middle) at ($(F)!0.5!(G)$); \node[tex] (C) [above = of Middle] {$\mathbf{C^5}$};

			\coordinate (Middle) at ($(B)!0.5!(C)$); \node[tex] (A) [above = of Middle] {$\mathbf{A^5}$};

			\path[-, line width=2pt] (A) edge (B); \path[-, line width=2pt] (A) edge (C);

			\path[-] (B) edge (D); \path[-] (B) edge (E);
			\path[-, line width=2pt] (C) edge (F); \path[-, line width=2pt] (C) edge (G);

			\path[-] (D) edge (H); \path[-] (D) edge (I);
			\path[-] (E) edge (K); \path[-] (E) edge (J);
			\path[-] (G) edge (N); \path[-] (G) edge (O);

			\end{tikzpicture}
		}\hspace{0.3cm}
		\resizebox{2.7cm}{!}{
		\begin{tikzpicture}[auto,node distance= 0.3cm and 0.1cm,>=stealth]
		\tikzstyle{tex}=[anchor=south]

		\node[tex] (H) []             {$H$};
		\node[tex] (I) [right = 0cm of H] {$I$};
		\node[tex] (J) [right = 0cm of I] {$J$};
		\node[tex] (K) [right = 0cm of J] {$K$};
		\node[tex] (L) [right = 0cm of K] {\phantom{$L$}};
		\node[tex] (M) [right = 0cm of L] {\phantom{$M$}};
		\node[tex] (N) [right = 0cm of M] {$N$};
		\node[tex] (O) [right = 0cm of N] {$O$};

		\coordinate (Middle) at ($(H)!0.5!(I)$); \node[tex] (D) [above = of Middle] {$D$};
		\coordinate (Middle) at ($(J)!0.5!(K)$); \node[tex] (E) [above = of Middle] {$E$};
		\coordinate (Middle) at ($(L)!0.5!(M)$); \node[tex] (F) [above = of Middle] {$L$};
		\coordinate (Middle) at ($(N)!0.5!(O)$); \node[tex] (G) [above = of Middle] {$G$};

		\coordinate (Middle) at ($(D)!0.5!(E)$); \node[tex] (B) [above = of Middle] {$B$};
		\coordinate (Middle) at ($(F)!0.5!(G)$); \node[tex] (C) [above = of Middle] {$\mathbf{C^6}$};

		\coordinate (Middle) at ($(B)!0.5!(C)$); \node[tex] (A) [above = of Middle] {$\mathbf{A^6}$};

		\path[-, line width=2pt] (A) edge (B); \path[-, line width=2pt] (A) edge (C);

		\path[-] (B) edge (D); \path[-] (B) edge (E);
		\path[-, line width=2pt] (C) edge (F); \path[-, line width=2pt] (C) edge (G);

		\path[-] (D) edge (H); \path[-] (D) edge (I);
		\path[-] (E) edge (K); \path[-] (E) edge (J);
		\path[-] (G) edge (N); \path[-] (G) edge (O);

		\end{tikzpicture}
	}\hspace{0.3cm}
	\resizebox{2.7cm}{!}{
		\begin{tikzpicture}[auto,node distance= 0.3cm and 0.1cm,>=stealth]
		\tikzstyle{tex}=[anchor=south]

		\node[tex] (H) []             {$H$};
		\node[tex] (I) [right = 0cm of H] {$I$};
		\node[tex] (J) [right = 0cm of I] {$J$};
		\node[tex] (K) [right = 0cm of J] {$K$};
		\node[tex] (L) [right = 0cm of K] {$L$};
		\node[tex] (M) [right = 0cm of L] {$M$};
		\node[tex] (N) [right = 0cm of M] {\phantom{$N$}};
		\node[tex] (O) [right = 0cm of N] {\phantom{$O$}};

		\coordinate (Middle) at ($(H)!0.5!(I)$); \node[tex] (D) [above = of Middle] {$D$};
		\coordinate (Middle) at ($(J)!0.5!(K)$); \node[tex] (E) [above = of Middle] {$E$};
		\coordinate (Middle) at ($(L)!0.5!(M)$); \node[tex] (F) [above = of Middle] {$F$};
		\coordinate (Middle) at ($(N)!0.5!(O)$); \node[tex] (G) [above = of Middle] {$O$};

		\coordinate (Middle) at ($(D)!0.5!(E)$); \node[tex] (B) [above = of Middle] {$B$};
		\coordinate (Middle) at ($(F)!0.5!(G)$); \node[tex] (C) [above = of Middle] {$\mathbf{C^7}$};

		\coordinate (Middle) at ($(B)!0.5!(C)$); \node[tex] (A) [above = of Middle] {$\mathbf{A^7}$};

		\path[-, line width=2pt] (A) edge (B); \path[-, line width=2pt] (A) edge (C);

		\path[-] (B) edge (D); \path[-] (B) edge (E);
		\path[-, line width=2pt] (C) edge (F); \path[-, line width=2pt] (C) edge (G);

		\path[-] (D) edge (H); \path[-] (D) edge (I);
		\path[-] (E) edge (K); \path[-] (E) edge (J);
		\path[-] (F) edge (L); \path[-] (F) edge (M);
		\end{tikzpicture}
	}\hspace{0.3cm}
\resizebox{2.7cm}{!}{
	\begin{tikzpicture}[auto,node distance= 0.3cm and 0.1cm,>=stealth]
	\tikzstyle{tex}=[anchor=south]

	\node[tex] (H) []             {$H$};
	\node[tex] (I) [right = 0cm of H] {$I$};
	\node[tex] (J) [right = 0cm of I] {$J$};
	\node[tex] (K) [right = 0cm of J] {$K$};
	\node[tex] (L) [right = 0cm of K] {$L$};
	\node[tex] (M) [right = 0cm of L] {$M$};
	\node[tex] (N) [right = 0cm of M] {\phantom{$N$}};
	\node[tex] (O) [right = 0cm of N] {\phantom{$O$}};

	\coordinate (Middle) at ($(H)!0.5!(I)$); \node[tex] (D) [above = of Middle] {$D$};
	\coordinate (Middle) at ($(J)!0.5!(K)$); \node[tex] (E) [above = of Middle] {$E$};
	\coordinate (Middle) at ($(L)!0.5!(M)$); \node[tex] (F) [above = of Middle] {$F$};
	\coordinate (Middle) at ($(N)!0.5!(O)$); \node[tex] (G) [above = of Middle] {$N$};

	\coordinate (Middle) at ($(D)!0.5!(E)$); \node[tex] (B) [above = of Middle] {$B$};
	\coordinate (Middle) at ($(F)!0.5!(G)$); \node[tex] (C) [above = of Middle] {$\mathbf{C^8}$};

	\coordinate (Middle) at ($(B)!0.5!(C)$); \node[tex] (A) [above = of Middle] {$\mathbf{A^8}$};

	\path[-, line width=2pt] (A) edge (B); \path[-, line width=2pt] (A) edge (C);

	\path[-] (B) edge (D); \path[-] (B) edge (E);
	\path[-, line width=2pt] (C) edge (F); \path[-, line width=2pt] (C) edge (G);

	\path[-] (D) edge (H); \path[-] (D) edge (I);
	\path[-] (E) edge (K); \path[-] (E) edge (J);
	\path[-] (F) edge (L); \path[-] (F) edge (M);
	\end{tikzpicture}
}
}
\caption[]{Example of reusing of totalizers of formula $\phi$ for some bucket $B_r$, when only one leaf changes
  every time.  In this example, $N=8$ and we assume that all variables $y_{i,r}$ are present in $B_r$, for $i \in 1..8$. The  tree on the top is never created but illustrates
  the virtual totalizer where no leaf node is missing. The nodes (variables)
  and edges (clauses) in boldface appear only in one totalizer, and the others are reused in most totalizers.}
\label{fig:glpw_reuse_totalizer}
\end{figure}

Once the constants are moved to the output of $\psi$,  for any two formulas $PW(P[x_l,X_i],\ {\cal X}\setminus \{X_i\})$, $PW(P[x_{l'},X_j],\ {\cal X}\setminus \{X_j\})$ with $X_i \neq X_j$, their corresponding buckets  $B_r$ for a bit $r$ only differ in one variable, that is $y_{j,r}$ in the first formula and $y_{i,r}$ in the second. From the perspective of a totalizer, this means that only one leaf node changes between the trees of the two totalizers for $PW(P[x_l,X_i],\ {\cal X}\setminus \{X_i\})$ and $PW(P[x_{l'},X_j],\ {\cal X}\setminus \{X_j\})$. Therefore, we can reuse most of the nodes of the totalizers, and their associated auxiliary variables, when constructing the formulas. This is illustrated in Figure~\ref{fig:glpw_reuse_totalizer}, where we consider a PB(AMO) with $N=8$, and hence GLPW requires introducing 8 totalizers of 7 leaf nodes. We can see that the proposed implementation requires 30 distinct totalizer nodes in total, while a na\"ive implementation without reusing nodes requires 104 distinct nodes.

With the proposed implementation of GLPW, the number of required auxiliary variables is $O(nNlog(q_{max}))$, and the number of clauses is $O(nN^2log(q_{max}))$. This follows from the fact that we are  only introducing one formula $\phi$ for each $i \in 1..N$ and for each bit $ r \in 0..p$. This reduces the total asymptotic size of formulas $\phi$, but not the one of formulas $\psi$ which become the dominating components of this size.

\subsubsection{Reuse of Auxiliary Variables}\label{sec:glpw_dynamic}
Now we describe how to reuse auxiliary variables and clauses among the different generalized polynomial watchdog formulas~\eqref{eq:glpw_pw}. The degree of reusability is especially high in the polynomial watchdog formulas generated for variables belonging to the same group $X_i$. To a minor degree, we can also reuse parts of generalized polynomial watchdog formulas related to variables belonging to distinct groups. We maintain three maps (\emph{MY}, \emph{MT}, \emph{MM}) that store variables that can be reused if needed.

\paragraph{Map \emph{MY}} This map is used to store  the variables $y_{i,r}$ that are defined by clauses~\eqref{eq:pw_y_def}, but avoiding having to introduce equivalent auxiliary variables. The key to an entry of  $MY$ is the set of variables $V$ that logically imply $MY(V)$ according to clauses~\eqref{eq:pw_y_def}. The following example shows a case in which variables $y_{i,r}$ are reused.
\begin{example}\label{ex:glpw-var-reuse}
  Consider a set $X_i \in {\cal X}$ such that $X_i=\{x_1,x_2,x_3\}$
  and the associated coefficients are $q_1=3,q_2=7,q_3=11$. The binary
  representations of the coefficients are $0011$, $0111$, and $1011$,
  respectively. We need at least 4 bits to represent these numbers,
  and therefore we need 4 buckets.  Looking at the bit of most weight
  ($r=3$), the set of variables which have $b_r(q)=1$ is just
  $\{x_3\}$. In this case, since it is only one variable, no new variable
  is added nor clauses ~\eqref{eq:pw_y_def}, and we set
  $MY(\{x_3\})=x_3$.  Similarly, looking at the following bit $r=2$,
  we set $MY(\{x_2\})=x_2$. For $r=1$ we introduce an auxiliary variable
  $y_{i,1}$, add clauses $\noo{x_1} \lor y_{i,1}$,
  $\noo{x_2} \lor y_{i,1}$ and $\noo{x_3} \lor y_{i,1}$, and define
  $MY(\{x_1,x_2,x_3\})=y_{i,1}$. For the least significant bit $r=0$,
  we can reuse variable $MY(\{x_1,x_2,x_3\})$ instead of introducing variable $y_{i,0}$ and the corresponding extra clauses.
  \end{example}

\paragraph{Map \emph{MT}} This map is used to implement the totalizers of formulas $\phi$, avoiding building a subtree twice with the same leaf nodes.
Given as key a set of variables corresponding to leaves of a totalizer, it returns the variables of the root of the totalizer. 

\paragraph{Map \emph{MM}} This map is used to store the output formulas $\psi$, i.e., the vectors $S_r$,  and reuse them when possible. The key of an entry is a triplet $(S_{r-1},c,U(\langle B_r\rangle))$, where:
\begin{itemize}
	\item $S_{r-1}$ is the vector of output variables of formula $\psi$ for the previous bit.
	\item $c=1$ if $S_{r-1}^{1/2}$ was defined considering
          $b_{r-1}(T)=1$, and 0 otherwise (see
          Figure~\ref{fig:glpw_move1}). That is, $c=1$ if the left
          input of $\psi$ are the even positions of $S_{r-1}$, and
          $c=0$ if the odd positions must be taken. See for instance that circuits $\phi$ and $\psi$ are identical in Figure~\ref{fig:glpw_move1_a} (right) and Figure~\ref{fig:glpw_move1_b}. The presence or not of a $1$ constant (i.e., of $c$) determines which bits $S_r^{1/2}$ (either odd or even) are taken as input of the next formula $\psi$ for bit $r+1$.
	\item $U(\langle B_r\rangle))$ is the output of $\phi(\langle B_r\rangle)$, without including the 1 constant if present.
\end{itemize}
Note that an entry of $\mathit{MM}$ contains all the bits of $S_r$.
 The choice of even or odd bits from $S_r$ used as
  input of $\phi$ for bucket $r+1$ can be different at each
  generalized local polynomial watchdog formula.

It is natural to ask whether preprocessing could achieve the same or similar reuse of variables and clauses when given a na\"ive GLPW encoding. None of the common subexpression elimination algorithms implemented in Savile Row~\cite{aij-savilerow}, which is the constraint reformulation tool that we use in our experiments, would extract the common subtrees because the auxiliary variables have distinct names in each subtree. In SAT preprocessing, Equivalent Literal Substitution (ELS) \cite{van2005toward} examines the set of binary clauses and detects sets of literals that take the same truth value in all solutions. 
The variables reused through maps \textit{MY}, \textit{MT}, and \textit{MM} do not necessarily meet the definition of equivalence. Consider the variables \(y_{i,1}\) and \(y_{i,0}\) in Example \ref{ex:glpw-var-reuse}: when all \(x\) variables are false, \(y_{i,1}\) and \(y_{i,0}\) are free and so can take different values, therefore they are not equivalent literals.


\section{Propagation Properties}\label{sec:pp}

In this section we review the principal propagation properties that SAT encodings can have, and we describe the propagation characteristics of the new generalized encodings. In summary, all the presented PB(AMO) encodings preserve the propagation strength of their counterpart PB encodings. 

\emph{Unit propagation} (UP) is the main propagation mechanism used in modern
SAT solvers. It is based on the principle that if a clause contains a
single literal (i.e., under a given assignment, all literals but one
are false), then every model must make that literal true. Hence, the
assignment can be extended with this literal. 
The principal consistency notions that a SAT encoding $E$ of a constraint $C$ can achieve with UP are the following:
\begin{itemize}
	\item  $E$ is said to be \emph{consistency checker} (CC) when: given any partial assignment $A$, if $A$ cannot be extended to a model of $C$,
	then unit propagating $A$ on $E$ will falsify some clause.
	\item  $E$ is said to be \emph{generalized arc consistent} (GAC) when: given any partial assignment $A$, if a variable $x$ in $C$ is
	true (respectively false) in every extension of $A$ satisfying $C$,
	then unit propagating $A$ on $E$ will extend $A$ to $A \cup \{x\}$
	(respectively $A\cup\{\noo{x}\}$).
\end{itemize}
Although we do not consider them in this work, it is worth mentioning that there also exist stronger consistency notions that do not relate to the encoded constraint $C$ but are defined on all variables of $E$, including auxiliary variables:
\begin{itemize}
	\item A formula $E$ is said to be \emph{unit refutation complete} (URC) when: given any partial assignment $A$ that cannot be extended to a model of $E$, then unit propagating $A$ on $E$ will falsify some clause.
	\item A formula $E$ is said to be \emph{propagation complete} (PC) when: given any partial assignment $A$, if a variable $x$ in $E$ is
	true (respectively false) in every extension of $A$ satisfying $E$,
	then unit propagating $A$ on $E$ will extend $A$ to $A \cup \{x\}$
	(respectively $A\cup\{\noo{x}\}$).
\end{itemize}
Note that any GAC encoding is also CC, any PC encoding is both URC and GAC, and any URC encoding is also CC. 

When clear from the context, these properties are usually attributed to the encoding methods. For instance, we say that the MDD encoding is GAC, meaning that all formulas obtained with the MDD encoding method are GAC, and we say that GMTO is not CC, meaning that there exist PB(AMO) constraints whose encoding obtained with the GMTO encoding method is not CC. 
\begin{table}[t]
	\caption{Propagation strength of the different encodings.}
	\centering
\begin{tabular}{|c|c|c|c|c|c|}\hline
	 \bf BDD & \bf SWC & \bf GT/RGT & \bf MTO & \bf GPW & \bf LPW \\
	  \bf MDD & \bf GSWC & \bf GGT/RGGT & \bf GMTO & \bf GGPW & \bf GLPW \\\hline
	  GAC & GAC & GAC & - & CC & GAC \\ \hline
\end{tabular}
\label{table:pp}
\end{table}

The propagation strength of the encodings considered in this paper is summarised in Table~\ref{table:pp}. Previous works have already proved the propagation strength for encoding PB constraints of  BDD~\cite{abio2012new}, SWC~\cite{holldobler2012compact}, GT~\cite{joshi2015generalized}, and GPW and LPW~\cite{BailleuxBR09}.
Regarding RGT, it can be easily proved that it is also GAC using the same reasoning that was applied in the GAC proof of GT in~\cite{joshi2015generalized}. The key idea of the proof is that UP propagates to a parent node the sum of the values reached by the two children given the current assignment. This also happens in RGT, where \Cref{lem:rggt-incl} ensures that any sum value is always represented by a corresponding interval.
\begin{theorem}
	RGT is a GAC encoding of PB constraints.
\end{theorem}

We are however interested in the propagation strength of encodings of PB(AMO) constraints. There are many GAC encodings of AMO constraints, for instance the pairwise encoding~\cite{HandbookOfSAT2009}. Since we consider monotonic PB(AMO) constraints, it is straightforward to see that the previous PB encodings can be used in conjunction with encodings of AMO constraints to obtain PB(AMO) encodings with the same propagation strength.
\begin{lemma}
	Let $\cal P$ be a PB(AMO) constraint of the form
	$P\land M_1 \land \dots \land M_N$. Let $E$ be any GAC (respectively CC) encoding of $P$. Then the conjunction of
	$E$ with a GAC (resp.\ CC) encoding of
	$M_1 \land \dots \land M_N$ is a GAC (resp.\ CC) encoding of~$\cal P$.
\end{lemma}
The generalized PB(AMO) encodings considered in this paper maintain the propagation properties of their counterpart non-generalized encodings. 
This is proved in Theorem~\ref{th:others_gac}. The intuition of its proof is that each one of the formulas obtained with MDD, GSWC, GGT, RGGT, GGPW and GLPW for an input $(P,{\cal X})$ can be roughly seen as a union of many encodings obtained respectively with BDD, SWC, GT, RGT, GPW and LPW. In particular, they encode all possible PB constraints obtained by keeping in $P$ just one variable for each $X_i$. For instance, $GSWC(q_1x_1+q_2x_2+q_3x_3+q_4x_4\leq K, \{\{x_1,x_2\},\{x_3,x_4\}\})$ is roughly an efficient union of $SWC(q_1x_1+q_3x_3\leq K)$, $SWC(q_1x_1+q_4x_4\leq K)$, $SWC(q_2x_2+q_3x_3\leq K)$ and $SWC(q_2x_2+q_4x_4\leq K)$, that shares auxiliary variables and clauses. Therefore, GSWC contains all the required clauses to enforce GAC on the possible PB constraints that satisfy the AMOs.
\begin{theorem}\label{th:others_gac}
	Let $\cal P$ be a PB(AMO) constraint of the form
	$P\land M_1 \land \dots \land M_N$, where ${\cal X}=\{X_1,\dots,X_N\}$
	is a partition of the variables in $P$ such that
	$X_i = \mathit{scope(M_i)}$. The following hold:
	\begin{itemize}
		\item Let $E$ be any encoding among MDD, GSWC, GGT, RGGT and GLPW. Then the conjunction of
	$\mathit{E}(P,{\cal X})$ with a GAC encoding of
	$M_1 \land \dots \land M_N$ is a GAC encoding of $\cal P$. 
		\item The conjunction of
	$\mathit{GGPW}(P,{\cal X})$ with a CC encoding of
	$M_1 \land \dots \land M_N$ is a CC encoding of $\cal P$. 
	\end{itemize}
\end{theorem}
\begin{proof}
	We prove the theorem for GSWC, but the proof for the other encodings is analogous. Let $S$ denote the conjunction of $\mathit{GSWC}(P,{\cal X})$ with
	a GAC encoding of $M_1 \land \dots \land M_N$.
	Let~$A$ be a partial assignment to the variables of $S$ which is
	extendible to a satisfying assignment of $\cal P$. Therefore, no AMO
	constraint $M_i$ is violated under $A$. We need to show that for
	every variable $x$ of $\cal P$ such that $x$ is not assigned in $A$,
	if $A \cup \{x\}$ cannot be extended to a satisfying assignment of
	$\cal P$, then $x$ is set to false by unit propagating $A$ on $S$
	(note that $A \cup \{\noo{x}\}$ can always be extended to a
	satisfying assignment due to decreasing monotonicity, so we don't need to consider this case).
	W.l.o.g., assume that $x_1 \in X_1$ is such variable. If
	$A \cup \{x_1\}$ cannot be extended to a satisfying assignment of
	$M_1 \land \dots \land M_N$ then, by the assumption that $S$
	contains a GAC encoding of
	$M_1 \land \dots \land M_N$, we have that $x_1$ is set to false by
	unit propagation. Assume now the contrary, i.e., that
	$A \cup \{x_1\}$ can be extended to an assignment satisfying the
	AMOs. In this case, the reason why UP should set $x_1$ to false is
	that $A \cup \{x_1\}$ cannot be extended to satisfy $P$. Since
	$A \cup \{x_1\}$ does not violate $M_1\land\dots\land M_N$, at most
	one variable in $X_i$ is true in $A$, for $2\leq i \leq N$, and no
	variable in $X_1$ is true in $A$.  Let us construct a PB constraint
	$P'$ from $P$ by picking one variable $x_{j_i}$ from each set $X_i$,
	$2 \leq i \leq N$, as follows: if $X_i$ contains a variable which is
	true in $A$, then this is the variable to be picked up from $X_i$,
	otherwise pick up any variable.  We define
	$P': q_1x_1 + \sum_{i=2}^N{q_{j_i}x_{j_i}} \leq K$.  Since $P'$
	contains all variables of $P$ which are true in $A$, and due to the
	monotonicity of $P$, we have that
	$q_1x_1 + \sum_{i=2}^N{q_{j_i}x_{j_i}}\leq K$ is equisatisfiable to
	$\sum_{i=1}^n{q_ix_i}\leq K$ under the assignment $A \cup \{x_1\}$,
	and therefore $A \cup \{x_1\}$ cannot be extended to a model of
	$P'$.  It is not hard to see that $\mathit{GSWC}(P,{\cal X})$
	contains all clauses of $\mathit{SWC}(P')$. Since the SWC encoding is GAC, $S$ contains all the clauses required to set $x_1$ to false by UP.
\end{proof}
An alternative proof that MDD is GAC was presented in~\cite{bofill2020mdd}.
Note that LPW as well as GLPW are the only GAC encodings of polynomial
size considered in this paper. The other ones have pseudo-polynomial size complexity.

It was stated in~\cite{BailleuxBR09} that the GPW encoding is not GAC. Since GGPW is a generalization of GPW, it is also not GAC. For instance, one could consider a PB(AMO) constraint where all AMO constraints have size 1. In this case, GPW and GGPW are identical. The same happens with MTO, which is also not GAC as stated in~\cite{zha2019n}, and therefore GMTO is not GAC either. In fact, MTO is not even a CC encoding of PB constraints, and therefore GMTO is not a CC encoding of PB(AMO) constraints.

\begin{theorem}
	MTO is not a CC encoding of PB constraints.
\end{theorem} 
\begin{proof}
	Consider the PB constraint and MTO encoding of Figure~\ref{fig:mto}. 
	Let partial assignment $A=\{x_6,x_8\}$ (i.e.\ $A=\{m_1^1,o_2^1\}$). Clearly $A$ cannot be extended to a model of $P$ since $4x_6+8x_8>10$.
	The only generated clauses containing $m_1^1$ and $o_2^1$ are:
	$$\eqref{eq:mto_sumnocarry_1}:\red{\noo{l_0^1}} \lor \noo{m_1^1} \lor f_1^1 \lor \red{\gamma_F^1}
	\qquad \eqref{eq:mto_sumnocarry_1}: \red{\noo{n_0^1}} \lor  \noo{o_2^1} \lor g_2^1 \lor \gamma_G^1 
	\qquad \eqref{eq:mto_sumnocarry_2}: \noo{n_1^1} \lor \noo{o_2^1} \lor \gamma_G^1$$
	where we paint in red the literals of type $\gamma^h_O$, $l_0^h$ and $r_0^h$ which are trivially false and are not included in these clauses by construction. Similarly we do not include clauses of type~\eqref{eq:mto_sumcarry_1} to~\eqref{eq:mto_sumcarry_3} which are  satisfied due to $\gamma_O^{h-1}$ being trivially false. 
	
	Unit propagation only assigns literal $f_1^1$, due to clause $\noo{m_1^1} \lor f_1^1$. The only remaining clauses containing variable $f_1^1$ are:
	$$\begin{array}{ll}
		\eqref{eq:mto_sumnocarry_1}: \noo{f_1^1}\lor \red{\red{\noo{g_0^1}}} \lor  c_1^1 \lor \gamma_C^1 
		&\quad \eqref{eq:mto_sumcarry_1}: \noo{\gamma_C^0} \lor \noo{f_1^1}\lor \red{\red{\noo{g_0^1}}} \lor  c_2^1 \lor \gamma_C^1  \\[4pt]
		\eqref{eq:mto_sumnocarry_1}:\noo{f_1^1} \lor \noo{g_1^1} \lor c_2^1 \lor \gamma_C^1 
		& \quad \eqref{eq:mto_sumcarry_2}:\noo{\gamma_C^0} \lor \noo{f_1^1} \lor \noo{g_1^1} \lor \gamma_C^1\\[4pt]
		\eqref{eq:mto_sumnocarry_2}: \noo{f_1^1} \lor \noo{g_2^1} \lor \gamma_C^1 
		&\quad \eqref{eq:mto_sumcarry_2}:\noo{\gamma_C^0} \lor \noo{f_1^1} \lor \noo{g_2^1} \lor \gamma_C^1 \quad \eqref{eq:mto_sumcarry_3}: \noo{\gamma_C^0} \lor \noo{f_1^1} \lor \noo{g_2^1} \lor c_1^1 
	\end{array}$$
	No clause is falsified and no other literal is unit-propagated.
\end{proof}
\begin{corollary}
	The conjunction of GMTO with an encoding of the AMO constraints is not a CC encoding of PB(AMO) constraints.
\end{corollary}
Although GMTO is the encoding with the worst propagation properties among all the encodings
considered in this paper, in Section~\ref{sec:results} it can be observed that it produces the smallest formulas by far in the
selected benchmark sets. As a result, the GMTO encoding ends up providing
the best performance for some benchmarks.


\section{Experiments}\label{sec:results}
In this section we report on a comparison between the different
encodings for PB(AMO) constraints, and also between those and the
classical encodings for PB constraints.  
For this purpose, we solve problems containing PB constraints as well as sources of incompatibility between
their variables, i.e. mutexes. We provide empirical evidence of the
usefulness of taking into account existing AMO constraints when
encoding PB constraints. We show that all new PB(AMO) encodings
perform significantly better than their counterpart PB encodings,
based on executions of two different SAT solvers which are
representative of the state-of-the-art. We also show the good
performance of the reduced generalized totalizer encodings (RGT /
RGGT). We study in detail the impact of the RGGT reduction algorithm
on the size of the generalized totalizers.

\subsection{Experimental Setting}\label{sec:res_setting}

On one hand we consider two problems consisting essentially of conjunctions of PB constraints and AMO constraints: the Multi-Choice
Multidimensional Knapsack Problem (MMKP) and Combinatorial Auctions (CA). For these problems, the AMO constraints between Boolean variables are explicitly stated in the problem definition, and hence we can directly define the PB(AMO) constraints, as we describe in Subsections~\ref{sec:mmkp} and~\ref{sec:ca}.

On the other hand we consider three challenging problems that are not essentially a set of PB(AMO) constraints but where PB constraints play an important role as well. Namely, we consider two extensions of the highly studied Resource-Constrained Project Scheduling Problem (RCPSP): Multi-mode RCPSP (MRCPSP)~\cite{brucker99} and RCPSP with Time-Dependent Resource Capacities and Requests (RCPSP/t)~\cite{hartmann2013project}.
We also consider the Nurse Scheduling Problem (NSP)~\cite{NSPLib}. These three problems have been modelled with the constraint programming modelling language Essence Prime~\cite{DBLP:journals/corr/NightingaleR16}. We automatically detect AMO constraints and generate the SAT formulas using Savile Row~\cite{savilerow-aij}, as described in~\cite{AnsoteguiBCDEMN19}. In brief, Savile Row detects pairs of Boolean variables that cannot be true at the same time (mutexes) by using constraint programming propagation schemes, and then builds disjoint AMO constraints from those mutexes. The detected AMO constraints are translated to SAT with the 2-product encoding~\cite{chen2010new}. In Subsections~\ref{sec:mrcpsp}, \ref{sec:rcpspt} and~\ref{sec:nsp} we provide a small description of such problems together with a discussion on the source of the mutexes, which in most cases are implicit, i.e.\ not stated by an explicit constraint in the model.

We have chosen MMKP, which is essentially a set of PB(AMO) constraints, in order to craft three different benchmark sets with different parameters,
with the aim of showing which encodings are better suited for
different kinds of PB(AMO) constraints (e.g.\ with different numbers of variables, coefficient values or AMO sizes). For the other problems, we have used representative benchmarks from the literature. We have considered the decision version of CA, MRCPSP, RCPSP/t and NSP, which are optimisation problems. This means that for maximisation problems (CA), and respectively minimisation problems (the others), we set a lower bound (resp. upper bound) on the objective function. In order to obtain both satisfiable and unsatisfiable instances, we consider each instance twice with two different bounds: a bound equal to the best known objective (in most cases the optimum), and a bound equal to the best plus one (in CA) or minus one (in the others). Table~\ref{table:sets} summarises the properties of each benchmark set. 

Overall, we consider all the problems that have been studied in the previous works related to PB(AMO) constraints~\cite{AnsoteguiBCDEMN19,bofill2020mdd}. All of them are NP-hard.
In each case the AMO constraints are encoded using only binary clauses and a small number of additional variables. The AMO encodings are invariant when comparing the various PB and PB(AMO) encodings, so the clauses and variables of the AMO encodings are not included in the reported formula sizes.

\begin{table}

	\caption{Summary of each set, containing in this order: number
          of instances; approximate number of PB constraints in each
          instance; approximate average number or range of number of variables
          in a PB constraint; average size of the AMOs; range of values of the
          coefficients.
        }
	\centering
	\begin{tabular}{|l|c|c|c|c|c|} \cline{2-6}
		\multicolumn{1}{c|}{} & \bf $\vert$set$\vert$ & \bf PB count & \bf PB size  & \bf AMO size & \bf coef. \\\hline
		\bf MMKP1 & 500 & 10 & 150 & 10 & [1,1000] \\ \hline
		\bf MMKP2 & 500 & 10 & 150 & 10 & [1,60] \\ \hline
		\bf MMKP3 & 400 & 50 & 75  & 5 & [1,10] \\ \hline
		\bf CA & 340 & 1 & [70,200]& 8 & [1,2500] \\ \hline
		\bf MRCPSPj30 & 1004 & 68 &29& 4 & [1,10]\\ \hline
		\bf RCPSP/Tj120 & 720 & 624 & 158 & 9 & [1,10] \\ \hline
		\bf NSP & 400 & 1 & 700 & 4 & [1,4] \\ \hline
	\end{tabular}
\label{table:sets}

\end{table}

\subsubsection{MMKP Instances}\label{sec:mmkp}
These have been generated using the MMKP instance generator from~\cite{HAN2010172}.
Each instance is defined by four
parameters: $L$ is the number of PB constraints, $N$ is the number of
AMO constraints, $M$ is the number of Boolean variables in each AMO
constraint, and $Q$ is the maximum coefficient of a variable in a PB
constraint.  The variables of the AMO constraints are disjoint, so
there are a total of $n=N\cdot M$ Boolean variables in each instance.
The PB constraints contain all $n$ variables.  The $j$-th variable in
the $i$-th AMO constraint is named $x_{i,j}$.  The coefficients in the
PB constraints are generated uniformly and independently at random in
the range $[1,Q]$.
The resulting instance has the following constraints:
\begin{align}
 &\qquad\sum_{i=1}^N{\sum_{j=1}^M{q_{i,j,k} \cdot x_{i,j}}} \leq K_k & 1\leq k \leq L \label{eq:mmkp_pb}\\
 &\qquad\sum_{j=1}^M{x_{i,j}} \leq 1 & 1\leq i \leq N\label{eq:mmkp_amo} \\
 &\qquad\sum_{j=1}^M{x_{i,j}} \geq 1 & 1\leq i \leq N\label{eq:mmkp_alo}
\end{align}
The conjunction of PB and AMO constraints~\eqref{eq:mmkp_pb}
and~\eqref{eq:mmkp_amo} is not a hard problem, since a trivial
solution is to set all variables $x_{i,j}$ to 0. For this reason
we add \emph{at-least-one} constraints~\eqref{eq:mmkp_alo}, requiring
that at least one variable in each AMO group is set to
true.  This way, \eqref{eq:mmkp_amo} together
with~\eqref{eq:mmkp_alo} form an \emph{exactly-one} constraint. When generating the SAT formulas, constraint~\eqref{eq:mmkp_amo} is encoded with the Regular encoding~\cite{ansotegui2004mapping}. 

We provide three different benchmark sets with different parameters.
The instances in a benchmark set
are distributed in families, and every family has values of $K_k$
distributed uniformly at random around a different mean in the range $[1,M\cdot Q]$.
The values of $K_k$ are proportional to the values of the coefficients in order to avoid introducing trivially satisfiable PB constraints.
We choose different values of $K_k$ to ensure that in the benchmark sets there are
instances of different hardness, and that approximately
half of the instances are satisfiable.
\begin{description}
\item[\emph{MMKP1}] 100 families of 5 instances, with $L=10$, $N=15$,
  $M=10$, $Q=1000$. The families have linearly increasing $K_k$ values 
  from family 1 (capacities of about 1000) to family 100 (capacities
  of about 14000).
  \item[\emph{MMKP2}] 100 families of 5 instances, with $L=10$, $N=15$,
  $M=10$, $Q=60$. The families have linearly increasing $K_k$ values
  from family 1 (capacities of about 100) to family 100 (capacities
  of about 800).
\item[\emph{MMKP3}] 20 families of 20 instances, with $L=50$, $N=15$,
  $M=5$, $Q=10$. The values of $K_k$
  linearly increase in each family, ranging between 65 and 100.
\end{description}
As can be seen in Table~\ref{table:results_size}, the three benchmark sets are diverse regarding the sizes of the formulas generated by all encodings. Namely, formulas obtained from MMKP1 are approximately one order of magnitude larger than those obtained from MMKP2, and these are approximately one order of magnitude larger than the formulas obtained from MMKP3.

\subsubsection{CA Instances}\label{sec:ca}
In the CA problem, there is a number of sets of items which are demanded (bids). 
Each bid has an associated
profit, and some bids are incompatible, i.e.\ they cannot be selected
together because they contain a same item. The problem consists of selecting a subset of bids to
maximise the obtained profit. This problem can be naturally modelled
with a Boolean variable for each bid: a variable is set to true if and
only if its associated bid is selected; AMO constraints appear when
setting incompatibilities between bids; finally, the maximisation of
the profit is modelled as a pseudo-Boolean objective function.
The AMO groups are constructed using the following greedy procedure:
we start with an empty list that will hold disjoint sets of bids and
we process each bid $b$ in turn; $b$ is added to the first
set in which all existing bids are pairwise incompatible with $b$; if
no such set exists, a new set containing $b$ is added to the list.

We consider the set of instances presented by Bofill et al~\cite{BofillPSV14}, which were
generated with the Combinatorial Auctions Test
Suite~\cite{leyton2006test}. There are a total of 170 optimisation
instances, each one containing a set of between 70 and 200 bids, and a
subset of these bids must be selected. As explained before, from each instance we have defined two instances of the decision problem with two different lower bounds, and hence there is a total of 340 instances.

\subsubsection{MRCPSP instances}\label{sec:mrcpsp}
The MRCPSP consists of deciding a start time and an execution mode for each of the activities of a project. These activities have demands of arbitrary quantities on shared resources of limited capacity, and it must be ensured that those capacities are never surpassed. These constraints can be naturally modelled with PB constraints. Also, there are predefined end-start precedence relations between activities that must be respected. Finally, the duration and demands on resources of each activity depends on the selected execution mode, and just one mode must be chosen. An optimal solution must minimise the total duration of the project. We consider the time-indexed model from~\cite{AnsoteguiBCDEMN19}, where Boolean variables express whether an activity is running in a particular mode at a particular time instant. There are mainly three reasons of pair-wise incompatibility between these variables: precedences between activities, the requirement of single-mode selection, and the limited capacity of the resources.

For this problem, we consider the 552 feasible instances of the j30 set from PSPLib~\cite{kolishPSPLIB}, each of them with two upper bounds. We refer to this set as MRCPSPj30.

\subsubsection{RCPSP/t instances}\label{sec:rcpspt}
RCPSP/t is another extension of RCPSP. Unlike MRCPSP, there is just one execution mode per activity. However, the demands of an activity over each resource can change during its execution, i.e.\ the demand depends on how many time units have passed since the activity started.
A natural way to deal with this characteristic is by introducing Boolean variables stating whether an activity has started at a particular time instant, and express PB resource constraints in terms of these variables~\cite{bofill2020smt}. Therefore there is a new source of pair-wise incompatibilities in addition to precedences and limited resource capacities, that is the fact that an activity cannot start at two different time instants.

For this problem, we consider a representative subset of 360 instances
out of the 3600 instances of set j120
from~\cite{hartmann2013project}. The J120 set is composed of 360 families of 10 instances with similar characteristics, and from every family we have chosen the first one. Again, each instance is considered with two upper bounds. We refer to this set as RCPSP/Tj120.

\subsubsection{NSP instances}\label{sec:nsp}
The NSP is the problem of deciding the daily shifts of nurses according to their preferences. Many variants of this problem have been proposed~\cite{DeCausmaecker2011,NSPLib}. Here we consider the basic definition that was also used in~\cite{AnsoteguiBCDEMN19} as well as the same Essence Prime model and set of instances. In particular, the considered problem consists in assigning a shift for each nurse and day, while ensuring that each shift has enough nurses and that nurses do not work too many days in a week. Also, the nurses have some preferences regarding their shift assignments, and there is a penalisation for each preference that is not satisfied. The sum of such penalisations must be minimised. For the decision version of the problem, this objective function becomes an upper bound on the total penalisation, which can be modelled with a PB constraint. AMO constraints appear since each nurse can only work in one shift each day. 

The instances we solve are a set of 200 instances taken at random from the N25 set from NSPLib~\cite{NSPLib}. Again, each instance is solved with two upper bounds.

\subsection{Comparison Between PB and PB(AMO) Encodings}\label{sec:res_pbVSpbamo}

All instances of all sets have been encoded into SAT
using the PB(AMO) encodings introduced in this paper, as well as using
the corresponding original PB encodings.  For completeness we also
report results on the BDD- and MDD-based encodings
from~\cite{abio2012new} and~\cite{bofill2020mdd} summarised in
Section~\ref{sec:enc_bdd}.

The generated SAT formulas have been solved with a timeout setting of
600 seconds using two different SAT solvers. The first solver is
CaDiCaL~\cite{Biere-SAT-Race-2019-solvers}, which was the system that
solved more instances than any other competitor in the SAT Race
2019~\cite{heule2019satrace}. The second solver is Glucose version
4.1~\cite{audemard2018glucose}, which has also obtained top positions
in previous SAT competitions and is currently the core of many other
state-of-the-art solvers such as MapleLCMD. All formulas have been
solved on the same machine: an 8GB, 3.10GHz
Intel\textsuperscript{\textregistered}
Xeon\textsuperscript{\textregistered} E3-1220v2.

\subsubsection{Formula Size Comparison}

Table~\ref{table:results_size} contains the sizes of the
encodings of PB and PB(AMO) constraints.
For each benchmark set and for each
encoding we report, in thousands, the mean number of variables
(\emph{vars.}) and clauses (\emph{cl.}) required to encode one PB or
PB(AMO) constraint.  We omit the number of variables and clauses
required to encode the AMO constraints because it is the same for each
encoding and negligible in magnitude. Column \emph{g.t.}\ contains the
mean computation time required to generate the SAT
encoding of the instances of the benchmark set. A long dash (---) means that
the encoding has been discarded for that benchmark set, because the formulas
are too large and the generation of most instances either ran out of
memory or did not finish in less than 600 seconds. 
We highlight in boldface the smallest number of variables and clauses among all encodings for each benchmark set and also distinguishing between PB and PB(AMO) encodings.

In all sets, using PB(AMO) encodings lets us reduce both the number of clauses and the number of
variables w.r.t.\ their counterpart PB encodings. The reduction rate ranges from one half to three orders of magnitude. It is noticeable that we obtain a high
reduction even in the sets with the smallest AMO constraints, with only 4 and 5 variables per AMO. The
decrease in size also clearly affects positively the generation time in MMKP and CA sets, which is at least
halved in most encodings. However the sets where formulas are generated with Savile Row have generally higher generation times, and the difference between PB and PB(AMO) encodings is not always significant. This is because the automatic AMO detection requires more time than the ad-hoc methods used in CA and MMKP. The generation times are particularly large in RCPSP/Tj120, where a huge number of mutexes are detected.

The use of PB(AMO) encodings is crucial in some cases. For instance,
with GSWC and GLPW in MMKP1 we are able to generate instances that run
out of memory with their counterpart encodings SWC and LPW. Similarly,
with RGGT in MMKP1 we obtain a reasonable generation time compared to
the one of RGT.

We also observe that GGPW and GMTO, which digit-wise decompose the
coefficients of the PB, produce dramatically smaller sizes compared to the other encodings.  This difference is of approximately two orders of magnitude in the number of clauses and variables in MMKP1 and CA. 
This is because these benchmark sets have large coefficients (and also large $K$) and this fact penalises the encodings with a size  proportional to the value of $K$. 
Although LPW and GLPW also use a digit-wise decomposition, they produce significantly larger encodings than GGPW and GMTO. This is particularly noticeable in RCPSP/Tj120 and NSP, where the generated formulas
 are many orders of magnitude larger than those obtained with other encodings. This is because GLPW needs to encode many polynomial watchdog formulas. Nevertheless, GLPW generates smaller formulas than the other GAC encodings in CA and MMKP1, which have large coefficients. Recall that GLPW is the only GAC encoding that generates polynomial size formulas w.r.t.\ the size of the PB.
We have observed that the refinements of GLPW proposed in Section~\ref{sec:glpw_totalizer} and Section~\ref{sec:glpw_dynamic} are crucial, since a na\"ive implementation generates huge formulas in all sets, as is the case with LPW. 

Regarding the reduction process applied in RGT/RGGT, we observe that it generally reduces the sizes of the formulas w.r.t.\ GT/GGT encodings, in some cases halving the number of variables and clauses. 
Also, comparing the minRatio heuristic to the default with GT and GGT encodings, we observe that the minRatio heuristic produces smaller formulas than the default (i.e.\ GT is smaller than GTd, and GGT is smaller than GGTd) in terms of the number of clauses. 
For MMKP1 and CA, GT and GGT are able to generate formulas but GTd and GGTd are not. However, the minRatio encodings generally produce more variables.

\begin{table*}[!htbp]
	\caption{Number of variables and clauses in thousands and
          generation time in seconds, for each set and PB(AMO) encoding.}
      \scriptsize
	\centering
	\begin{tabular}{|c|l|r|r|r||l|r|r|r|}
		\cline{2-9}
		\multicolumn{1}{c|}{}& \multicolumn{4}{c||}{\bf PB}&\multicolumn{4}{c|}{\bf PB(AMO)}\\ \cline{2-9}
		\multicolumn{1}{c|}{}& \bf enc. &\bf vars.&\bf cl.& \bf g.t.& \bf enc. &\bf vars.&\bf cl.& \bf g.t.\\ \hline
		\multirow{9}{*}{\rotatebox{90}{MMKP1}}& \bf BDD&596.60&1193.20&19.84& \bf MDD&25.09&263.37&2.68\\ \cline{2-9}
		& \bf SWC&---&---&---& \bf GSWC&104.84&1072.04&5.97\\ \cline{2-9}
		& \bf GTd&---&---&---& \bf GGTd&---&---&---\\ \cline{2-9}
		& \bf GT&831.49&1805.75&88.71& \bf GGT&61.91&676.54&6.52\\ \cline{2-9}
		& \bf RGT&634.81&1411.53&116.13& \bf RGGT&25.00&275.01&4.53\\ \cline{2-9}
		& \bf MTO&\bf 3.56&\bf 10.30&0.17& \bf GMTO&\bf 0.49&\bf 1.95&0.09\\ \cline{2-9}
		& \bf GPW&5.90&76.93&0.48& \bf GGPW&0.99&4.43&0.04\\ \cline{2-9}
		& \bf LPW&---&---&---& \bf GLPW&30.52&220.56&2.09\\ \hline
		\hline
		\multirow{9}{*}{\rotatebox{90}{MMKP2}}& \bf BDD&40.60&81.20&1.10& \bf MDD&2.04&19.59&0.19\\ \cline{2-9}
		& \bf SWC&68.11&135.49&0.82& \bf GSWC&6.41&61.64&0.36\\ \cline{2-9}
		& \bf GTd&10.00&1639.65&10.47& \bf GGTd&1.93&120.67&0.82\\ \cline{2-9}
		& \bf GT&47.85&101.03&4.01& \bf GGT&4.15&41.65&0.36\\ \cline{2-9}
		& \bf RGT&37.34&80.00&3.99& \bf RGGT&2.01&20.94&0.24\\ \cline{2-9}
		& \bf MTO&\bf 2.40&\bf 7.15&0.08& \bf GMTO&\bf 0.33&\bf 1.19&0.02\\ \cline{2-9}
		& \bf GPW&3.48&42.33&0.26& \bf GGPW&0.59&2.46&0.03\\ \cline{2-9}
		& \bf LPW&---&---&---& \bf GLPW&13.26&90.08&0.80\\ \hline
		\hline
		\multirow{9}{*}{\rotatebox{90}{MMKP3}}& \bf BDD&3.27&6.55&0.37& \bf MDD&0.46&2.18&0.11\\ \cline{2-9}
		& \bf SWC&6.05&12.02&0.40& \bf GSWC&1.16&5.52&0.18\\ \cline{2-9}
		& \bf GTd&1.31&30.90&1.03& \bf GGTd&0.38&4.58&0.17\\ \cline{2-9}
		& \bf GT&4.27&8.78&1.10& \bf GGT&0.80&3.89&0.18\\ \cline{2-9}
		& \bf RGT&3.15&6.54&1.04& \bf RGGT&0.45&2.32&0.13\\ \cline{2-9}
		& \bf MTO&0.83&\bf 1.78&0.10& \bf GMTO&\bf 0.21&\bf 0.58&0.04\\ \cline{2-9}
		& \bf GPW&\bf 0.82&5.11&0.17& \bf GGPW&0.33&1.17&0.05\\ \cline{2-9}
		& \bf LPW&19.38&213.59&7.45& \bf GLPW&4.38&24.04&1.04\\ \hline
		\hline
		\multirow{9}{*}{\rotatebox{90}{CA}}& \bf BDD&155.96&312.06&0.44& \bf MDD&38.60&160.71&0.22\\ \cline{2-9}
		& \bf SWC&376.91&747.11&0.45& \bf GSWC&109.55&381.73&0.23\\ \cline{2-9}
		& \bf GTd&---&---&---& \bf GGTd&---&---&---\\ \cline{2-9}
		& \bf GT&166.92&458.77&1.07& \bf GGT&78.94&247.68&0.35\\ \cline{2-9}
		& \bf RGT&116.17&349.09&1.62& \bf RGGT&34.90&150.98&0.52\\ \cline{2-9}
		& \bf MTO&\bf 1.21&\bf 6.04&0.02& \bf GMTO&\bf 0.41&\bf 1.35&0.01\\ \cline{2-9}
		& \bf GPW&1.86&17.96&0.02& \bf GGPW&0.72&3.81&0.01\\ \cline{2-9}
		& \bf LPW&67.03&1327.85&0.84& \bf GLPW&16.60&154.37&0.12\\ \hline
		\hline
			\multirow{9}{*}{\rotatebox{90}{MRCPSPj30}}& \bf BDD&0.42&0.81&5.37&\bf MDD&0.06&0.21&5.33\\ \cline{2-9}
		&	\bf SWC&0.85&1.68&5.35&\bf GSWC&0.15&0.50&5.30\\ \cline{2-9}
		&	\bf GTd&0.18&1.60&5.46&\bf GGTd&0.06&0.41&5.35\\ \cline{2-9}
		&	\bf GT&0.50&1.03&5.70&\bf GGT&0.11&0.33&5.37\\ \cline{2-9}
		&	\bf RGT&0.34&0.73&5.68&\bf RGGT&0.05&0.21&5.36\\ \cline{2-9}
		&	\bf MTO&0.26&0.85&5.37&\bf GMTO&0.06&0.16&5.30\\ \cline{2-9}
		&	\bf GPW&0.28&1.32&5.35&\bf GGPW&0.08&0.21&5.30\\ \cline{2-9}
		&	\bf LPW&4.77&42.24&5.64&\bf GLPW&0.72&2.57&5.40\\ \hline
		\hline
		\multirow{9}{*}{\rotatebox{90}{RCPSP/Tj120}}& \bf BDD&2.61&5.19&35.52&\bf MDD&0.30&1.03&33.27\\ \cline{2-9}
		&\bf SWC&3.92&7.97&34.88&\bf GSWC&0.42&1.39&32.92\\ \cline{2-9}
		&\bf GTd&0.78&4.77&35.15&\bf GGTd&0.21&2.09&33.55\\ \cline{2-9}
		&\bf GT&2.44&4.99&81.93&\bf GGT&0.44&1.37&34.25\\ \cline{2-9}
		&\bf RGT&2.24&4.59&78.65&\bf RGGT&0.34&1.20&34.12\\ \cline{2-9}
		&\bf MTO&1.79&22.30&36.86&\bf GMTO&0.19&0.59&33.29\\ \cline{2-9}
		&\bf GPW&2.27&43.82&34.96&\bf GGPW&0.26&1.12&33.16\\ \cline{2-9}
		&\bf LPW&190.51&11269.63&60.49&\bf GLPW&5.51&46.33&35.41\\ \hline
		\hline
		\multirow{9}{*}{\rotatebox{90}{NSP}}& \bf BDD&113.28&226.31&2.03&\bf MDD&4.22&12.38&1.70\\ \cline{2-9}
		&\bf SWC&175.64&351.48&1.85&\bf GSWC&5.06&14.74&1.72\\ \cline{2-9}
		&\bf GTd&6.59&249.81&2.09&\bf GGTd&1.28&15.96&1.78\\ \cline{2-9}
		&\bf GT&143.97&288.45&14.30&\bf GGT&4.80&13.77&2.34\\ \cline{2-9}
		&\bf RGT&112.45&225.41&14.36&\bf RGGT&4.23&12.55&2.36\\ \cline{2-9}
		&\bf MTO&7.72&14.12&1.85&\bf GMTO&1.34&7.19&1.74\\ \cline{2-9}
		&\bf GPW&7.93&242.72&1.87&\bf GGPW&1.91&26.99&1.73\\ \cline{2-9}
		&\bf LPW&1180.69&101527.19&4.47&\bf GLPW&109.52&3961.55&2.35\\\hline
	\end{tabular}
	\label{table:results_size}
\end{table*}

\subsubsection{Solving Time Comparison}
Tables~\ref{table:results_cadical} and \ref{table:results_glucose} contain statistics regarding the solving times using the SAT solvers CaDiCaL and Glucose respectively. The tables contain, for each benchmark set and encoding: first quartile (\emph{Q1}), median (\emph{med}) and third quartile (\emph{Q3}) of solving time in seconds, where t.o.~means execution aborted at 600 seconds; and the
number of instances that timed out before being solved (\emph{t.o.}). We highlight in boldface the best encoding regarding number of timeouts, breaking ties with Q3, for each solver, for each benchmark set, and also distinguishing between PB and PB(AMO) encodings. The globally best values for each dataset considering both solvers are underlined.

Even though both are CDCL solvers, CaDiCaL incorporates many inprocessing techniques that modify the formula on the fly, as well as local search. These are not included in Glucose, and this might explain that in some cases they perform quite differently. In any case, the results clearly show that using  PB(AMO)s  substantially improves the solving times.
This solving time reduction is consistent with the observed reduction in the sizes of the formulas. 
In most cases the solving times are reduced by approximately one order of magnitude. Also the number of timeouts is significantly reduced, sometimes more than halved. 
There are cases where this improvement is even more evident, as in RCPSP/Tj120 with CaDiCaL, and especially in NSP where almost no solutions are found with original PB encodings with any solver.

Regardless of which solver is used, the GMTO encoding is clearly the best in MMKP1 and MMKP2, and is very close to the best encoding for CA, namely GGPW. As mentioned before, all encodings except GMTO and GGPW generate huge formulas for MMKP1 and CA, and large formulas for MMKP2. These large sizes adversely affect the solving times. Among GAC encodings, the best one in MMKP1 and CA is GLPW, which is among the smallest GAC encodings for those benchmark sets. 
For datasets with smaller coefficient values, i.e.\ MMKP3, MRCPSPj30, RCPSP/Tj120 and NSP, there is not a clear winner, although GGPW, RGGT and GMTO perform the best in many cases.

In all of our benchmark sets, GGPW is better than GLPW even though the latter is GAC. This is possibly because GLPW always generates significantly larger formulas.

Looking at the crafted MMKP datasets, instances in MMKP3 contain more PB constraints than the others, and the values of $K$ are distributed around the
transition value from unsatisfiable instances to satisfiable instances. We have observed empirically that it is in this transition where the instances become harder. It is precisely in MMKP3 where we can observe that non-GAC encodings worsen significantly their performance with respect to MMKP1 in comparison to GAC encodings, such as RGGT.

The picture regarding the minRatio heuristic is somewhat complicated. For MMKP1 and CA, it was not even possible to generate the formulas for GTd and GGTd. 
However, for MMKP2, MMKP3 and MRCPSPj30 with the PB encodings (comparing GT to GTd), the default heuristic is superior with both solvers. 
In this case minRatio causes GT to generate many more variables than GTd. In other datasets, there is no clear best option.
When using the PB(AMO) encodings, the sizes of GGT and GGTd are more similar than the sizes of GT and GTd.  
Overall it seems that the minRatio heuristic avoids catastrophic worst-case behaviour when the range of coefficients is large, but otherwise is not clearly better or worse than the default heuristic for the PB(AMO) encodings.

Finally, we observe that the reduction technique introduced in the 
RGT and RGGT encodings improves the solving times in all
MMKP and CA sets and with both solvers compared to GT and GGT respectively, and
the number of timeouts is also reduced in all cases.

\begin{table*}[!htbp]
	\caption{Solving times and number of timeouts using CaDiCaL.}
	\scriptsize
	\centering
	\begin{tabular}{|c|l|r|r|r|r||l|r|r|r|r|}
		\cline{2-11}
		\multicolumn{1}{c|}{}& \multicolumn{5}{c||}{\bf PB}&\multicolumn{5}{c|}{\bf PB(AMO)}\\ \cline{2-11}
		\multicolumn{1}{c|}{}& \bf enc. & \bf Q1 & \bf med& \bf Q3 & \bf t.o.&\bf enc. & \bf Q1 & \bf med& \bf Q3 &  \bf t.o.\\ \hline
		\multirow{9}{*}{\rotatebox{90}{MMKP1}}& \bf BDD&27.75&47.62&t.o.&149& \bf MDD&3.63&5.64&274.05&118\\ \cline{2-11}
		& \bf SWC&---&---&---&---& \bf GSWC&21.32&37.32&292.34&107\\ \cline{2-11}
		& \bf GTd&---&---&---&---& \bf GGTd&---&---&---&---\\ \cline{2-11}
		& \bf GT&48.86&65.47&t.o.&146& \bf GGT&14.60&19.83&314.57&116\\ \cline{2-11}
		& \bf RGT&31.93&49.30&t.o.&146& \bf RGGT&3.90&5.75&181.21&110\\ \cline{2-11}
		& \bf MTO&0.21& 0.97&7.75&\bf \ull{43}& \bf GMTO&0.03&0.07&1.03&\bf \ull{27}\\ \cline{2-11}
		& \bf GPW&1.79&2.08&24.87&68& \bf GGPW&0.07&0.11&4.30&55\\ \cline{2-11}
		& \bf LPW&---&---&---&---& \bf GLPW&5.64&6.38&90.49&89\\ \hline
		\hline
		\multirow{9}{*}{\rotatebox{90}{MMKP2}}& \bf BDD&1.95&2.90&346.01&116& \bf MDD&0.25&0.48&20.38&73\\ \cline{2-11}
		& \bf SWC&3.17&4.43&110.42&78& \bf GSWC&1.27&1.91&11.14&51\\ \cline{2-11}
		& \bf GTd&39.46&58.30&192.44&91& \bf GGTd&1.93&2.22&39.48&65\\ \cline{2-11}
		& \bf GT&2.35&3.88&487.41&120& \bf GGT&0.89&1.16&16.69&69\\ \cline{2-11}
		& \bf RGT&2.06&2.86&445.05&118& \bf RGGT&0.27&0.47&12.58&66\\ \cline{2-11}
		& \bf MTO&0.13&0.60&5.98&\bf \ull{43}& \bf GMTO&0.02&5&0.78&\bf \ull{23}\\ \cline{2-11}
		& \bf GPW&0.83&1.10&18.74&71& \bf GGPW&0.04&0.07&2.98&52\\ \cline{2-11}
		& \bf LPW&---&---&---&---& \bf GLPW&1.92&2.32&35.59&71\\ \hline
		\hline
		\multirow{9}{*}{\rotatebox{90}{MMKP3}}& \bf BDD&t.o.&t.o.&t.o.&318& \bf MDD&84.74&248.11&t.o.&135\\ \cline{2-11}
		& \bf SWC&167.89&558.67&t.o.&189& \bf GSWC&32.97&127.61&t.o.&101\\ \cline{2-11}
		& \bf GTd&213.55&t.o.&t.o.&223& \bf GGTd&71.75&259.29&t.o.&142\\ \cline{2-11}
		& \bf GT&t.o.&t.o.&t.o.&330& \bf GGT&48.68&208.53&t.o.&131\\ \cline{2-11}
		& \bf RGT&t.o.&t.o.&t.o.&329& \bf RGGT&46.53&202.19&t.o.&126\\ \cline{2-11}
		& \bf MTO&81.57&447.52&t.o.&\bf \ull{170}& \bf GMTO&22.91&100.24&551.00&\bf 98\\ \cline{2-11}
		& \bf GPW&114.80&507.51&t.o.&194& \bf GGPW&78.22&374.97&t.o.&172\\ \cline{2-11}
		& \bf LPW&t.o.&t.o.&t.o.&349& \bf GLPW&t.o.&t.o.&t.o.&311\\ \hline
		\hline
		\multirow{9}{*}{\rotatebox{90}{CA}}& \bf BDD&0.01&0.05&9.77&28& \bf MDD&0.01&0.04&5.33&14\\ \cline{2-11}
		& \bf SWC&0.03&0.20&27.05&37& \bf GSWC&0.02&0.09&10.89&23\\ \cline{2-11}
		& \bf GTd&---&---&---&---& \bf GGTd&---&---&---&---\\ \cline{2-11}
		& \bf GT&0.02&0.11&13.77&18& \bf GGT&0.02&0.08&6.32&17\\ \cline{2-11}
		& \bf RGT&0.02&0.08&8.65&18& \bf RGGT&0.02&0.07&3.18&13\\ \cline{2-11}
		& \bf MTO&0.02&0.05&0.35& 0& \bf GMTO&0.01&0.03&0.22& 0\\ \cline{2-11}
		& \bf GPW&0.01&0.04&\bf 0.23&\bf 0& \bf GGPW&0.01&0.03&\bf0.14&\bf 0\\ \cline{2-11}
		& \bf LPW&0.03&0.57&6.02&0& \bf GLPW&0.02&0.08&1.53&0\\ \hline
		\hline
		\multirow{9}{*}{\rotatebox{90}{MRCPSPj30}}&\bf BDD&0.03&0.07&2.17&20&\bf MDD&0.02&0.04&0.11&8\\ \cline{2-11}
		&\bf SWC&0.04&0.11&3.03&24&\bf GSWC&0.03&0.06&0.17&7\\ \cline{2-11}
		&\bf GTd&0.04&0.09&2.57&40&\bf GGTd&0.02&0.05&0.14&8\\ \cline{2-11}
		&\bf GT&0.03&0.07&1.95&\bf \ull{19}&\bf GGT&0.02&0.05&0.14&8\\ \cline{2-11}
		&\bf RGT&0.02&0.06&1.55&21&\bf RGGT&0.02&0.05&\bf \ull{0.12}&\bf \ull{7}\\ \cline{2-11}
		&\bf MTO&0.04&0.09&3.19&41&\bf GMTO&0.02&0.05&0.16&7\\ \cline{2-11}
		&\bf GPW&0.03&0.08&2.04&25&\bf GGPW&0.02&0.05&0.12&8\\ \cline{2-11}
		&\bf LPW&0.06&0.48&20.67&49&\bf GLPW&0.04&0.09&0.38&10\\ \hline
		\hline
		\multirow{9}{*}{\rotatebox{90}{RCPSP/Tj120}}&\bf BDD&0.21&4.87&257.43&117&\bf MDD&0.16&1.03&70.42&41\\ \cline{2-11}
		&\bf SWC&0.30&8.44&479.54&162&\bf GSWC&0.19&1.70&85.12&41\\ \cline{2-11}
		&\bf GTd&0.26&3.33&177.85&\bf \ull{75}&\bf GGTd&0.16&1.79&76.00&35\\ \cline{2-11}
		&\bf GT&0.25&4.83&311.58&142&\bf GGT&0.19&1.52&76.98&38\\ \cline{2-11}
		&\bf RGT&0.23&4.56&346.34&154&\bf RGGT&0.16&1.28&83.68&38\\ \cline{2-11}
		&\bf MTO&0.39&25.64&t.o.&223&\bf GMTO&0.15&6.39&209.17&66\\ \cline{2-11}
		&\bf GPW&0.32&7.91&t.o.&187&\bf GGPW&0.18&1.10&73.26&\bf \ull{26}\\ \cline{2-11}
		&\bf LPW&10.48&t.o.&t.o.&467&\bf GLPW&0.32&12.53&599.24&180\\ \hline
		\hline
		\multirow{9}{*}{\rotatebox{90}{NSP}}&\bf BDD&t.o.&t.o.&t.o.&394&\bf MDD&0.03&0.18&1.89&15\\ \cline{2-11}
		&\bf SWC&t.o.&t.o.&t.o.&374&\bf GSWC&0.06&0.34&1.76&17\\ \cline{2-11}
		&\bf GTd&t.o.&t.o.&t.o.&\bf \ull{333}&\bf GGTd&0.04&0.15&1.56&17\\ \cline{2-11}
		&\bf GT&t.o.&t.o.&t.o.&389&\bf GGT&0.05&0.33&3.36&29\\ \cline{2-11}
		&\bf RGT&t.o.&t.o.&t.o.&391&\bf RGGT&0.03&0.23&3.31&27\\ \cline{2-11}
		&\bf MTO&t.o.&t.o.&t.o.&375&\bf GMTO&0.08&0.32&1.92&23\\ \cline{2-11}
		&\bf GPW&t.o.&t.o.&t.o.&336&\bf GGPW&0.05&0.16&\bf 1.32&\bf 15\\ \cline{2-11}
		&\bf LPW&t.o.&t.o.&t.o.&400&\bf GLPW&0.73&5.71&27.49&22\\
		\hline
	\end{tabular}
	\label{table:results_cadical}
\end{table*}

\begin{table*}[!htbp]
	\caption{Solving times and number of timeouts using Glucose.}
\scriptsize
	\centering
	\begin{tabular}{|c|l|r|r|r|r||l|r|r|r|r|}
		\cline{2-11}
		\multicolumn{1}{c|}{}& \multicolumn{5}{c||}{\bf PB}&\multicolumn{5}{c|}{\bf PB(AMO)}\\ \cline{2-11}
		\multicolumn{1}{c|}{}& \bf enc. & \bf Q1 & \bf med& \bf Q3 & \bf t.o.&\bf enc. & \bf Q1 & \bf med& \bf Q3 &  \bf t.o.\\ \hline
\multirow{9}{*}{\rotatebox{90}{MMKP1}}& \bf BDD&9.80&12.18&t.o.&138& \bf MDD&3.60&8.66&144.55&104\\ \cline{2-11}
& \bf SWC&---&---&---&---& \bf GSWC&4.42&5.83&249.15&112\\ \cline{2-11}
& \bf GTd&---&---&---&---& \bf GGTd&---&---&---&---\\ \cline{2-11}
& \bf GT&12.06&15.74&t.o.&137& \bf GGT&2.57&2.77&97.69&103\\ \cline{2-11}
& \bf RGT&8.95&11.43&t.o.&136& \bf RGGT&2.49&7.14&110.08&99\\ \cline{2-11}
& \bf MTO&0.23&0.93&24.53&\bf 68& \bf GMTO&0.02&0.03&0.99&\bf 37\\ \cline{2-11}
& \bf GPW&0.92&0.96&16.01&81& \bf GGPW&0.04&0.04&6.39&69\\ \cline{2-11}
& \bf LPW&---&---&---&---& \bf GLPW&2.06&2.16&66.93&90\\ \hline
\hline
\multirow{9}{*}{\rotatebox{90}{MMKP2}}& \bf BDD&2.94&3.52&403.91&119& \bf MDD&0.19&0.29&15.79&69\\ \cline{2-11}
& \bf SWC&4.08&5.53&145.60&102& \bf GSWC&0.49&0.57&6.15&58\\ \cline{2-11}
& \bf GTd&5.33&6.98&244.01&116& \bf GGTd&2.49&8.78&71.88&99\\ \cline{2-11}
& \bf GT&2.79&3.47&491.06&122& \bf GGT&0.32&0.37&18.47&73\\ \cline{2-11}
& \bf RGT&2.36&2.92&355.14&120& \bf RGGT&0.16&0.25&11.83&69\\ \cline{2-11}
& \bf MTO&0.14&0.58&13.52&\bf 62& \bf GMTO&0.02&0.02&0.56&\bf 38\\ \cline{2-11}
& \bf GPW&0.46&0.48&14.08&80& \bf GGPW&0.02&0.03&3.65&67\\ \cline{2-11}
& \bf LPW&---&---&---&---& \bf GLPW&0.90&0.98&21.37&74\\ \hline
\hline
\multirow{9}{*}{\rotatebox{90}{MMKP3}}& \bf BDD&t.o.&t.o.&t.o.&334& \bf MDD&46.49&155.70&t.o.&104\\ \cline{2-11}
& \bf SWC&228.59&t.o.&t.o.&215& \bf GSWC&33.69&140.21&584.10&100\\ \cline{2-11}
& \bf GTd&219.36&t.o.&t.o.&215& \bf GGTd&72.66&263.05&t.o.&146\\ \cline{2-11}
& \bf GT&t.o.&t.o.&t.o.&333& \bf GGT&47.52&153.11&t.o.&109\\ \cline{2-11}
& \bf RGT&t.o.&t.o.&t.o.&332& \bf RGGT&40.92&135.16&548.87& \bf \ull{94}\\ \cline{2-11}
& \bf MTO&118.02& 521.95&t.o.&\bf 195& \bf GMTO&74.93&411.42&t.o.&172\\ \cline{2-11}
& \bf GPW&144.17&t.o.&t.o.&214& \bf GGPW&128.82&t.o.&t.o.&226\\ \cline{2-11}
& \bf LPW&t.o.&t.o.&t.o.&349& \bf GLPW&483.77&t.o.&t.o.&286\\ \hline
\hline
\multirow{9}{*}{\rotatebox{90}{CA}}& \bf BDD&0.01&0.03&5.16&20& \bf MDD&0.01&0.02&1.88&12\\ \cline{2-11}
& \bf SWC&0.01&0.35&14.47&34& \bf GSWC&0.01&0.08&3.56&11\\ \cline{2-11}
& \bf GTd&---&---&---&---& \bf GGTd&---&---&---&---\\ \cline{2-11}
& \bf GT&0.01&0.09&5.23&15& \bf GGT&0.01&0.05&2.05&9\\ \cline{2-11}
& \bf RGT&0.01&0.04&4.36&11& \bf RGGT&0.01&0.02&1.26&4\\ \cline{2-11}
& \bf MTO&0.01&0.02&0.26&0& \bf GMTO&0.01&0.01&0.13&0\\ \cline{2-11}
& \bf GPW&0.01&0.02&\bf \ull{0.09}&\bf 0& \bf GGPW&0.01&0.01&\bf\ull{0.05}&\bf 0\\ \cline{2-11}
& \bf LPW&0.00&0.34&3.01&0& \bf GLPW&0.01&0.04&0.56&0\\ \hline
\hline
\multirow{9}{*}{\rotatebox{90}{MRCPSPj30}}&\bf BDD&0.00&0.07&1.17&34&\bf MDD&0.00&0.02&0.07&13\\ \cline{2-11}
&\bf SWC&0.00&0.13&1.44&31&\bf GSWC&0.00&0.03&0.09&13\\ \cline{2-11}
&\bf GTd&0.00&0.13&1.51&49&\bf GGTd&0.00&0.03&0.11&13\\ \cline{2-11}
&\bf GT&0.00&0.07&0.97&33&\bf GGT&0.00&0.02&0.08&13\\ \cline{2-11}
&\bf RGT&0.00&0.05&\bf 0.86&\bf 31&\bf RGGT&0.00&0.02&0.07&\bf 11\\ \cline{2-11}
&\bf MTO&0.00&0.04&2.51&56&\bf GMTO&0.00&0.02&0.06&15\\ \cline{2-11}
&\bf GPW&0.00&0.05&1.09&34&\bf GGPW&0.00&0.02&0.05&13\\ \cline{2-11}
&\bf LPW&0.00&0.66&7.07&66&\bf GLPW&0.00&0.11&0.38&14\\ \hline
\hline
\multirow{9}{*}{\rotatebox{90}{RCPSP/Tj120}}&\bf BDD&0.34&7.62&t.o.&241&\bf MDD&0.09&1.65&t.o.&201\\ \cline{2-11}
&\bf SWC&0.78&10.15&t.o.&269&\bf GSWC&0.13&1.93&t.o.&194\\ \cline{2-11}
&\bf GTd&0.34&6.16&t.o.&237&\bf GGTd&0.13&2.91&t.o.&198\\ \cline{2-11}
&\bf GT&0.39&5.40&t.o.&\bf 201&\bf GGT&0.11&1.68&t.o.&208\\ \cline{2-11}
&\bf RGT&0.37&6.00&t.o.&203&\bf RGGT&0.09&1.53&t.o.&204\\ \cline{2-11}
&\bf MTO&0.44&67.49&t.o.&315&\bf GMTO&0.12&8.24&t.o.&\bf 189\\ \cline{2-11}
&\bf GPW&0.48&5.19&t.o.&282&\bf GGPW&0.11&1.18&t.o.&245\\ \cline{2-11}
&\bf LPW&3.70&t.o.&t.o.&460&\bf GLPW&0.37&6.49&t.o.&268\\ \hline
\hline
\multirow{9}{*}{\rotatebox{90}{NSP}}&\bf BDD&t.o.&t.o.&t.o.&398&\bf MDD&0.01&0.07&0.82&19\\ \cline{2-11}
&\bf SWC&t.o.&t.o.&t.o.&400&\bf GSWC&0.02&0.09&1.27&19\\ \cline{2-11}
&\bf GTd&t.o.&t.o.&t.o.&398&\bf GGTd&0.01&0.08&0.75&21\\ \cline{2-11}
&\bf GT&t.o.&t.o.&t.o.&398&\bf GGT&0.02&0.10&2.12&37\\ \cline{2-11}
&\bf RGT&t.o.&t.o.&t.o.&398&\bf RGGT&0.01&0.09&2.64&39\\ \cline{2-11}
&\bf MTO&t.o.&t.o.&t.o.&396&\bf GMTO&0.03&0.12&2.61&27\\ \cline{2-11}
&\bf GPW&t.o.&t.o.&t.o.&\bf 395&\bf GGPW&0.02&0.07&0.52&\bf \ull{13}\\ \cline{2-11}
&\bf LPW&t.o.&t.o.&t.o.&400&\bf GLPW&0.71&3.21&9.57&28\\ 
\hline

	\end{tabular}
	\label{table:results_glucose}
\end{table*}


\subsubsection{RGT and RGGT}

\begin{figure}
  \centering

  \subcaptionbox[0.95\textwidth]{The mean reduction factor per node as depth increases
    \label{fig:rggt-redu-graph-by-depth}}{
    \includegraphics{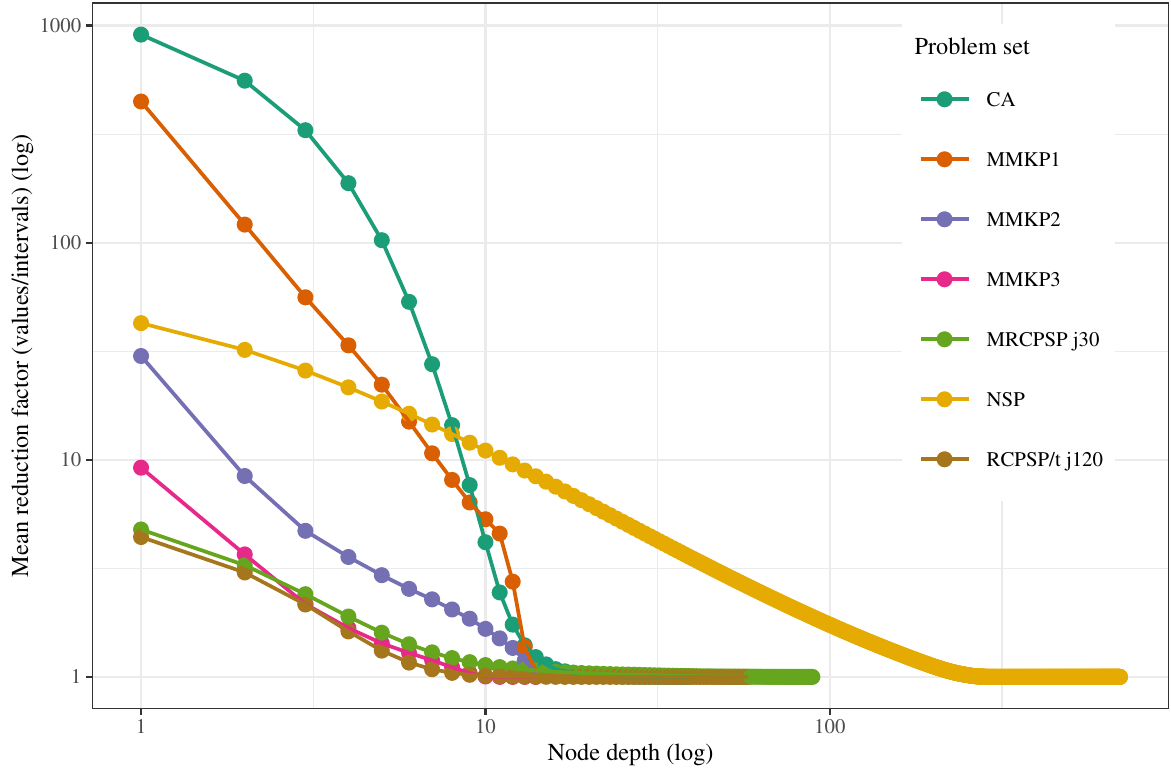} } \hfill
  \vspace{1cm}
  \subcaptionbox[0.95\textwidth]{Comparison of \# of values versus \#
    of intervals for entire GGT resp. RGGT trees
    \label{fig:rggt-redu-graph-by-tree}}{
    \includegraphics{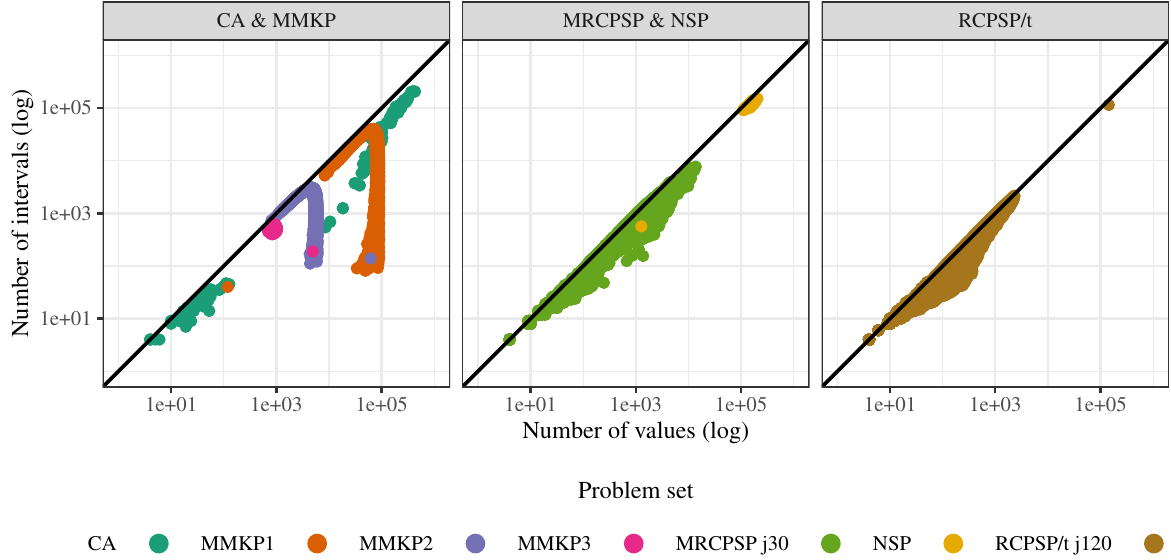}
  }
  \caption[]{Effects of reduction observed in RGGT (with MinRatio heuristic)}
  \label{fig:rggt-redu-graph}
\end{figure}

Figure~\ref{fig:rggt-redu-graph} illustrates how the reduction step of RGGT affects
the number of variables required to encode the PB(AMO) constraint (in comparison with GGT). We omit the root node because GT, GGT, RGT, and RGGT all use only one SAT variable at the root. Figure~\ref{fig:rggt-redu-graph-by-depth} shows how much reduction happens on average at different
depths in the RGGT tree. The reduction factor (defined as $\frac{\mid A.\mathit{vals} \mid}{\mid A.\mathit{intervals} \mid}$ for each node \(A\)) is largest close to the root. It is interesting that considerable reduction is still
taking place beyond depth 10 for some benchmark sets. This may be
the result of the minRatio heuristic which tends to produce
unbalanced trees (as described in \Cref{sec:heur_ggt}), spreading the leaf nodes over a wide range
of depths.  GGT leaf nodes typically have smaller $\mathit{vals}$ sets than internal nodes and are therefore
more likely to permit non-trivial intervals to be created in
their sibling node.

 Both plots show that less reduction occurs as we progress through sets
MMKP1, MMKP2, and MMKP3, just as in Table~\ref{table:results_size} the
relative reduction in SAT variables and clauses from GGT to RGGT
decreases across these three problem sets.  Recall that the
coefficients are being sampled with a maximum value of 1000, 60, and
10 respectively -- this progression reduces the likelihood of
selecting close but distinct coefficients which can be merged into an
interval.  
The reduction appears to benefit all CA instances, the highest reduction factor being observed in the middle-sized instances.
The reduction factors for MRCPSP and RCPSP/t are modest compared to most of the other problem classes. 
NSP has a significant average reduction factor even beyond depth 100, but the trees are very deep and the reduction for entire trees (\Cref{fig:rggt-redu-graph-by-tree}) is modest.

\section{Related Work}\label{sec:related}

Encodings of PB constraints based on Multi-valued Decision Diagrams (MDDs) have been extensively studied. 
It was in this context that PB(AMO) constraints were first introduced by Bofill, Coll, Suy, and Villaret~\cite{BCSV2017amopb}. Originally PB(AMO) constraints were defined in a slightly different way and were referred to as AMO-PB constraints. An MDD-based SAT encoding of AMO-PB constraints was introduced and successfully applied to solve variants of the Resource-Constrained Project Scheduling Problem (RCPSP). 
The MDD encoding approach was later revisited by the same authors~\cite{bofill2020mdd} within a more general framework named PB modulo $\cal C$, or PB($\cal C$), where $\cal C$ stands for any kind of collateral constraint over the variables of the PB constraint. 
PB(AMO) constraints were presented as a particular case of PB($\cal C$).  Other collateral constraints were also considered, such as \emph{exactly-one} and \emph{implication chains} (i.e.\ monotonically non-decreasing sequences)~\cite{bofill2020mdd}. 
Earlier, Ab\'io et al~\cite{abio2015encoding} studied decision diagram encodings of PB constraints in conjunction with implication chains, and also showed that implication chains can be used to represent AMO constraints. Bofill et al~\cite{bofill2020mdd} compared their MDD encoding of PB(AMO) constraints to Ab\'io et al's encoding~\cite{abio2015encoding} (with AMOs represented as implication chains), and it was shown that the two MDD-based techniques had similar performance. 

A number of alternatives to decision diagrams exist for encoding PB constraints into SAT. 
In this work we have reviewed a large sample of state-of-the-art SAT
encodings for PB constraints, and the related work is presented in
detail from Section~\ref{sec:enc_bdd} to Section~\ref{sec:enc_glpw}:
an encoding based on decision diagrams in Section~\ref{sec:enc_bdd};
the Sequential Weight Counter (SWC) in Section~\ref{sec:enc_gswc}; the Generalized Totalizer (GT) in
Section~\ref{sec:enc_ggt}; the n-Level Modulo Totalizer (MTO) in
Section~\ref{sec:enc_gmto}; the Global Polynomial Watchdog (GPW) in
Section~\ref{sec:enc_ggpw}; and the Local Polynomial Watchdog (LPW) in Section~\ref{sec:enc_glpw}.
Each of these encodings have been generalized to encode PB(AMO) constraints, either here or elsewhere~\cite{BCSV2017amopb}.

There are a number of other PB encodings that have not been generalized to PB(AMO) constraints, 
but most likely could be using techniques similar to the ones used here. 
For instance, Manthey et al~\cite{manthey2014more} proposed an alternative
to GPW and LPW called Binary Merger, in
which formulas $\phi$ and $\psi$ are built using sorting
networks and odd-even merger circuits~\cite{AsinNOR11}
respectively. The size bound of formulas obtained with Binary Merger is
asymptotically smaller than that of GPW. However, in this work we have
chosen the original GPW definition because in our preliminary experiments GPW has shown slightly better
performance. 
E\'en and S\"orensson~\cite{een2006translating} presented a BDD-based encoding very similar to the one 
we use, and also two further SAT encodings for PB constraints. The first,
named \emph{Sorters}, consists of a sequence of digit-wise sums
similar to GPW but also accepting mixed radix bases. A PB(AMO)
version of this
encoding could be easily defined by introducing auxiliary variables
for each group $X_i$ and bucket, as done in GGPW but considering mixed
radix bases. The other one, named \emph{Adders}, performs digit-wise
sums using a circuit of full adders and hence introducing carry
bits. Again, we could straightforwardly generalize \emph{Adders} by
introducing auxiliary variables for each bit and group $X_i$. 
In E\'en and S\"orensson's own experiments the Adders encoding was inferior to
their BDD-based encoding and the Sorters encoding~\cite{een2006translating}. 

Conjunctions of PB constraints and AMO constraints have also been
considered in the context of Mixed Integer Linear
Programming. Achterberg et al~\cite{achterberg2020presolve} describe a presolving step where
an AMO constraint is used to replace a set of 0/1 variables (taken from the scope of the AMO constraint) with an integer variable, 
and to redefine a PB constraint to use the integer variable in place of the 0/1 variables. 

Ans\'otegui et al.\ \cite{AnsoteguiBCDEMN19} integrated the MDD-based PB(AMO) encoding (described in Section~\ref{sec:enc_bdd}) into the automatic reformulation pipeline of Savile Row~\cite{aij-savilerow}. Similarly to our work, the output of that reformulation process is a SAT formula. However, in that case the input is a CP model written in Essence Prime, which is an expressive constraint programming language that supports finite domain variables and global constraints, among others. The input CP model contains a set of linear constraints, and AMO constraints over their variables are automatically detected. Some AMO constraints are detected by means of a syntactic check. Other AMOs are retrieved by detecting cliques in a graph of mutexes between pairs of literals. Mutexes are detected by probing: a literal is set to true and constraint propagation is applied to detect other literals that are incompatible with it. The focus of Ans\'otegui et al.\ is on automatic detection of AMOs, therefore it is complementary to developing new PB(AMO) encodings.


\section{Conclusions and Future Work}\label{sec:conclusions}

When solving a combinatorial problem with SAT, the size and properties of the encoding are of vital
importance. Arithmetic can be challenging to encode into SAT, and there has been a great deal of work on encoding the PB constraint in particular. 
Our focus has been on PB(AMO) constraints, which are conjunctions of one PB constraint and any number of AMO constraints. 
We have defined five new encodings for PB(AMO) constraints by generalising existing state-of-the-art encodings of PB constraints.
In each case, the size of the PB(AMO) encoding is substantially reduced compared to its corresponding PB encoding. Moreover,
the propagation properties of the original encodings are preserved in the new ones. 

We performed experiments with two recent CDCL SAT solvers (Glucose and CaDiCaL) using five problem classes:  the 
Multi-Choice Multidimensional Knapsack Problem where we can control the parameters of the PB(AMO) constraints, the Combinatorial Auctions problem, the Multi-mode Resource-Constrained Project Scheduling Problem, the Resource-Constrained Project Scheduling Problem with Time-Dependent Resource Capacities and Requests,  and the Nurse Scheduling Problem.  The new PB(AMO) encodings are dramatically smaller and more efficient than their
counterpart PB encodings. 
We have observed size reductions of an order of magnitude, and also solving
time improvements of an order of magnitude in several cases (comparing median times). 
In almost every case, the PB(AMO) encoding solves more instances within the time limit than its corresponding PB encoding.

We have also shown that there is no single best encoding for PB(AMO)
constraints, but it depends on the characteristics of the instances at hand. The
benchmark instances that we consider expose some strengths and weaknesses of the
different encodings. For example, the GGPW and GMTO encodings generate extremely small formulas, and they represent the best choices for some benchmark sets despite their poor propagation properties. 

We also contribute a new encoding that improves the Generalized Totalizer by collecting equivalent values into intervals. When applied to PB constraints, the new encoding is named Reduced Generalized Totalizer (RGT), and for PB(AMO) constraints it is the Reduced Generalized Generalized Totalizer (RGGT). In terms of size, RGT and RGGT are never worse than GT and GGT respectively, and they are often significantly better. The improvement in formula size translates into faster solving times, and in fact RGGT is the overall best choice for two of the benchmark sets.

The success of the PB(AMO) encodings immediately suggests two avenues of future work. The first is to investigate whether other constraints could be exploited in addition to the set of AMO constraints to further reduce the size of the encodings. For example, in the Combinatorial Auctions problem the AMO constraints represent cliques in a mutex graph. Other mutexes (outside the cliques) are not presently used in any way. In an encoding based on Generalized Totalizer, the additional mutexes could potentially rule out values of internal nodes, saving both variables and clauses. Similarly, in the MDD encoding the additional mutexes could potentially rule out nodes and edges. 
The second avenue is to automatically select an appropriate PB(AMO) encoding based on properties of the PB(AMO) constraint, such as the number of variables, the magnitude of the coefficients, and the sizes and number of the cells in the AMO partition. Given that there is no single best encoding, and differences in performance are often substantial, an accurate encoding selection method would be valuable.


\subsection*{Acknowledgements}
Work partially supported by
grant RTI2018-095609-B-I00 (MCIU/AEI/FEDER, UE). Jordi Coll is supported by grant  \emph{Ayudas para Contratos Predoctorales 2016} (grant number
BES-2016-076867, funded by MINECO and co-funded by FSE), and partially funded by the French Agence Nationale de la Recherche, reference ANR-19-CHIA-0013-01, and by Archimedes institute, Aix-Marseille University.
Felix Ulrich-Oltean is supported by grant EP/R513386/1 from the
UK Engineering and Physical Sciences Research Council.


\bibliography{main}

\end{document}